\documentclass{article}

\usepackage[nonatbib, main, final]{neurips_2026}
\usepackage[square,numbers]{natbib}
\usepackage[utf8]{inputenc} 
\usepackage[T1]{fontenc}    
\usepackage{hyperref}       
\usepackage{url}            
\usepackage{booktabs}       
\usepackage{amsfonts}       
\usepackage{amssymb}
\usepackage{nicefrac}       
\usepackage{array}

\usepackage{microtype}      
\usepackage{xcolor}         
\usepackage{graphicx}
\usepackage{multirow}

\usepackage{float}
\usepackage{caption}
\usepackage{subcaption}

\usepackage{tabularx}

\usepackage{amsmath}
\usepackage{pifont}
\usepackage{wrapfig}

\usepackage[dvipsnames]{xcolor}
\definecolor{blue}{HTML}{1f77b4}
\definecolor{orange}{HTML}{ff7f0e}
\definecolor{green}{HTML}{2ca02c}
\definecolor{tab-red}{HTML}{d62728}
\definecolor{purple}{HTML}{9467bd}
\definecolor{brown}{HTML}{8c564b}
\definecolor{pink}{HTML}{e377c2}
\definecolor{olive}{HTML}{bcbd22}
\definecolor{dark-purple}{HTML}{440154}
\definecolor{grey}{HTML}{808080}
\definecolor{red}{HTML}{FF0000}
\definecolor{bright-yellow}{HTML}{fde725}

\newtheorem{theorem}{Theorem}

\newcommand{\cmark}{\ding{51}}%
\newcommand{\xmark}{\ding{55}}%
\title{Data-Free Metrics Don't Work: The Missing Invariance  Under Functionally Preserving Reparameterisations}

\author{
  Gabryel Mason-Williams\thanks{Marks equal contribution.} \\
  Queen Mary University of London\\
  London, United Kingdom \\
  \texttt{g.t.mason-williams@qmul.ac.uk} \\
  \And
  Israel Mason-Williams$^*$ \\
  King's College London \& Imperial College London \\
  London, United Kingdom \\
  \texttt{israel.mason-williams@kcl.ac.uk} \\
  \AND
  Fredrik Dahlqvist \\
  Queen Mary University of London \\
  London, United Kingdom \\
  \texttt{f.dahlqvist@qmul.ac.uk} \\
}

\begin{document}

\maketitle

\begin{abstract}
Data-free methods for analysing and understanding the layers of neural networks offer many metrics for quantifying notions of `strong' versus `weak' layers, with the promise of increased interpretability. In particular,  random matrix theory (RMT) offers data-free metrics that claim predictive power over the quality of pre-trained models at a layerwise precision and the ability to identify model pathologies, indicating a unique relationship with generalisation. As a result, metrics from RMT are championed for pre-training optimisation strategies and post-training compression.  We establish that RMT-based metrics are unrelated to performance or training by showing that functionally preserving reparametrisations of a pre-trained model can have arbitrary metrics. We show this across a range of architectures and scales. To create functionally preserving reparametrisations, we exploit the well-established phenomenon of criticality: some layers can be re-initialised or re-randomised without affecting the functional behaviour of the model -- they are called robust -- while others cannot -- they are called critical. Re-initialising or re-randomising robust layers provides functionally preserving reparametrisations for which RMT-based metrics are arbitrary. Moreover, we show that relationships between metrics offered by RMT are spuriously related to generalisation. We conclude by showing that if many pre-trained models have data-free metrics in a `good' range, it is, in part, dependent on model initialisation. 
\\
\\
Project page available at: \url{https://gmw99.github.io/data-free-metrics-dont-work/}

\end{abstract}

\section{Introduction}

Understanding and interpreting deep learning models is a critical area of research, especially as the prevalence of these models increases in real-world applications. The holy grail of neural network interpretability lies in identifying computationally efficient metrics that can provide deep insights into the effectiveness of neural networks and their components. 
Data-free methods typify this endeavour by analysing the properties of neural network parameters without regard for the training or testing data, with Random Matrix Theory (RMT) being suggested as ideally suited to understanding deep neural networks~\cite[Chap. 5]{couillet2022random} with numerous applications of RMT to deep neural networks. 

Prominent among RMT-inspired data-free methods is a family of metrics that can, it is claimed, predict the quality of pre-trained models by only examining their weights. These metrics are based on the theory of Heavy-Tailed Self Regularization (HTSR) developed in~\citep{mahoney2019traditional,martin2020heavy,weightwatcher} and are described in a Nature paper~\citep{martin2021predicting} that shows empirically that well-trained models tend to have metrics lying in `good' ranges and concludes that these metrics can be used to discriminate between well-trained and poorly-trained models as well as identify model pathologies. In this paper, we will show that reversing the implication `good models $\Rightarrow$ good metrics' into `good metrics $\Rightarrow$ good models' is not only a logical fallacy, it is also empirically incorrect as we show over a range of scales and architectures that bad models can have good metrics. We also invalidate the original premise that `good models $\Rightarrow$ good metrics' by similarly showing that good models can have bad metrics.  We argue that if pre-trained models do indeed exhibit `good' metrics, it is, in part, due to our commonly used initialisation strategies and not because such data-free properties are a prerequisite for generalisation.

Scrutinising the claim that the metrics of~\citep{weightwatcher,martin2021predicting} can predict the quality of a model is important as numerous lines of research have built on this claim, with many assuming that specific data-free metric properties are fundamental for generalisation. Data-free metrics have been used in the context of layer-wise pruning ratios in LLMs~\citep{lu2024alphapruning}, explaining and understanding stages of the grokking phenomenon~\citep{prakash2025grokking}, identifying layers to apply LoRA adapters to~\citep{qing2024alphalora}, adapting layer-wise learning rates~\citep{liu-etal-2024-model} and weight decay values~\citep{he2026alphadecay}, informing early stopping conditions~\citep{he2025earlystoppingtransformermodeltraining} and improving optimizers~\citep{pang2026htmuon}; with data-free metrics often considered to have a unique relationship with generalisation. 
These metrics thus promise a lot: a better understanding of optimisation, redundancy, transitions from memorisation to generalisation, and interpretability of deep neural networks more generally. Our scrutiny of these metrics should be commensurate with these promises. 

Our strategy to understand whether data-free metrics of~\citep{weightwatcher,martin2021predicting} are causal for performance or generalisation follows the well-known work of Dinh \textit{et al.\ }\citep{dinh2017sharp} which showed that measures of flatness around local minima are not casual for generalisation. They did so by constructing functionally equivalent networks for which measures of flatness around a local minimum could be arbitrarily far apart, thereby disproving links between flatness and generalisation. We follow the same strategy, but focus on empirical results showing that functionally \emph{preserving} models can have very distant metrics, thereby disproving the link between data-free metrics and model performance or training. 

An experimentally useful class of functionally preserving models is provided by the work of Zhang \textit{et al.\ }\citep{zhang2017understanding,create-equal} that identified an interesting and unexpected phenomenon in neural network layers: some layers within a network are robust, while others are critical. A robust layer can be either re-initialisatised or re-randomised without any noticeable effect on performance. A critical layer is a layer that cannot be re-initialisatised or re-randomised without dramatically affecting the performance of the network. Re-initialisation sets the layer back to its initial parameters before training, whilst re-randomisation sets the parameters of the layer to random values by re-sampling from the same distribution used for initialisation. It was observed that in some cases, re-initialisation and re-randomisation can result in significant performance differences for a given layer, with re-initialisation maintaining performance but re-randomisation significantly degrading it~\citep{create-equal}. In other cases, re-initialisation and re-randomisation of a layer lead to a negligible difference in performance. This distinction between layer behaviours offers a core tool for testing the robustness of data-free metrics. For our purpose, re-initialisation and/or re-randomisation of robust layers will be used to build functionally preserving models for which the HTSR-based data-free metrics will take a large range of (mostly `bad') values. In other words, the metrics offered by HTSR are not reparameterisation-invariant under re-initialisations or re-randomisations of robust layers and therefore only offer spurious insights into generalisation or well-trainedness. We also show that these metrics cannot distinguish between robust and critical layers, nor predict performance differences between re-initialisation and re-randomisation. 

As a natural follow-up question from our layer-wise analysis of reparametrisation and data-free metrics, we investigate global data-free properties (average value of these metrics over all layers) and generalisation. Our results show that global data-free metric values can also be arbitrary for well-generalising models: neither layer-wise nor average data-free metrics relate to generalisation. 

We hope that our work can push the field forward in two core ways. Firstly, by providing a strong test ground for the development of data-free metrics that are reparametrisation invariant, or at least do not differ too much on functionally preserving reparametrisations of models. This is precisely what the work of~\citep{dinh2017sharp} led to, with the development of new sharpness metrics such as Relative Flatness~\citep{petzka2021relative} and Fisher-Rao norm~\citep{liang2019fisher} that are reparameterisation invariant and now considered state of the art sharpness metrics. Secondly, our work offers concrete evidence for the prioritisation of data-\emph{based} metrics to improve efforts in interpretability and compression (at least until adequate data-free metrics are discovered). 

\section{Background}\label{sec:background}

As discussed briefly above, data-free methods of interpretability aim to understand the inner workings of neural networks by studying the properties of the network parameters. Data-free approaches often focus on the matrix norm properties of layer weight matrices to understand learning or improve the performance of neural networks~\citep{yunis2024approachingdeeplearningspectral, martin2021predicting, Sanyal2020Stable, feng2022rank, salimans2016weight, NIPS2017_b22b257a,wei2022more, qing2024alphalora, thamm2024random, he2026alphadecay}. However,~\cite{create-equal} showed in their work that matrix norms, such as the Frobenius norm, are too coarse to understand the generalisation properties of neural networks.~\cite{weightwatcher,martin2021predicting} use RMT to analyse the weight matrices (excluding biases) of neural network layers through training to create their theory of heavy-tailed self-regularisation (HTSR). With this theory, they construct a set of predominately power-norm metrics related to generalisation that is applied after training to assess layer performance. Given a weight matrix $W\in \mathbb{R}^{m\times n}$ and $\lambda_i, 1\leq i\leq n$ the eigenvalues of $W^\top W$ the following metrics are defined:
\vspace{-1mm}
\begin{itemize}
    \itemsep0em
    \item $\alpha$ and weighted-$\alpha$: the exponent of the power-law that best describes the distribution of the eigenvalues $\rho(\lambda)\sim \lambda^\alpha$~\cite{weightwatcher}, and $\alpha\lambda_{\mathrm{max}}$. Estimating $\alpha$ is not straightforward~\cite[3.2]{weightwatcher} and we show through several experiments in \S~\ref{app:reliability_of_alpha} that the distribution of  eigenvalues is typically poorly modelled by power-laws, undermining the statistical basis for $\alpha$ and its derivatives.
    \item $\alpha$-norm and log-$\alpha$-norm: $\sum_{i}\lambda_i^\alpha, \ln\left(\sum_{i}\lambda_i^\alpha\right)$
    \item Frobenius and log-Frobenius norm: $\sum_{i}\lambda_i, \ln\left(\sum_{i}\lambda_i\right)$
    \item Spectral and log-spectral norm: $\lambda_{\mathrm{max}}, \ln \lambda_{\mathrm{max}}$
\end{itemize}
\vspace{-1mm}
\newcommand{\relu}{\mathrm{ReLU}}
Following in the footsteps of~\citep{dinh2017sharp}, we can already dismiss several of these metrics as potential indicators of performance or well-trainedness. For this, consider a two-layer network $\phi(x) = W_2 \cdot\relu(W_1 \cdot x)$ and the reparametrisation maps $T_b: (W_1,W_2)\mapsto (bW_1, b^{-1}W_2), b> 0$. It is clear that $\phi$ is invariant under this reparametrisation, i.e.\ $\phi(x) = b^{-1}W_2 \cdot\relu(bW_1 \cdot x)$.  Since the spectrum of $(bW^\top)bW$ is the spectrum of $W^\top W$ scaled by the constant positive factor $b^2$, it follows immediately that:
\begin{theorem}\label{thm:DinhCopy}
Let $M$ be any of the metrics above except $\alpha$. For any value $K>0$, there exists $b>0$ such that the value of $M$ applied to the layers of the network $\phi$ reparametrised by $T_b$ is greater (resp.\ smaller) than $K$.
\end{theorem}
Thus, two functionally equivalent networks can differ under all the metrics above -- except $\alpha$ -- by an arbitrary amount. It would therefore be unreasonable to expect these metrics to have any predictive power on the quality or well-trainedness of a given layer in a network. Note that taking the average of the metrics over both layers (as done in~\cite{martin2021predicting}) will lead to the exact same conclusion.

However,\space $\alpha$ itself is invariant or close to invariant (depending on estimation techniques) under $T_b$. For example, the standard maximum likelihood estimator for $\alpha$ is given by 
$\widehat{\alpha}=\frac{n}{\sum_{i}\ln(\nicefrac{\lambda_i}{\lambda_{min}})}$
which is clearly invariant under rescaling the eigenvalues. Other metrics involving quotients of eigenvalues, for example stable rank, are also outside of the scope of Theorem~\ref{thm:DinhCopy}. We will thus focus on the metric $\alpha$ in the rest of this paper (results for other metrics can be found in \S~\ref{app:data_free_metrics}). According to~\cite{weightwatcher,martin2021predicting}, well-trained models have $\alpha$ values in the optimal range $[2,6]$ for most of their layers, e.g.~\citep[Table 6, Fig. 12]{weightwatcher},~\citep[Figs. 6,7]{martin2021predicting} with values of $\alpha<2$ corresponding to overfitting/memorisation and $\alpha>6$ to underfitting (see also  \href{https://weightwatcher.ai}{https://weightwatcher.ai}).

The map $T_b$ defined above provides an example of \emph{reparametrisation}, i.e.\ of a map $g: \Theta\to\Theta$ on the space of parameters of a model, typically assumed to be a homeomorphism or at least a bijection. Note that whilst reparametrisations are often mentioned in the context of invariance properties under said reparametrisations, these are disjoint concepts, and in this work we will simply understand a reparametrisation as a homeomorphism on the space of parameters of a network without implying any invariance property. We identify a model with its defining set of parameters $\theta\in\Theta$ and say that a metric $M:\Theta\to \mathbb{R}^+$ is invariant under a reparametrisation $g$ if $M = M\circ g$, for example $\alpha=\alpha\circ T_ b$ in the example above. Any re-initialisation or re-randomisation of the weight matrix $W_l$ associated with the layer $l$ of a model defines a unique linear bijective (and thus homeomorphic) reparametrisation $g:\Theta\to\Theta$. It stands to reason that for a model $\theta$, if a reparametrisation $g(\theta)$ of this model is functionally indistinguishable from $\theta$, then any metric $M$ meant to capture model generalisation should also be (close to) invariant under $g$ for $\theta$, i.e.\ $M(\theta)\approx M(g(\theta))$.

\section{Experimental Setup}\label{sec:setup}

\begin{wrapfigure}{R}[0pt]{0.45\textwidth}
\vspace{-0.2cm}
    \centering
    \includegraphics[width=0.40\textwidth]{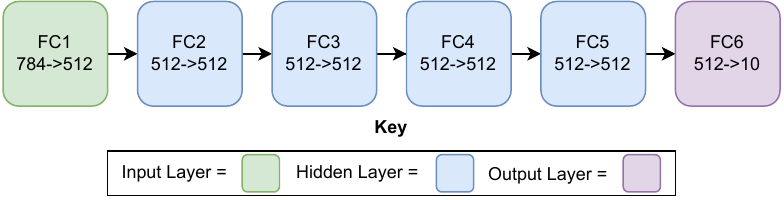}
    \caption{ReLU FCN Architecture.}
    \label{fig:simple_model}
\end{wrapfigure}
\textbf{Reparametrisation Invariance, Robust and Critical Layers:}
\cite{create-equal} demonstrated the robust and critical layer phenomenon across a range of trained architectures -- MLPs, VGGs~\citep{simonyan2014very}, ResNets~\citep{he2016deep}, Transformers~\citep{vaswani2017attention}, Vision Transformers~\citep{dosovitskiy2020image}, MLPMixers~\citep{tolstikhin2021mlpmixer} -- and of datasets -- MNIST~\citep{lecun2010mnist}, CIFAR10~\citep{cifar10}, ImageNet~\citep{deng2009imagenet} and LM1B\citep{lm1b}. 

In \S~\ref{sec:small-scale-analysis} we start with the simplest model (ReLU FCN 5x512), Figure~\ref{fig:simple_model}, and dataset (MNIST) that demonstrates this phenomenon~\cite{create-equal} in order to systematically explore the behaviour of data-free metrics under layer re-initialisations and re-randomisations of a large number of trained models. 

\noindent An added benefit of the ReLU FCN 5x512 model is that it offers a clear performance contrast between re-initialisation and re-randomisation.~\cite{create-equal} showed that residual blocks are robust to re-randomisation and attributed this to the residual layer potentially playing a lesser role in the network and thus having smaller activations than the skip connection. To analyse how effective the data-free metrics are at disambiguating between re-initialisation and re-randomisation, we analyse the correlation between data-free metrics and test accuracy using the Spearman correlation coefficient $\rho$, coefficient of determination $R^2$, the root mean square error (RMSE) of the linear regression and Kendall’s tau measure (K-$\tau$). Where $\rho$ and K-$\tau$ score of -1 indicates a very strong negative correlation, 0 indicates no correlation, and 1 indicates a very strong positive correlation and $R^2$ score of 1 indicates perfect predictions and a score of 0 indicates imperfect predictions. We use $R^2$, RMSE and Kendall's Tau measure for this study as they are the correlation metrics used by~\cite{martin2021predicting} to highlight the predictive capacity of the data-free metrics. We trained 100 ReLU FCN 5x512 models, creating 100 initialisations and 100 trained models, to obtain a representative sample of possible initialisations and trained models. The model weights and biases are initialised and re-randomised from the same distribution $\mathcal{U}(-\sqrt{k}, \sqrt{k})$ where $k$ is $\frac{1}{\text{in\_features}}$, e.g.\ FC1 has $k=\frac{1}{784}$. We record the data-free metric properties of these trained models' layers when they undergo re-initialisation and re-randomisation and observe whether these metrics:
\vspace{-1mm}
\begin{enumerate}
    \itemsep0em
    \item[\textbf{1}] Are invariant under re-initialisation and re-randomisation, particularly for robust layers, 
    \item[\textbf{2}] Can distinguish critical from robust layers,
    \item[\textbf{3}] Can predict the performance difference between re-initialisation and re-randomisation. 
\end{enumerate}
\vspace{-1mm}
In the process we demonstrate the overall lack of predictive capacity of data-free metrics after training. 

In \S~\ref{sec:large-scale}, we extend our analysis to large-scale pre-trained computer vision models (ResNet34 and ViT) both trained on ImageNet~\citep{imagenet15russakovsky}) and pre-trained large language models GPT2 and GPT2-Large evaluated on WikiText103~\citep{merity2016pointer}). As we employ pre-trained open source models we do not have access to the starting initialisation and as we result we only focus on the case of re-randomisation to test the predictive capacity of the $\alpha$ metric at this scale, concretely questions \textbf{1} and \textbf{2}. We perform re-initialisation and re-randomisation at large scale for the Pythia-1B model~\citep{pythia} and confirm our findings at the MNIST scale in Section\S~\ref{app:pythia-1B}. Furthermore, we provide empirical support for linking functional preservation to accuracy preservation under reparametrisations (re-initialisation  and re-randomisation) in the main body for MNIST and in\S~\ref{app:reparam-func-pres} for all other models explored. In \S~\ref{sec:gen_norms} we additionally explore the following metrics: Alpha Weighted ($\hat{\alpha}$, Log Alpha Norm, MP Soft Rank~\cite{weightwatcher}, Spectral Norm, Stable Rank~\cite{rudelson2007sampling} and Generalized von-Neumann Matrix Entropy~\cite{weightwatcher}: which are formally defined in \S~\ref{app:data_free_metrics}. The metrics are collected using the weightwatcher\footnote{\href{https://weightwatcher.ai}{https://weightwatcher.ai}} tool.

\noindent \textbf{Average Alpha vs Generalisation.}
We perform two sets of experiments to investigate the (lack of) relationship between average/global alpha and generalisation. In both cases we rely on the singular value decomposition (SVD) $W=U\Sigma V^T$ of the weight matrices of the model's layers. 

In our first set of experiments (\S~\ref{sec:alpha_init_training}), we study the impact of alpha \emph{at initialisation} on performance. For this, we construct a family of initialisations by power laws that interpolate the max-normalised singular values $\Sigma_\lambda$ between a given initial sampled spectrum $\Sigma_{\lambda=1}$ and
two natural endpoints: the full rank spectrum given by the identity matrix $\boldsymbol{I}$ (which corresponds to $\lambda=0, \alpha\to \infty$), and the rank 1 spectrum (1,0,..,0) (which corresponds to $\lambda=100001, \alpha\to 0$). 
For each interpolation levels $\lambda$ we initialise the weight matrices of each layer of the model as $W_\lambda = U\Sigma_\lambda V^T$ for some randomly initialised orthogonal matrices $U,V$. This controlled intervention induces different values of $\alpha$ at initialisation which allows us to systematically study how initialisation affects both performance and the value of $\alpha$ after training. In particular it allows us to assess if models whose layers are all initialised with an $\alpha$ within the `good' range $[2,6]$ achieve better than random accuracy.

In the second experiments (\S~\ref{sec:fixed_alpha}) we extend this analysis to \emph{training} with a fixed $\alpha$. Concretely, we sample from the random initialisation for each layer to obtain singular values that correspond to a desired $\alpha$ value,\space $\Sigma_{\alpha}$, and then construct $W_{\alpha} = U \Sigma_{\alpha}V^T$ at each layer and keep $\Sigma_{\alpha}$ fixed throughout training. We then solely train $U$ and $V^T$ while maintaining the unitary properties. We explore $\alpha$ within the ranges of $2 \leq \alpha \leq 6$,\space $10 \leq \alpha \leq 12$, and $18 \leq \alpha \leq 20$ for all layers.

\section{Results and Discussion}\label{sec:results}
In the main body we present our results for $\alpha$. We show that these results generalise to other metrics in~\S~\ref{app:data_free_metrics}. In the tables below, we report the mean and the standard error of the mean (SEM)~\citep{belia2005researchers}. We show that alpha is not invariant under functionality-preserving reparametrisations in small-scale (MNIST) settings in~\S~\ref{sec:small-scale-analysis}. We then show that our findings hold for other scales and modalities by exploring ImageNet models and the GPT family in~\S~\ref{sec:large-scale}, with full re-initialisation and re-randomisation for the Pythia-1B model in~\S~\ref{app:pythia-1B}. In \S~\ref{sec:data-based-metrics} we show that data-\emph{based} metrics explain the performance difference from re-initialisation and re-randomisation through their implicit relationship with the lottery ticket hypothesis~\cite{frankle2018lottery}. We then explore initialisation and average alpha, showing that the optimal alpha range is somewhat arbitrary in \S~\ref{sec:alpha_init_training} and unreliable in \S~\ref{app:reliability_of_alpha}. Additionally, in~\S~\ref{app:reparam-func-pres} we conduct a full functional similarity assessment of reparametrisation for all models to show that when accuracy is preserved so is function. Finally, we provide evidence that optimal global alpha values are not causal for well-generalising models by optimising models with arbitrary alpha values in~\S~\ref{sec:fixed_alpha} and through a full SVD ablation for learning dynamics in \S~\ref{sec:exp-spectral-vs-training}. 

\subsection{Small Scale Analysis of Alpha Under Reparametrisation and Training}
\label{sec:small-scale-analysis}
\begin{figure}[htb]
    \centering
    \begin{subfigure}[b]{0.29\linewidth}
        \includegraphics[width=\linewidth]{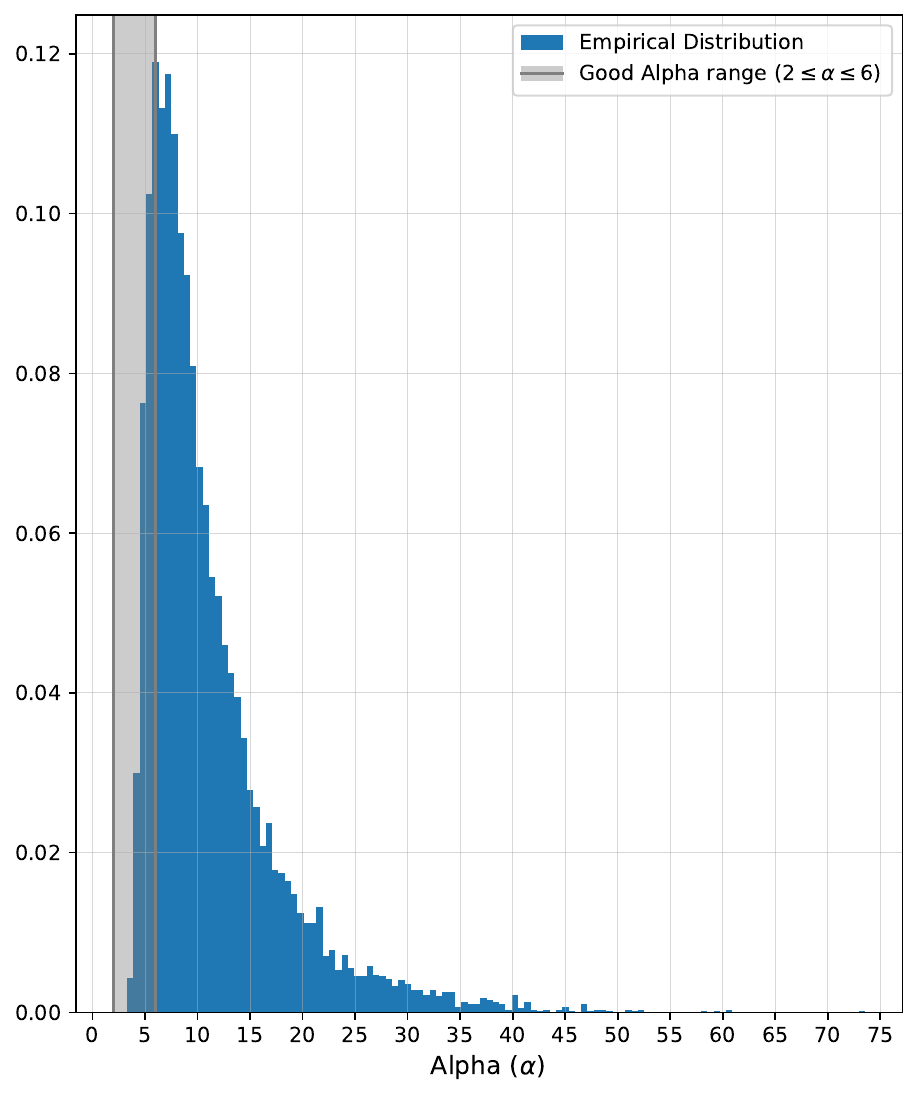}
        \caption{}
        \label{fig:alpha-init-dist}
    \end{subfigure}
    \begin{subfigure}[b]{0.7\linewidth}
        \includegraphics[width=\linewidth]{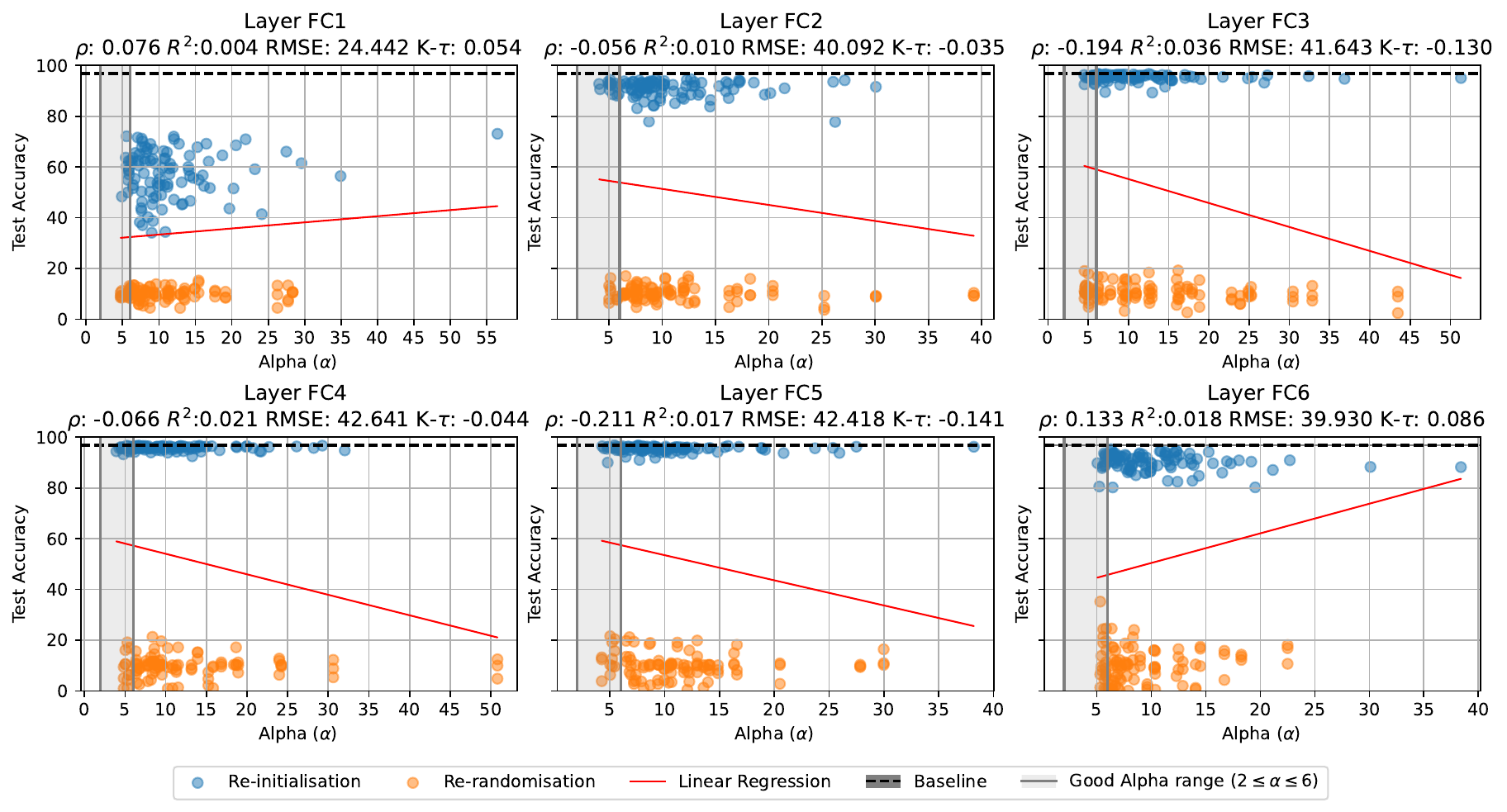}
        \caption{}
        \label{fig:alpha-init-vs-rand}
    \end{subfigure}
    \caption{(a) Empirical distribution of $\alpha$ values on a 512x512 fully connected layer, sampled from 10,000 initialisations. (b) Layer re-initialisation ({\color{blue}\textbf{$\bullet$}}) and re-randomisation {(\color{orange}{\textbf{$\bullet$}}}) test accuracy vs $\alpha$. In~\S~\ref{app:init_method_and_aplha} we present the descriptive statistics for sub-figure (a) along with other initialisation schemes.}
\end{figure}

\noindent \textbf{Quality of training vs alpha.} The idea that well-trained layers have $\alpha\in [2,6]$ seems to be borne out by training 100 ReLU FCN 5x512 models to good values of test accuracy, achieving $\alpha\in [2,6]$ for every layer (see Table~\ref{tab:mean_trained_alpha}).  
Whilst well-trained may imply $\alpha\in[2,6]$ (but more on this below), a simple experiment shows that the converse does not hold. We plot the empirical distribution of $\alpha$ for a 512x512 fully connected layer in Figure~\ref{fig:alpha-init-dist}, sampled from 10,000 potential initialisations. The resulting distribution shows that an initialised, \emph{untrained} layer of this network, can fall, with a small but non-negligible probability, within the optimal $\alpha$ value range of 2 and 6. This highlights that a random layer can be considered optimal under RMT which suggests a core failing of the metric.

\begin{table}[H]
\centering
\caption{Alpha ($\alpha$) of the layers in ReLU FCN 5x512 and test accuracy of the model on MNIST.}
\label{tab:mean_trained_alpha}
\resizebox{\textwidth}{!}{%
\begin{tabular}{|c|cccccc|c|}
\hline
\multirow{2}{*}{\textbf{Metric}} & \multicolumn{6}{c|}{\textbf{Layer}}                                                                                                                                                                                           & \multirow{2}{*}{\textbf{Test Accuracy}} \\ \cline{2-7}
                                 & \multicolumn{1}{c|}{\textbf{FC1}}     & \multicolumn{1}{c|}{\textbf{FC2}}      & \multicolumn{1}{c|}{\textbf{FC3}}      & \multicolumn{1}{c|}{\textbf{FC4}}      & \multicolumn{1}{c|}{\textbf{FC5}}      & \textbf{FC6}      &                                         \\ \hline
Alpha ($\alpha$)                 & \multicolumn{1}{c|}{4.820 $\pm$ 0.025} & \multicolumn{1}{c|}{4.205 $\pm$ 0.039} & \multicolumn{1}{c|}{4.126 $\pm$ 0.038} & \multicolumn{1}{c|}{4.135 $\pm$ 0.035} & \multicolumn{1}{c|}{4.193 $\pm$ 0.034} & 3.793 $\pm$ 0.805 & \multicolumn{1}{l|}{96.822 $\pm$ 0.057} \\ \hline
\end{tabular}%
}
\end{table}

\vspace{-2mm}

\noindent \textbf{Reparametrisation-invariance of alpha.} 
Next, we perform independent re-initialisations ({\color{blue}{\textbf{$\bullet$}}}) and re-randomisations {(\color{orange}{\textbf{$\bullet$}}}) in Figure~\ref{fig:alpha-init-vs-rand} of each layer for 100 trained networks and record the impact on the networks' test accuracy and alpha values. Figure~\ref{fig:alpha-init-vs-rand} shows that a non-negligible proportion of networks whose performance is severely degraded by re-randomisation maintain a good value of $\alpha$ around 5. Conversely, many networks maintain good performance after re-initialisations (particularly of layers FC3-FC5) but with $\alpha$ values significantly outside of $[2,6]$. As is clear from Figure~\ref{fig:alpha-init-vs-rand}, whilst all values of $\alpha$ start in the `good' range $[2,6]$, reparametrisation in the form of re-initialisation or re-randomisation scatters these values over a very wide range (typically deep into `underfitted' territory), irrespective of the change in accuracy incurred by this reparametrisation. Alpha is not invariant under these reparametrisations, even when accuracy is (i.e.\ on robust layers).

\noindent \textbf{The robust vs critical phenomenon and alpha.}
Figure~\ref{fig:alpha-init-vs-rand} shows the stark contrast in how layers respond to re-initialisation and re-randomisation. For example, we only see a large drop in test accuracy when applying re-initialisation to the Layer FC1, re-initialising other layers leaves performance almost unchanged. From this perspective, FC1 is a critical layer whilst FC2-FC6 are robust (to re-initialisation). In this experiment $\alpha$ cannot distinguish between these behaviours, the range of values taken by $\alpha$ shows no noticeable difference between critical and robust layers.

\noindent \textbf{The re-initialisation vs re-randomisation phenomenon and alpha.}
We observe different results when re-randomisation is applied. We find that re-randomising any layer degrades accuracy to circa random accuracy on the test set, in other words none of the layers are robust to re-randomisation. Surprisingly, this is not reflected in the corresponding $\alpha$ values of these two conditions, as the distribution of $\alpha$ values is relatively similar for each layer and each condition. If $\alpha$ had predictive power we would observe a negative correlation between re-initialisation (corresponding to `good' alphas) to re-randomisation (corresponding to `bad' alphas). However, there is no difference between the $\alpha$ values of re-initialisation and re-randomisation, with a mean Spearman correlation coefficient and Kendall's tau measure across layers of -0.053 and -0.035, respectively. Thus $\alpha$ does not distinguish between these behaviours either. 

\noindent \textbf{Generalisation of Reparametrisation Invariance to other Data-Free Metrics.}
We extend our analysis to other data-free metrics that have been defined in the literature. In spite of Theorem~\ref{thm:DinhCopy} and the lack of predictive power of the $\alpha$ metric described above, it is important to verify across other representative measures that data-free metrics are not invariant under reparametrisation (of robust layers), and cannot disambiguate between critical and robust layers, nor between re-initialisation and re-randomisation. When we conduct a similar analysis across six other metrics, Alpha Weighted, Log Alpha Norm, MP Soft Rank, Spectral Norm, Stable Rank and Generalized von-Neumann Matrix Entropy we find, as shown in \S~\ref{app:data_free_metrics} Figure~\ref{fig:genralised_fig}, that we can draw exactly the same conclusion as with $\alpha$. In~\S~\ref{sec:data-based-metrics} we show that data-based metrics offer some predictive capacity in reference to criticality.

\noindent \textbf{Reparametrisation, Functional Similarity and Accuracy Preservation.} 
 As we perform interventions on particular layers of specific models, preserving accuracy can be seen as a proxy for function preservation. To quantify the faithfulness of this proxy in our experiments, we employ functional and representational similarity measures~\citep{klabunde2305similarity} to quantify the output similarity of the original model and the model with the intervention (re-initialisation, Table~\ref{tab:MNIST_functional_preservation_re_init}, or re-randomisation, Table~\ref{tab:MNIST_functional_preservation_re_rand}) on the test set. In Table~\ref{tab:MNIST_functional_preservation_re_init} we observe that applying re-initialisation across layers leads to varying levels of functional preservation, with layers that have a low relative change in accuracy (FC2-FC5) maintaining functional similarity across the respective metrics while layers with higher accuracy deviation (FC1 and FC6) preserving less of the predictive structure of the original model. In the case of re-randomisation in Table~\ref{tab:MNIST_functional_preservation_re_init}, there are persistent high deviations in accuracy which is accompanied by increased divergence from the original model's output distribution across similarity metrics. Overall, the correlation between functional similarity and model accuracy presented in Table~\ref{tab:MNIST_functional_correlation} (for several functional similarity and correlation metrics) shows that increased functional similarity corresponds to increased accuracy, justifying the following statement: 

\textit{When accuracy is preserved under the reparametrisations offered by re-initialisation and re-randomisation predictive structure of the original model is approximately preserved too}. 

Analysis of larger models (ResNet's,ViT's, GPT2 and Pythia-1B) in~\S~\ref{app:reparam-func-pres} confirms this conclusion.

\begin{table}[H]
\centering
\caption{Functional preservation of the ReLU FCN 5x512 by layer under the re-initialisation reparametrisation. ($\uparrow/$$\downarrow$) denotes the preferred direction for each metric.}
\label{tab:MNIST_functional_preservation_re_init}
\resizebox{\linewidth}{!}{
\begin{tabular}{|c|c|c|c|c|c|c|}
\hline
\textbf{Model Layer} & \textbf{$\Delta$ Accuracy} & \textbf{$\Delta$ Loss} & \textbf{Prediction Similarity ($\uparrow$)} & \textbf{Jensen-Shannon Distance ($\downarrow$)}      & \textbf{Procrustes Distance ($\downarrow$)} & \textbf{KL Divergence ($\downarrow$)}   \\ \hline 
FC1                  & -39.859 $\pm$ 0.936   & 1.226 $\pm$ 0.038      & 0.574 $\pm$ 0.009       & 0.459 $\pm$ 0.003 & 0.340 $\pm$ 0.003   & 1.014 $\pm$ 0.024 \\ \hline
FC2                  & -6.330 $\pm$ 0.328    & 0.285 $\pm$ 0.007      & 0.917 $\pm$ 0.003       & 0.272 $\pm$ 0.002 & 0.087 $\pm$ 0.001   & 0.267 $\pm$ 0.004 \\ \hline
FC3                  & -1.899 $\pm$ 0.131    & 0.297 $\pm$ 0.003      & 0.962 $\pm$ 0.001       & 0.294 $\pm$ 0.001 & 0.053 $\pm$ 0.001   & 0.289 $\pm$ 0.002 \\ \hline
FC4                  & -1.384 $\pm$ 0.080    & 0.317 $\pm$ 0.003      & 0.968 $\pm$ 0.001       & 0.308 $\pm$ 0.001 & 0.051 $\pm$ 0.001   & 0.314 $\pm$ 0.002 \\ \hline
FC5                  & -1.716 $\pm$ 0.114    & 0.364 $\pm$ 0.003      & 0.963 $\pm$ 0.001       & 0.329 $\pm$ 0.001 & 0.054 $\pm$ 0.001   & 0.356 $\pm$ 0.002 \\ \hline
FC6                  & -6.966 $\pm$ 0.332    & 0.522 $\pm$ 0.004      & 0.910 $\pm$ 0.003       & 0.383 $\pm$ 0.001 & 0.061 $\pm$ 0.001   & 0.487 $\pm$ 0.003 \\ \hline
\end{tabular}}
\end{table}

\begin{table}[H]
\centering
\caption{Functional preservation of the ReLU FCN 5x512 by layer under the re-randomisation reparametrisation. ($\uparrow/$$\downarrow$) denotes the preferred direction for each metric.}
\label{tab:MNIST_functional_preservation_re_rand}
\resizebox{\linewidth}{!}{
\begin{tabular}{|c|c|c|c|c|c|c|}
\hline
\textbf{Model Layer} & \textbf{$\Delta$ Accuracy} & \textbf{$\Delta$ Loss} & \textbf{Prediction Similarity ($\uparrow$)} & \textbf{Jensen-Shannon Distance ($\downarrow$)}      & \textbf{Procrustes Distance ($\downarrow$)} & \textbf{KL Divergence ($\downarrow$)}   \\ \hline 
FC1   & -86.928 $\pm$ 0.217 & 3.704 $\pm$ 0.052 & 0.102 $\pm$ 0.002 & 0.721 $\pm$ 0.001 & 0.894 $\pm$ 0.003 & 2.601 $\pm$ 0.020 \\ \hline
FC2   & -86.692 $\pm$ 0.273 & 2.533 $\pm$ 0.019 & 0.104 $\pm$ 0.003 & 0.712 $\pm$ 0.001 & 0.806 $\pm$ 0.005 & 2.303 $\pm$ 0.010 \\ \hline
FC3   & -86.562 $\pm$ 0.332 & 2.348 $\pm$ 0.009 & 0.105 $\pm$ 0.003 & 0.711 $\pm$ 0.000 & 0.647 $\pm$ 0.008 & 2.237 $\pm$ 0.006 \\ \hline
FC4   & -87.350 $\pm$ 0.415 & 2.300 $\pm$ 0.007 & 0.097 $\pm$ 0.004 & 0.711 $\pm$ 0.000 & 0.466 $\pm$ 0.008 & 2.228 $\pm$ 0.006 \\ \hline
FC5   & -87.007 $\pm$ 0.477 & 2.315 $\pm$ 0.009 & 0.101 $\pm$ 0.005 & 0.711 $\pm$ 0.001 & 0.335 $\pm$ 0.007 & 2.232 $\pm$ 0.006 \\ \hline
FC6   & -86.848 $\pm$ 0.680 & 2.356 $\pm$ 0.012 & 0.102 $\pm$ 0.007 & 0.711 $\pm$ 0.001 & 0.221 $\pm$ 0.007 & 2.263 $\pm$ 0.011 \\ \hline
\end{tabular}}
\end{table}

\begin{table}[htb]
\centering
\caption{Correlation between the change in accuracy due to re-initialisation (Table~\ref{tab:MNIST_functional_preservation_re_init}) and re-randomisation (Table~\ref{tab:MNIST_functional_preservation_re_rand}) and the functional preservation of the ReLU FCN 5x512.}
\footnotesize
\label{tab:MNIST_functional_correlation}
\begin{tabular}{|c|c|c|c|c|}
\hline
\textbf{Similarity Metric}   & \textbf{$\boldsymbol{\rho}$} & \textbf{$\boldsymbol{R^2}$}  & \textbf{RMSE} & \textbf{K-$\boldsymbol{\tau}$} \\ \hline
Prediction Similarity       & 0.997           & 1.000          & 0.004         & 0.964             \\ \hline
Jensen-Shannon Distance     & -0.869          & 0.977          & 0.030         & -0.672            \\ \hline
Procrustes Distance         & -0.809          & 0.635          & 0.189         & -0.604            \\ \hline
KL Divergence & -0.864          & 0.976          & 0.149         & -0.665            \\ \hline
\end{tabular}
\end{table}

\subsection{Large Scale Analysis of Alpha Under Reparametrisation}
\label{sec:large-scale}
This section explores the predictive power of alpha with respect to performance in large scale models that have been trained for high-complexity tasks using the reparametrisation by re-randomisation setup at the hundred million parameter scale.
We find, in line with our small scale experiments, that alpha has no predictive capacity in this regime either. For these experiments, we use pre-trained open source models, and we therefore have no access to pre-training initialisations. As a result, we only focus on reparametrisation by re-randomisation to test the invariance and predictive capacity of the $\alpha$ metric. The results in this section are  consistent with other data-free metrics in this experimental setup which we present in \S~\ref{app:ResNet34},~\ref{app:ViT},~\ref{app:GPTsmall} and~\ref{app:GPTlarge}.

\begin{wraptable}{r}{0.4\textwidth}
  \centering
  \caption{Model test performance on ImageNet and mean layer Alpha ($\alpha)$ value.}
  \label{tab:imagenet}
    \resizebox{0.4\textwidth}{!}{
    \begin{tabular}{|c|c|c|c|c|}
    \hline
    \textbf{Model} & \textbf{\begin{tabular}[c]{@{}c@{}}Number of\\ Parameters\end{tabular}} & \textbf{Loss} & \textbf{Accuracy} & \textbf{\begin{tabular}[c]{@{}c@{}}Mean Layer\\ Alpha Value\end{tabular}} \\ \hline
    ResNet34       & 21.8M                                                                   & 1.071         & 73.302            & 3.380 $\pm{}$ 0.195                                                       \\ \hline
    ViT            & 86.6M                                                                   & 0.848         & 79.136            & 4.287 $\pm{}$ 0.231                                                       \\ \hline
    \end{tabular}}
\end{wraptable}
\noindent \textbf{ImageNet.} We explore the relationship of a layer's $\alpha$ value with the corresponding performance of the model on ImageNet with pre-trained ResNet34 and ViT models. The respective performance and model details are shown in Table~\ref{tab:imagenet}; from this table it can be observed that these models are well-trained for the task of ImageNet. In line with the RMT perspective, the competitive performance of these models is aligned with a mean $\alpha$ value across layers within the optimal range.

\begin{figure}[b]
    \centering
    \begin{subfigure}[b]{0.49\linewidth}
        \includegraphics[width=\linewidth]{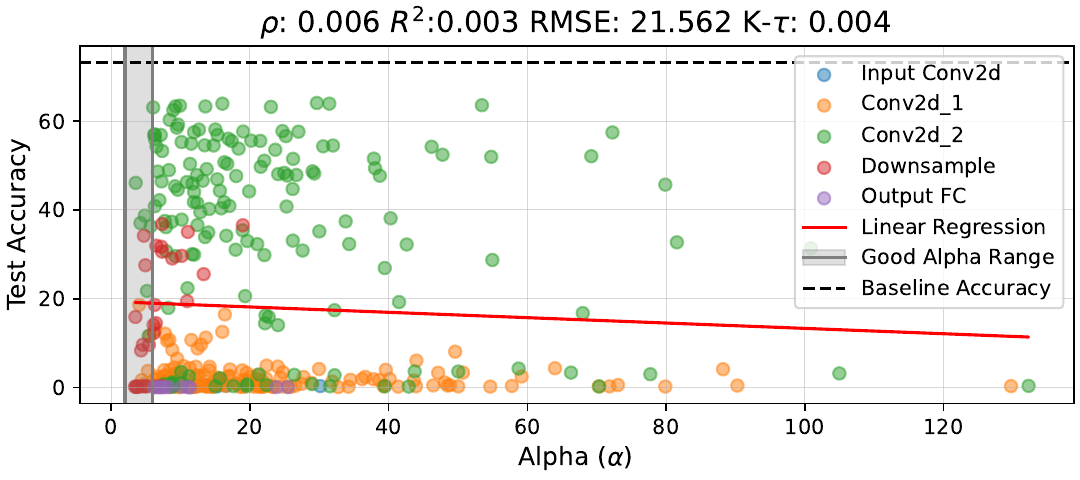}
        \caption{ResNet34}
    \end{subfigure}
    \begin{subfigure}[b]{0.49\linewidth}
        \includegraphics[width=\linewidth]{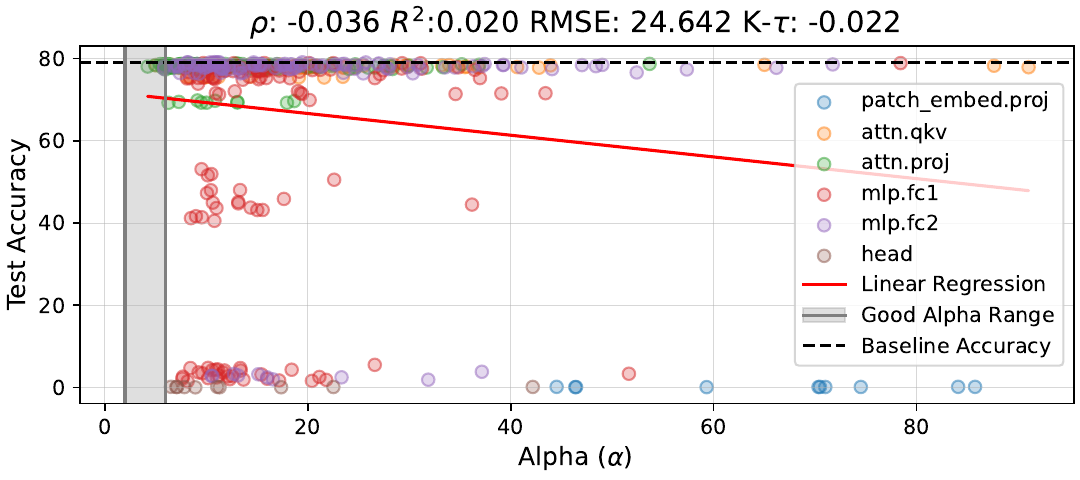}
        \caption{ViT}
    \end{subfigure}
    \caption{Layer re-randomisation Test Accuracy vs Alpha ($\alpha$) on ImageNet. See stratified analysis in Appendix \ref{app:stratified_analysis}.}
    \label{fig:imagenet}
\end{figure}

To conduct this experiment we randomise a layer independently and then record the corresponding test accuracy and  $\alpha$ value post re-randomisation. Given the different implicit biases offered by the ResNet and ViT we believe that our selection of models is representative for the core architectural differences, allowing a robust analysis of the impacts of re-randomisation and $\alpha$ metrics at scale. We repeat our re-randomisation process 10 times per layer to capture stochastic variation in randomisations. In both ResNet34 and ViT, Figure~\ref{fig:imagenet}, we find no relationship between $\alpha$ and test accuracy, as found in the small scale experiment. Both models can be re-randomised to $\alpha$ values outside of the desired range and maintain accuracy. The mean Spearman correlation coefficients for alpha and accuracy are 0.006 and -0.036 respectively, and Kendall’s tau values are 0.004 and -0.022, respectively. In the case of ResNet34 we see that the model is not so robust to re-randomisation, having a lower accuracy than the baseline; a large proportion of layers with optimal $\alpha$ values result in a model with essentially random accuracy, i.e.\ `good alpha does not imply good model'. The ViT is more robust to layer re-randomisation and represents the inverse case where many layers are significantly outside of the optimal $\alpha$ range but display a high preservation of the baseline accuracy, i.e.\ `good model does not imply good alpha'. As a result, these results confirm in large scale settings that $\alpha$ is not invariant under reparametrisation by re-randomisation, even on robust layers, and has very limited predictive capacity over post re-randomisation performance. 

\begin{figure}[H]
    \centering
    \begin{subfigure}[b]{0.49\linewidth}
        \includegraphics[width=\linewidth]{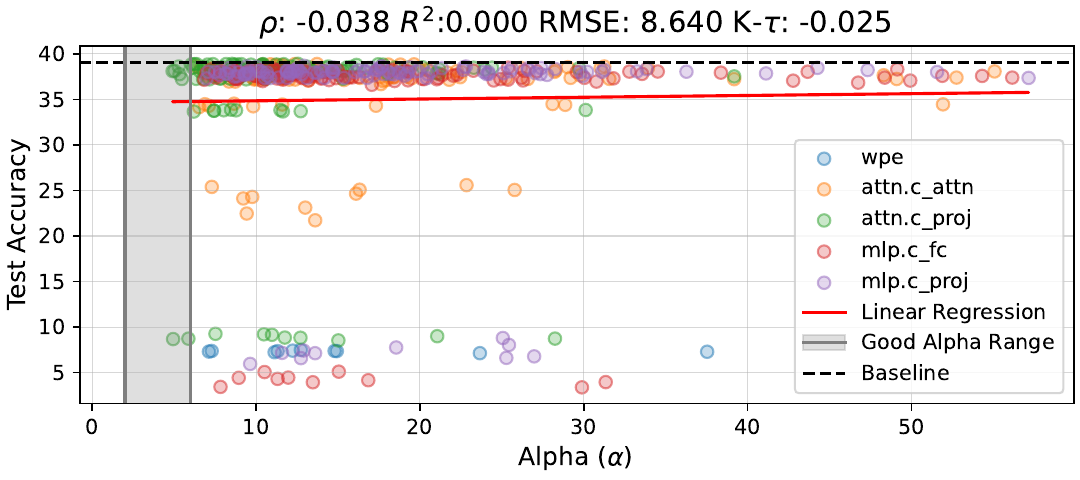}
        \caption{GPT2}
    \end{subfigure}
    \begin{subfigure}[b]{0.49\linewidth}
        \includegraphics[width=\linewidth]{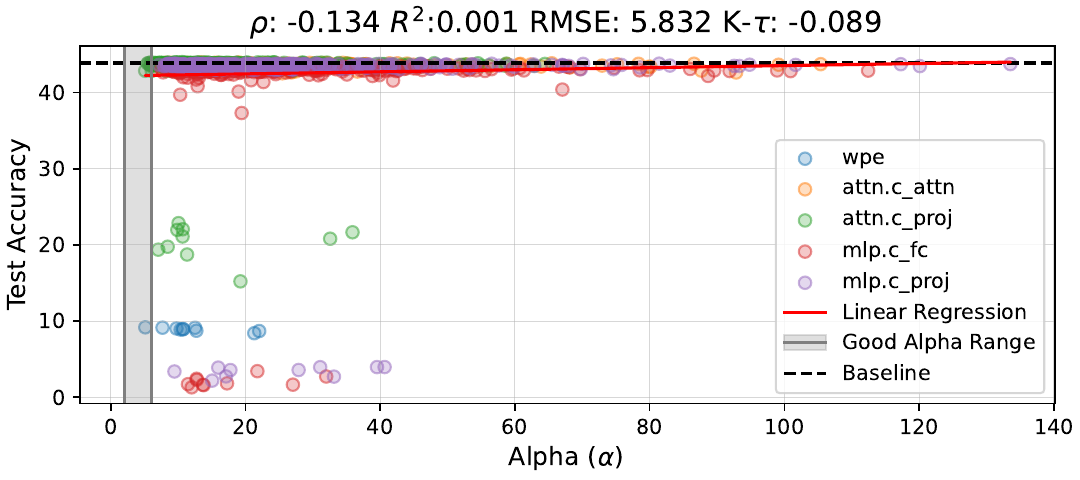}
        \caption{GPT2-Large}
    \end{subfigure}
    \caption{Layer re-randomisation Test Accuracy vs $\alpha$ on WikiText103. See stratified analysis in Appendix \ref{app:stratified_analysis}}
    \label{fig:gpt}
\end{figure}

\begin{wraptable}{r}{0.45\textwidth}
  \centering
    \caption{Model test performance on WikiText-103 and mean layer Alpha ($\alpha)$ value.}
    \label{tab:gpt}
    \resizebox{0.45\textwidth}{!}{
    \begin{tabular}{|c|c|c|c|c|c|}
    \hline
    \textbf{Model} & \textbf{\begin{tabular}[c]{@{}c@{}}Number of\\ Parameters\end{tabular}} & \textbf{Loss} & \textbf{Accuracy} & \textbf{Perplexity} & \textbf{\begin{tabular}[c]{@{}c@{}}Mean Layer\\ Alpha Value\end{tabular}} \\ \hline
    GPT            & 127M                                                                    & 3.399    & 39.095     & 29.941              & 3.865 $\pm$ 0.136                                                         \\ \hline
    GPT Large      & 774M                                                                    & 2.967 & 43.877 & 19.436              & 4.013  $\pm$ 0.091                                                        \\ \hline
    \end{tabular}}
\end{wraptable}
\noindent \textbf{Large Language Models.} We confirm our negative results for the predictive capacity of the $\alpha$ metric in the vision domain under re-randomisation, by examining a different modality. We extend our exploration of the relationship between layer $\alpha$ values and the corresponding performance of the model to language models on WikiText103 with pre-trained GPT2 and GPT2-Large. The respective performance and model details is shown in Table~\ref{tab:gpt}. Both models have competitive performance with low loss and perplexity on the WikiText-103~\cite{merity2016pointer} datasets and, in-line with HTSR Theory, have mean layer $\alpha$ values that are within the optimal range. 
We use the pre-trained models as the base model and then we randomise a layer and record its $\alpha$ value and the corresponding test perplexity and loss of the model.  We do this 10 times for each layer to obtain a representative set of random layers. For both GPT2 and GPT2-Large, Figure~\ref{fig:gpt}, we observe that their is no relationship between the performance of the model and a layer's $\alpha$ value, with a mean Spearman correlation coefficient of -0.038, -0.134  and Kendall’s tau measure of -0.025 and -0.089 for GPT2 and GPT2-Large respectively. We also find that randomising layers in the GPT2 and GPT2-Large model often has a negligible affect on the performance of the model irrespective of the $\alpha$ value associated with the layer. In other words, re-randomising generates functionally preserving reparametrisations of the original model, but with arbitrary values of alpha for the layer under consideration. Furthermore, when re-randomising robust layers, we observe that most of these layers lie far outside the optimal range of $\alpha$. Again, we see that $\alpha$ is not invariant under reparametrisation by re-randomisation, even on robust layers, and has no predictive power over performance. 

\subsection{Alpha At Initialisation \& Training}
\label{sec:alpha_init_training}
The previous sections (\S~\ref{sec:small-scale-analysis} and \S~\ref{sec:large-scale}) showed that the $\alpha$ of a layer has no (causal) relationship with model performance. However, it could be argued that the metric is best interpreted as a global metric averaged across all layers, as done in~\cite{martin2021predicting}. Therefore, we explore the average model $\alpha$ and its relationship to generalisation. In this section we report on the first of the two sets of SVD-based experiments described in \S~\ref{sec:setup}. All experiments are carried out on the ReLU FCN 5x512 model of \S~\ref{sec:small-scale-analysis}.

\begin{figure}[H]
    \centering
    \includegraphics[width=0.75\linewidth]{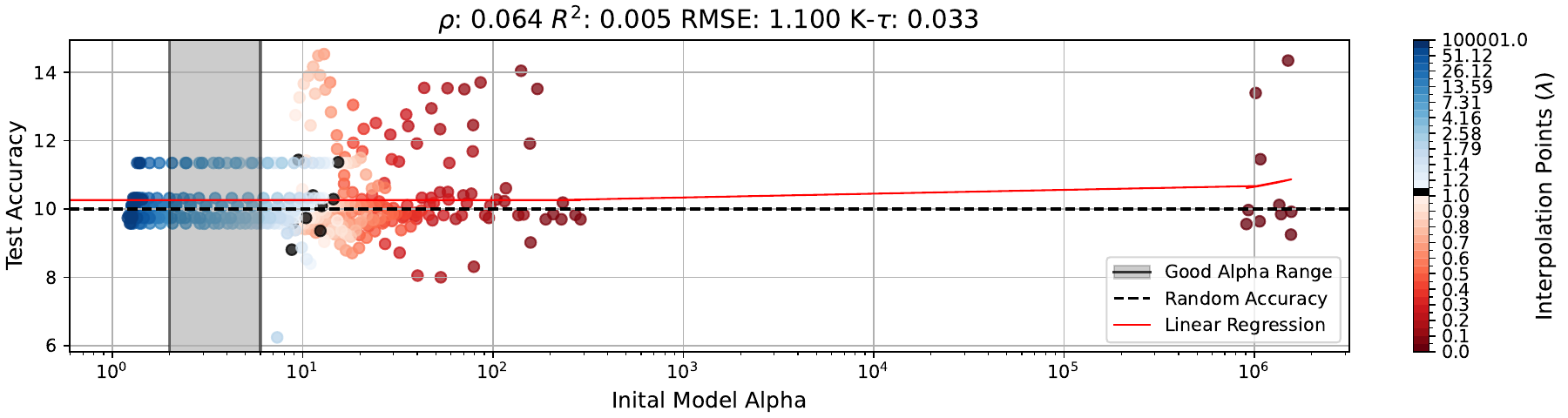}
    \caption{Test Accuracy vs Model $\alpha$ at initialisation.}
    \label{fig:model_init_alpha_test_acc}
\end{figure}

\begin{figure}[H]
    \centering
    \includegraphics[width=0.75\linewidth]{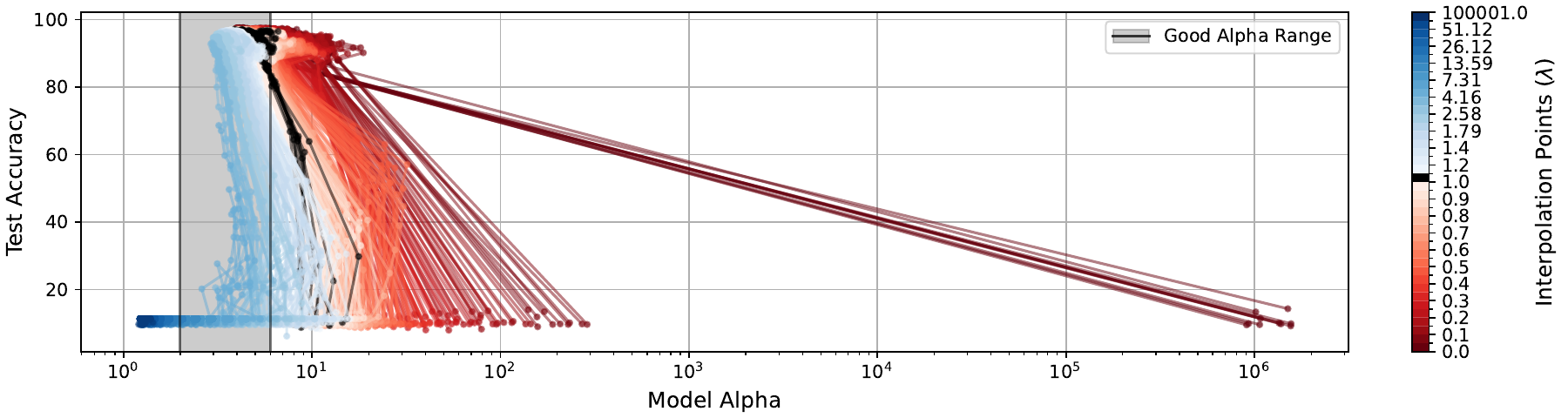}
    \caption{Test Accuracy vs Model $\alpha$ through training.}
    \label{fig:model_alpha_test_acc_through_training}
\end{figure}

\begin{figure}[H]
    \centering
    \includegraphics[width=0.8\linewidth]{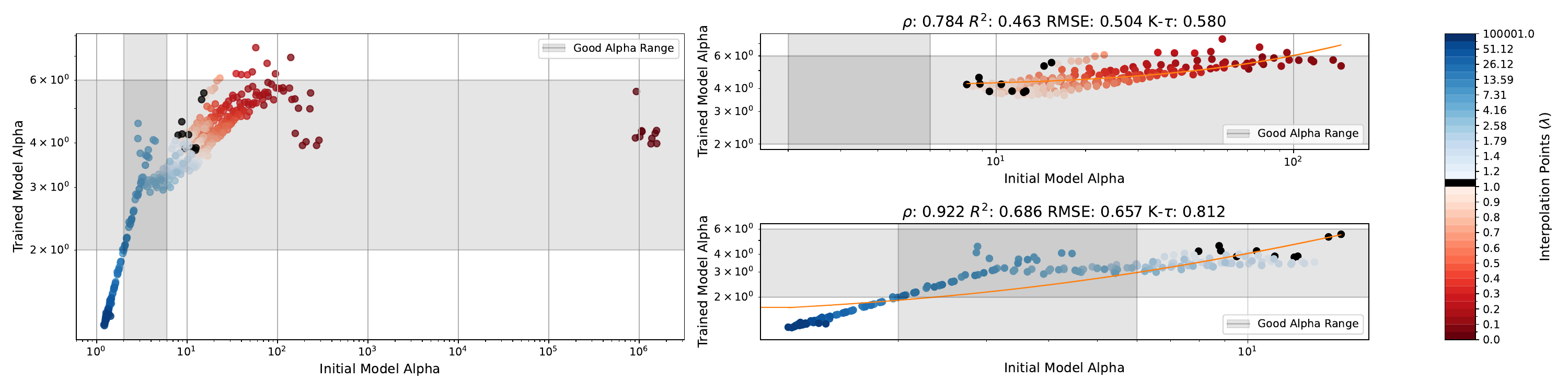}
    \caption{Trained model $\alpha$ vs initial model $\alpha$ (\textbf{Left}). Trained model $\alpha$ vs initial model $\alpha$ for interpolations $\lambda=0.1$ to $\lambda=1$ (\textbf{Top Right}) and $\lambda=1$ to $\lambda=100001$ (\textbf{Bottom Right}).}
    \label{fig:model_init_vs_model_end}
\end{figure}

\noindent\textbf{Alpha and Initialisation.}
In Figure~\ref{fig:model_init_alpha_test_acc} we explore the test accuracy of untrained models initialised with a specific model $\alpha$ following the singular value interpolation $\Sigma_\lambda$ described in \S~\ref{sec:setup}. The ten black dots ($\bullet$) represent standard uniform initialisations of a model, i.e.\ ten starting spectra $\Sigma_1$, that each generate 20 interpolations spectra towards $(1,0,..,0)=\Sigma_\infty$ (\textcolor{blue}{$\bullet$}) and another 20 interpolations  towards $\boldsymbol{I}=\Sigma_{0}$ (\textcolor{red}{$\bullet$}). We observe that models within the `good' range $2 \leq \alpha \leq 6$ (the \textcolor{grey}{\textbf{grey}} region in Figure~\ref{fig:model_init_alpha_test_acc}) exhibit no performance increase versus models outside of this range (\textcolor{blue}{$\bullet$} for $\alpha < 2$ and \textcolor{red}{$\bullet$} for $\alpha > 6$) although a global alpha in this range is meant to be indicative of a good, well trained model. Figure~\ref{fig:model_init_alpha_test_acc} thus highlights the lack of causal relationship with test accuracy. We observe negligible correlation between an untrained model's initial $\alpha$ and its untrained test accuracy. 
\noindent \textbf{Alpha and Training.}
As noted in~\citep{weightwatcher,martin2021predicting}, models that generalise do consistently have an average model $\alpha\in [2,6]$, which does suggest that it is an interesting property of a model. If, as we have shown repeatedly,\space $\alpha$ is not connected to generalisation, how can we explain this empirical observation? To answer this question we explore how $\alpha$ changes throughout training in Figure~\ref{fig:model_alpha_test_acc_through_training}, where the lines of the black dots ($\bullet$) represent the standard initialisations training trajectories. The models initialised with $\alpha$ between 1 and 3 fail to generalise and maintain the corresponding $\alpha$ value throughout training. The rest of the models initialised within the good $\alpha$ range do generalise and stay within this range throughout training. The model's initialised with an $\alpha>6$ do generalise throughout training, and witness a decrease in model alpha tending towards the `good' range of $2 \leq \alpha \leq 6$ in training. 
Figure~\ref{fig:model_init_vs_model_end} shows a positive correlation between the initialisation $\alpha$ and the trained $\alpha$ of the models. Models initialised using the standard uniform sampling (i.e.\ with $\Sigma_1$ initialisations) land in the `good' range $[2,6]$, initialisations $\Sigma_\lambda, \lambda<1$ land beyond $\alpha>6$. It is thus likely that if most trained models have an average $\alpha\in[2,6]$, it is at least in part due to the commonly used initialisation procedures.
It should be noted that the correlation is not as strong for the interpolation between 1 and 0 which we attribute to the difficulty in estimating $\alpha$ reliably~\citep{clauset2009power} (the reliability of the power-law assumption is explored in depth in \S~\ref{app:reliability_of_alpha}, which shows inconclusive evidence for fitting a power law to the eigenvalues and when adjusting for under estimated tail samples we observe a strong monatomic increasing relationship (Figure~\ref{fig:model_init_vs_model_end_fixed}). Notably, well performing models exist inside and outside of the good alpha range at the end of training (see Figure~\ref{fig:model_alpha_test_acc_through_training}).
These results can also be interpreted as providing additional credence to the fact that SGD performs heavy-tail regularisation on the singular values throughout training, which can also explain why we observe $\alpha\in[2,6]$ in well-trained models. Thus, we propose that if the average alpha of well-trained models tends to lie in the `good' range $[2,6]$ it is as a consequence of our standard initialisation procedures and regularisation from SGD, not because this metric is indicative of generalisation.

\subsection{Generalisation With Fixed Alpha}
\label{sec:fixed_alpha}
\begin{table}[H]
\caption{Resulting test performance of models trained with fixed Alpha ($\alpha$) values. Mean and standard error of the mean derived from 10 trained models trained on MNIST.}
\label{tab:training-with-fixed-alpha}
\resizebox{\textwidth}{!}{%
\begin{tabular}{|c|c|c|c|c|c|c|c|}
\hline
\textbf{Fix Alpha Range} & \textbf{FC1}      & \textbf{FC2}      & \textbf{FC3}      & \textbf{FC4}      & \textbf{FC5}      & \textbf{FC6}      & \textbf{Test Accuracy} \\ \hline
$2 \leq \alpha \leq6$ & 5.4799 $\pm$ 0.1312  & 5.0627 $\pm$ 0.1209  & 5.1361 $\pm$ 0.1836  & 4.9623 $\pm$ 0.1957  & 5.209 $\pm$ 0.194    & 5.4213 $\pm$ 0.1224  & 90.178 $\pm$ 0.0214       \\ \hline
$10 \leq \alpha \leq12$ & 10.7025 $\pm$ 0.1442 & 11.0621 $\pm$ 0.2103 & 10.8441 $\pm$ 0.1443 & 10.8218 $\pm$ 0.1284 & 11.0165 $\pm$ 0.138  & 11.2001 $\pm$ 0.2024 & 90.223 $\pm$ 0.0237       \\ \hline
$18 \leq \alpha \leq20$ & 19.0752 $\pm$ 0.1244 & 17.4638 $\pm$ 1.2448 & 18.8759 $\pm$ 0.1885 & 18.7325 $\pm$ 0.1573 & 18.6775 $\pm$ 0.2023 & 18.8356 $\pm$ 0.1289 & 90.219 $\pm$ 0.0253       \\ \hline
\end{tabular}%
}
\end{table}
To further highlight that that the average alpha is not causally related to generalisation, we train models with arbitrary fixed $\alpha$ values between $2 \leq \alpha \leq 6$,\space $10 \leq \alpha \leq 12$, and $18 \leq \alpha \leq 20$ using the SVD-based methodology with fixed-alpha spectra $\Sigma_\alpha$ described at the end of \S~\ref{sec:setup}. The results in Table~\ref{tab:training-with-fixed-alpha} show that the models achieve no statistically meaningful difference in performance regardless of the fixed $\alpha$. All models, regardless of fixed $\alpha$, achieve similar performance. Therefore,\space $\alpha$ between 2 and 6 is not causally related or required for good performance. It should be noted that the model does not achieve the same performance as in \S~\ref{sec:small-scale-analysis}; however, this can be attributed to the fixed singular values $\Sigma_{\alpha}$, which have a strong regularisation effect. Furthermore, this shows that despite models generally falling within good alpha values during training, this is not a prerequisite of generalisation and is a quirk of the current optimisation process afforded by SGD and similar variants. In \S~\ref{sec:exp-spectral-vs-training} we further this analysis by conducting a full ablation study of the interactions of the SVD factors during training and perform controlled interventions to understand the role the singular value spectra has on learning which further confirm the lack of causality for optimal $\alpha$ values and generalisation.

\section{Conclusion}\label{sec:conclusion}
We showed that RMT-based metrics cannot predict or explain model generalisation, nor the robust vs critical layer phenomenon, nor the re-initialisation vs re-randomisation phenomenon. We designed a range of experiments covering all aspects of the problem -- individual layers \S~\ref{sec:small-scale-analysis},~\ref{sec:large-scale} versus entire model \S~\ref{sec:alpha_init_training},~\ref{sec:fixed_alpha}, initialisation \S~\ref{sec:alpha_init_training}, training \S~\ref{sec:alpha_init_training},~\ref{sec:fixed_alpha}, model performance \S~\ref{sec:results}, eigenvalue power-law fitting \S~\ref{app:reliability_of_alpha} -- and a wide set of architectures, problems and scales -- MNIST, ImageNet, LLMs -- that all support our conclusion that
\vspace{-1.5mm}
\begin{center}
    `good model $\nRightarrow$ good data-free metrics' and `good data-free metrics $\nRightarrow$ good model'
\end{center} 
\vspace{-1.5mm}
based on the fact that data-free metrics are affected in completely arbitrary ways by functionality-preserving reparametrisations of models (concretely given by the re-randomisations of robust layers). We have provided an alternative explanation for the fact that data-free metrics like $\alpha$ tend to fall in certain ranges for well-trained models.

Uncovering or disproving causality relations is a core scientific endeavour, often with concrete engineering consequences. Our results advocate for a reappraisal of the way that we approach interpretability of neural networks. Instead of using metrics with spurious predictive capacity, we argue that there is a need for an in-depth exploration of data-free methods that can reliably disambiguate between the robust and critical layer phenomenon, i.e.\ display some level of invariance under functionality-preserving reparametrisations. Additionally, we advocate for a focus on methods that consider more than just weight distributions of models and instead seek metrics which further understand the more nuanced interplay between between weights and data. We argue that data-free metrics fail because they do not capture the nuanced interplay between both the data passed through the model and the weights that process it. Concretely our recommendations are as follows: \textbf{1} Avoid the use of data-free that metrics as they only have spurious relations with model performance, \textbf{2} Ensure that novel data-free metrics are able to meaningfully capture performance differences offered by re-initialisation and re-randomisation, \textbf{3} In lieu of such data-free metrics, pursue the understanding of neural networks with metrics that consider the crucial interplay between weights and data and \textbf{4} Prioritise data-based approaches for interpretability based on their ability to distinguish between re-initialisation and re-randomisation.

\bibliographystyle{plainurl}
\bibliography{neurips}
\newpage
\appendix
\begin{ack}
\textbf{1}: Calculations were performed using the Sulis Tier 2 HPC platform hosted by the Scientific Computing Research Technology Platform at the University of Warwick. Sulis is funded by EPSRC Grant EP/T022108/1 and the HPC Midlands+ consortium. \textbf{2}:  This research utilised Queen Mary's Apocrita HPC facility, supported by QMUL Research-IT~\citep{qmul_hpc}. \textbf{3}: Calculations were performed using the King’s College London HPC \citep{kingshpc}. \textbf{4}: This work was supported by UK Research and Innovation grant number EP/S023356/1, in the UKRI Centre for Doctoral Training in Safe and Trusted Artificial Intelligence\footnote{UKRI Centre for Doctoral Training in Safe and Trusted Artificial: \url{www.safeandtrustedai.org}.}.
\end{ack}
\section{Limitations}
\label{app:Limitations}
In the main body of the paper we show across modalities, learning tasks and parameter scales that the alpha metric offered by Random Matrix Theory is not reparametrisation invariant under the re-randomisation and re-initialisation setting offered by critical layer theory. In \S~\ref{app:data_free_metrics}, we extend our analysis to eight other data-free metrics across the same scales. While this is an extensive assessment of data-based metrics, we do not study all available data-free metrics. Given that data-free metrics are typically derived from similar matrix properties, one would expect that our findings should extend to all or most metrics built from neural network weights. However, caution would recommend that this be empirically confirmed for all data-free metrics. For the metrics we explore in the main body and in the appendix, we find no data-free metric that is reparametrisation invariant. In the paper we study two prominent modalities, vision and language, but do not explore data domains such as audio, as a result a natural extension of our experiments would be to confirm our findings in this data domain. From the results presented, we believe that the same findings would hold. In \S~\ref{sec:data-based-metrics} we show that data-based metrics are reparametrisation invariant, and thus we argue that they should be prioritised in future studies, but we only survey three data-based metrics (Activation Disagreement, Jenson-Shannon Divergence, and Linear CKA). As a result, future work could focus on which data-based metrics provide the strongest reparametrisation invariance - once again, we argue that this is more of an evaluative task and is outside of the scope of the core insights provided in this paper. 

\section{Pythia-1B Re-initialisation and Re-randomisation}
\label{app:pythia-1B}
\begin{wraptable}{r}{0.45\textwidth}
\centering
\caption{Model test performance on WikiText-103 and mean layer Alpha ($\alpha)$ value.}
\label{tab:pythia1b}
\resizebox{0.45\textwidth}{!}{
\begin{tabular}{|c|c|c|c|c|}
\hline
\textbf{Model} & \textbf{Params} & \textbf{Loss} & \textbf{Acc} & \textbf{Mean $\boldsymbol{\alpha}$} \\ \hline
Pythia         & 1B              & 2.576         & 47.680        & 4.801$\pm$0.202            \\ \hline
\end{tabular}}
\end{wraptable}
\textbf{Pythia-1B} To conduct re-initialisation and re-randomisation experiments at large scale we use the Pythia-1B model~\cite{pythia} whose initialisations are publicly available. To gauge the impact of re-initialisation and re-randomisation on the model we use the same WikiText-103 dataset that was used in the main body for the GPT experiments. Table~\ref{tab:pythia1b} presents the unmodified model accuracy on the dataset and the alpha fit of the the layer weights within the training. Here, the model can be described as well-trained given that the mean $\alpha$ value is within the optimal range at the end of training.

\begin{table}[H]
\centering
\caption{The impact of re-initialisation across a range of layers within the Pythia-1B model, alongside the correlation statistics of $\alpha$ and test accuracy.}
\label{tab:Pythia-1B_re_init}
\resizebox{\linewidth}{!}{
\begin{tabular}{|c|c|c|c|c|c|c|c|}
\hline
\textbf{Layer Type} & \textbf{Accuracy} & \textbf{Alpha ($\boldsymbol{\alpha}$)} & \textbf{\begin{tabular}[c]{@{}c@{}}\textbf{\% in Optimal $\boldsymbol{\alpha}$ Range}\end{tabular}} & \textbf{$\boldsymbol{\rho}$} & \textbf{$\boldsymbol{R^2}$} & \textbf{RSME} & \textbf{K-$\boldsymbol{\tau}$} \\ \hline
Attention QKV       & 42.187$\pm$0.896      & 29.157$\pm$4.429      & 0.000                                                                       & 0.388           & 0.103            & 3.395         & 0.283          \\ \hline
Attention Out       & 42.681$\pm$1.743      & 21.945$\pm$2.703      & 0.000                                                                       & 0.174           & 0.119            & 6.541         & 0.117          \\ \hline
MLP 1               & 33.264$\pm$2.996      & 25.337$\pm$3.831      & 0.000                                                                       & -0.106          & 0.045            & 11.710         & -0.083         \\ \hline
MLP 2               & 41.093$\pm$2.729      & 28.064$\pm$4.798      & 0.000                                                                       & -0.247          & 0.563            & 7.216         & -0.200           \\ \hline
\end{tabular}}
\end{table}
In Table~\ref{tab:Pythia-1B_re_init} we present the impact of re-initialisation of a range of layers on the Pythia-1B model. For layers Attention QKV, Attention Out, MLP1 and MLP2 the accuracy is typically preserved under re-initialisation. Alongside this accuracy preservation, we record alpha values that are far outside of the optimal range. Moreover, when analysing the correlation between $\alpha$
 values and accuracy, we see across correlation metrics that $\alpha$ has weak predictive capacity. These results agree with the existing analysis for the MNIST ReLU FCN 5x512 architecture explored in the main body. 
\begin{table}[H]
\centering
\caption{The impact of re-randomisation across a range of layers within the Pythia-1B model, alongside the correlation statistics of $\alpha$ and test accuracy.}
\label{tab:Pythia-1B_re_rand}
\resizebox{\textwidth}{!}{%
\begin{tabular}{|c|c|c|c|c|c|c|c|}
\hline
\textbf{Layer Type} & \textbf{Accuracy}   & \textbf{Alpha ($\boldsymbol{\alpha}$)} & \textbf{\% in Optimal $\boldsymbol{\alpha}$ Range} & \textbf{$\boldsymbol{\rho}$} & \textbf{$\boldsymbol{R^2}$} & \textbf{RSME} & \textbf{K-$\boldsymbol{\tau}$} \\ \hline
Attention QKV       & $41.581 \pm  1.193$ & $33.577 \pm  6.225$       & 0.000                            & -0.059          & 0.003          & 4.767         & -0.067            \\ \hline
Attention Out       & $42.437 \pm  1.817$ & $23.361 \pm  2.651$       & 0.000                            & -0.671          & 0.148          & 6.710          & -0.450             \\ \hline
MLP1                & $32.241 \pm  3.140$ & $29.251 \pm  4.495$       & 0.000                            & 0.124           & 0.010           & 12.498        & 0.050              \\ \hline
MLP2                & $40.954 \pm  2.727$ & $22.579 \pm  1.961$       & 0.000                            & -0.238          & 0.059          & 10.58         & -0.150             \\ \hline
\end{tabular}%
}
\end{table}

\begin{figure}[H]
    \centering
    \includegraphics[width=\linewidth]{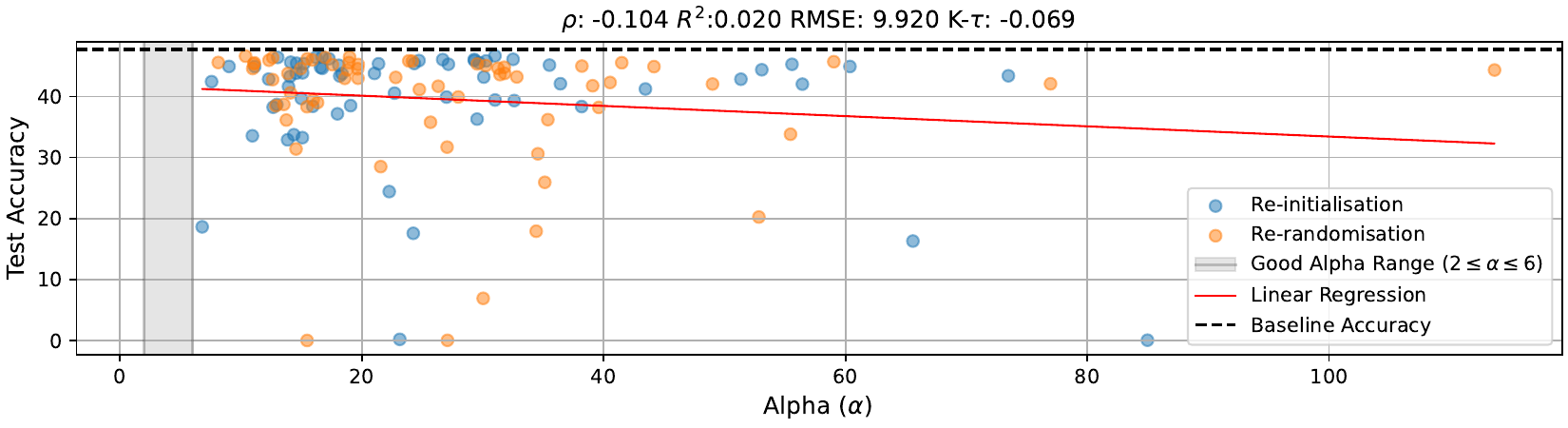}
    \caption{Pythia-1B layer re-initialisation (\textcolor{blue}{\textbf{$\bullet$}}) and re-randomisation (\textcolor{orange}{\textbf{$\bullet$}}) test accuracy vs $\alpha$.}
    \label{fig:pythia-alpha}
\end{figure}

When considering the case of re-randomisation Table~\ref{tab:Pythia-1B_re_rand} shows that accuracy is preserved less than in the case of re-randomisation. Similar to the results for re-initialisation we record alpha values that are far outside of the optimal range of alpha values. Given that for both re-randomisation and re-initialisation that we record alpha values that are outside of the optimal value range, yet there exists a meaningful performance impact between these different conditions highlights that alpha is not an effective metric for disambiguating these important differences between these reparametrisations, see Figure \ref{fig:pythia-alpha} for the global relationship between layer re-initialisation and re-randomisation test accuracy against $\alpha$.

Overall, the results of re-initialisation and re-randomisation at the 1B parameter scale further highlights the inadequacies of $\alpha$ to meaningfully correspond to generalisation under functionally preserving reparametrisations. 

\section{Reparametrisation and Function Preservation}
\label{app:reparam-func-pres}
In this paper we suggest that preservation of model accuracy after layer reparametrisation (re-initialisation or re-randomisation) implies preservation of the function encoded by that model. In the main paper, we show that when comparing the original model to the re-parametrised model across a range of functional similarity and representational similarity metrics~\citep{klabunde2305similarity} accuracy is preserved when the predictive structure of the original model is also preserved. 

Below we provide the corresponding functional/representational similarity analysis for the larger scale models presented in the paper. Overall, the correlation Tables~\ref{tab:model_functional_correlation} and~\ref{tab:Pythia_1B_functional_correlation}, show that functional/representational similarity holds a strong correlation with the change in accuracy observed. Furthermore, this provides strong evidence that when accuracy is preserved, the reparametrisations offered by re-randomisation and re-initialisation preserve the original function of the model.

\begin{table}[H]
\centering
\caption{Correlation between the change in accuracy due to layer re-randomisation and the functional preservation of the model.}
\label{tab:model_functional_correlation}
    \begin{subtable}[b]{\textwidth}
        \caption{ResNet34}
        \centering
        \begin{tabular}{|c|c|c|c|c|}
            \hline
            \textbf{Metric}              & \textbf{$\boldsymbol{\rho}$} & \textbf{$\boldsymbol{R^2}$}  & \textbf{RMSE} & \textbf{K-$\boldsymbol{\tau}$} \\ \hline
            Prediction Similarity       & 1.000               & 0.999          & 0.005         & 0.989             \\ \hline
            Jensen-Shannon Distance     & -0.976          & 0.965          & 0.029         & -0.879            \\ \hline
            Procrustes Distance         & -0.753          & 0.722          & 0.121         & -0.659            \\ \hline
            Kullback-Leibler Divergence & -0.940           & 0.857          & 0.851         & -0.814            \\ \hline
        \end{tabular}%
    \end{subtable}
\end{table}

\begin{table}[H]
    \ContinuedFloat
    \centering
    
    \begin{subtable}[b]{\textwidth}
        \caption{ViT}
        \centering
            \begin{tabular}{|c|c|c|c|c|}
            \hline
            \textbf{Metric}              & \textbf{$\boldsymbol{\rho}$} & \textbf{$\boldsymbol{R^2}$}  & \textbf{RMSE} & \textbf{K-$\boldsymbol{\tau}$} \\ \hline
            Prediction Similarity       & 0.935           & 0.995          & 0.021         & 0.824             \\ \hline
            Jensen-Shannon Distance     & -0.872          & 0.866          & 0.087         & -0.752            \\ \hline
            Procrustes Distance         & -0.923          & 0.858          & 0.099         & -0.771            \\ \hline
            Kullback-Leibler Divergence & -0.877          & 0.970           & 0.342         & -0.759            \\ \hline
            \end{tabular}%
    \end{subtable}

    \begin{subtable}[b]{\textwidth}
        \caption{GPT2}
        \centering
            \begin{tabular}{|c|c|c|c|c|}
            \hline
            \textbf{Metric}        & \textbf{$\boldsymbol{\rho}$} & \textbf{$\boldsymbol{R^2}$}  & \textbf{RMSE} & \textbf{K-$\boldsymbol{\tau}$} \\ \hline
            Prediction Similarity & 0.747           & 0.978          & 0.030          & 0.579             \\ \hline
            \end{tabular}%
    \end{subtable}
    \begin{subtable}[b]{\textwidth}
        \caption{GPT2-Large}
        \centering
            \begin{tabular}{|c|c|c|c|c|}
            \hline
            \textbf{Metric}        & \textbf{$\boldsymbol{\rho}$} & \textbf{$\boldsymbol{R^2}$}  & \textbf{RMSE} & \textbf{K-$\boldsymbol{\tau}$} \\ \hline
            Prediction Similarity &   0.842 &   0.949 &   0.030 &      0.655       \\ \hline
            \end{tabular}%
    \end{subtable}
\label{tab:my-table}
\end{table}

\begin{table}[H]
\centering
\caption{Correlation between the change in accuracy due to re-initialisation and re-randomisation and the functional preservation of Pythia-1B.}
\label{tab:Pythia_1B_functional_correlation}
\begin{tabular}{|c|c|c|c|c|}
\hline
\textbf{Metric} & \textbf{$\boldsymbol{\rho}$} & \textbf{$\boldsymbol{R^2}$} & \textbf{RMSE} & \textbf{K-$\boldsymbol{\tau}$} \\ \hline
Prediction Sim  & 0.991           & 0.968            & 0.036         & 0.933          \\ \hline
\end{tabular}
\end{table}

\section{Initialisation Scheme and Alpha (\texorpdfstring{$\alpha$}{α})}
\label{app:init_method_and_aplha}
In the main body of the paper, we use the default initialisation scheme offered PyTorch. In these results and Figure~\ref{fig:alpha-init-dist} in particular, we observe that there is a non-negligible probability of a randomly initialised layer having an $\alpha$ value in the `good' range $[2,6]$. In this section, we identify the probability of such ranges emerging at initialisation across a range of different layer types, namely:  Linear Layer 512x512, Resnet34 Layer2 Conv\_2 (Conv2d(128, 128, kernel\_size=(3, 3), stride=(1, 1), padding=(1, 1)) and ViT Base 16 Out Projection. 
\\
\\
Additionally, as many different initialisation schemes are offered by PyTorch such as Xavier (normal ($\mathcal{X}_{n}$), uniform ($\mathcal{X}_{u}$)) \cite{glorot_init} and Kamming (normal ($\mathcal{K}_{n}$) , uniform ($\mathcal{K}_{u}$))\cite{kaiming_init}  we have reproduced the analysis in each of the following tables to also capture these initialisations. The following tables present the descriptive statistics of each initialisation type and their relationships with alpha for the layers named above. 

\begin{table}[H]
\centering
\caption{Distribution statistics of $\alpha$ across initialisation schemes and layer types.}
\label{tab:distrubtion_alpha_init_schema}
\begin{subtable}[htbp]{\linewidth}
    \centering
    \caption{Linear Layer $512\times512$.}
    \label{tab:layer_512_512}
    \resizebox{\linewidth}{!}{
        \begin{tabular}{|c|c|c|c|c|c|c|c|c|}
            \hline
            \textbf{\begin{tabular}[c]{@{}c@{}}Initialisation \\ Scheme\end{tabular}} & \textbf{$\boldsymbol{\bar{x}}$} & \textbf{$\boldsymbol{\Tilde{x}}$} & \textbf{$\boldsymbol{\sigma}$} & \textbf{Skew} & \textbf{Kurtosis} & \textbf{Min} & \textbf{Max} & \textbf{\begin{tabular}[c]{@{}c@{}}\% in  Optimal \\ $\boldsymbol{\alpha}$ Range\end{tabular}} \\ \hline
            PyTorch default                                                                 & 10.985      & 9.019       & 6.390             & 2.088         & 5.992             & 3.500        & 56.640       & 16.640                                                                  \\ \hline
            $\mathcal{X}_{u}$                                                       & 11.112      & 8.993       & 6.587             & 2.017         & 5.360             & 3.398        & 64.359       & 16.910                                                                  \\ \hline
            $\mathcal{X}_{n}$                                                       & 10.960      & 9.001       & 6.448             & 2.083         & 5.938             & 3.446        & 63.332       & 17.510                                                                  \\ \hline
            $\mathcal{K}_{u}$                                                       & 10.877      & 8.948       & 6.340             & 2.191         & 6.950             & 3.353        & 68.180       & 16.700                                                                  \\ \hline
            $\mathcal{K}_{n}$                                                       & 10.993      & 9.040       & 6.524             & 2.216         & 7.038             & 3.403        & 68.135       & 17.120                                                                  \\ \hline
        \end{tabular}
    }
\end{subtable}

\centering
\begin{subtable}[htbp]{\linewidth}
    \centering
    \caption{ResNet34 Layer2 Conv$\_$2 (Conv2d(128, 128,
    kernel$\_$ size=(3, 3), stride=(1, 1), padding=(1, 1)).}
    \label{tab:resnet_conv2}
    \resizebox{\linewidth}{!}{
        \begin{tabular}{|c|c|c|c|c|c|c|c|c|}
            \hline
            \textbf{\begin{tabular}[c]{@{}c@{}}Initialisation \\ Scheme\end{tabular}} & \textbf{$\boldsymbol{\bar{x}}$} & \textbf{$\boldsymbol{\Tilde{x}}$} & \textbf{$\boldsymbol{\sigma}$} & \textbf{Skew} & \textbf{Kurtosis} & \textbf{Min} & \textbf{Max} & \textbf{\begin{tabular}[c]{@{}c@{}}\% in  Optimal \\ $\boldsymbol{\alpha}$ Range\end{tabular}} \\ \hline
            PyTorch default                                                         & 18.126             & 13.518               & 14.302            & 2.835         & 14.455            & 3.594        & 211.387      & 5.990                                                                   \\ \hline
            $\mathcal{X}_{u}$                                                       & 18.341             & 13.940               & 14.146            & 2.748         & 14.229            & 3.494        & 191.422      & 6.210                                                                   \\ \hline
            $\mathcal{X}_{n}$                                                       & 17.729             & 13.540               & 13.747            & 3.585         & 32.047            & 3.405        & 288.763      & 6.510                                                                   \\ \hline
            $\mathcal{K}_{u}$                                                       & 18.186             & 13.791               & 13.986            & 2.790         & 15.779            & 3.367        & 221.806      & 6.230                                                                   \\ \hline
            $\mathcal{K}_{n}$                                                       & 17.919             & 13.750               & 13.271            & 2.390         & 9.203             & 3.583        & 152.902      & 6.350                                                                   \\ \hline
        \end{tabular}
    }
\end{subtable}
\end{table}

\begin{table}[H]
\ContinuedFloat
\begin{subtable}[htbp]{\linewidth}
    \centering
    \caption{ViT Base 16 Out Projection (Linear Layer $768\times768$).}
    \label{tab:ViT_out}
    \resizebox{\linewidth}{!}{
        \begin{tabular}{|c|c|c|c|c|c|c|c|c|}
            \hline
            \textbf{\begin{tabular}[c]{@{}c@{}}Initialisation \\ Scheme\end{tabular}} & \textbf{$\boldsymbol{\bar{x}}$} & \textbf{$\boldsymbol{\Tilde{x}}$} & \textbf{$\boldsymbol{\sigma}$} & \textbf{Skew} & \textbf{Kurtosis} & \textbf{Min} & \textbf{Max} & \textbf{\begin{tabular}[c]{@{}c@{}}\% in  Optimal \\ $\boldsymbol{\alpha}$ Range\end{tabular}} \\ \hline
            PyTorch default                                                         & 13.002             & 10.587               & 7.953             & 2.183         & 6.524             & 3.873        & 76.666       & 8.020                                                                   \\ \hline
            $\mathcal{X}_{u}$                                                       & 13.197             & 10.667               & 8.143             & 2.161         & 6.974             & 3.836        & 101.310      & 8.310                                                                   \\ \hline
            $\mathcal{X}_{n}$                                                       & 12.977             & 10.531               & 7.969             & 2.200         & 6.844             & 3.990        & 74.792       & 8.190                                                                   \\ \hline
            $\mathcal{K}_{u}$                                                       & 13.183             & 10.664               & 8.071             & 2.099         & 5.917             & 4.001        & 68.743       & 7.840                                                                   \\ \hline
            $\mathcal{K}_{n}$                                                       & 13.185             & 10.618               & 8.204             & 2.191         & 6.667             & 3.857        & 88.848       & 7.740                                                                   \\ \hline
        \end{tabular}
    }
\end{subtable}
\end{table}

Table~\ref{tab:distrubtion_alpha_init_schema} presents summary statistics for $\alpha$ across different initialisation schemes for different layer types of models. These results  complement that of those presented in Figure~\ref{fig:alpha-init-dist} by showing that initialised layers can fall within optimal value range of $\alpha$ with a somewhat large non-zero probability. Given that HTSR argues that such $\alpha$ values should only emerge at the end of training and quantify a good layer, this evidence further undermines the ability of $\alpha$ to effectively capture generalisation in neural networks.

\section{Alpha (\texorpdfstring{$\alpha$}{α}) Threshold-Prediction Tests}

The $\alpha$ metric can be considered a threshold based classification metric, where $2\leq\alpha\leq6$ denotes a `good' layer or model with high accuracy and $\alpha<2$ or $\alpha>6$ is considered a `bad' layer or model. With this perspective we analyse the performance of $\alpha$ of predicting a good layer, after reparametrisation offered by re-reinitialisation and re-randomisation. A `good' layer is considered to be a layer that results in the model being within 10\% of the original test accuracy after the reparametrisation and has an $\alpha$ within $2\leq\alpha\leq6$. In Table \ref{tab:alpha-evaluation} we report the global (for all layers) AUC and the F1-score, Precision, True Positive Rate (TPR), False Positive Rate (FPR) of $\alpha$ predicting a good layer; to clarify, a good layer corresponds to the resulting model being within 10\% of the original test accuracy after the reparametrisation. For the Pythia 1B model, we are not able to provide the AUC score nor the true positive rate score as none of the reparametrisations resulted in layers within the 2 and 6 alpha range. 

\begin{table}[H]
\centering
\caption{Effectiveness of $\alpha$ at predicting a good layer $2\leq\alpha\leq6$ where model performance is within 10\% of the original test accuracy after layer reparametrisation.}
\label{tab:alpha-evaluation}
\resizebox{\textwidth}{!}{%
\begin{tabular}{|l|c|c|c|c|c|}
\hline
\textbf{Model} & \textbf{AUC} & \textbf{F1-Score} & \textbf{Precision} & \textbf{True Positive Rate} & \textbf{False Positive Rate} \\ \hline
MLP            & 0.520        & 0.242             & 0.167              & 0.439        & 0.387        \\ \hline
ResNet34       & 0.449        & 0.000             & 0.000              & 0.000        & 0.020        \\ \hline
ViT            & 0.686        & 0.056             & 0.029              & 1.000        & 0.836        \\ \hline
GPT2           & 0.442        & 0.022             & 0.011              & 0.714        & 0.901        \\ \hline
GPT2-Large     & 0.603        & 0.010             & 0.005              & 0.875        & 0.973        \\ \hline
Pythia-1B      & nan          & 0.000             & 0.000              & nan          & 0.797        \\ \hline
\end{tabular}%
}
\end{table}

For all models evaluated, $\alpha$ achieves a poor AUC. For the ResNet34 and GPT, this is worse than random; the MLP is slightly better than random, and for the GPT-Large and ViT it is better than random; however, still poor. Furthermore, the F1-Score and Precision are very low, with the highest F1-score and Precision of 0.242 and 0.167, respectively, in the MLP case. With regard to the true positive rate (TPR) and false positive rate (FPR) of the optimal $\alpha$ range, a high TPR corresponds with a high FPR, showing the metric is poor at effectively identifying good layers. In the case of ResNet, the $\alpha$ metric fails to identify any of the layers that correspond to a model maintaining performance, whilst flagging a small number that do not maintain performance as good layers. Overall, these results strongly corroborate the main findings in the paper.

\newpage
\section{Alpha (\texorpdfstring{$\alpha$}{α}) Analysis Stratified Over Layer Type}
\label{app:stratified_analysis}
Table \ref{tab:stratified-layers}, presents the stratified analysis of the model layers by type, for the ResNet34 and ViT on ImageNet and the GPT2 and GPT2-Large on WikiText103. In the tables there is no consistent trend across the layer types, with some layers reporting a positive relationship between test accuracy and $\alpha$ and others a negative relationship. Which strongly supports the main findings in the paper that $\alpha$ values are not a reliable indicator of performance and generalisation.
\begin{table}[H]
\centering
\caption{Stratified layer re-randomisation Test Accuracy vs $\alpha$}
\label{tab:stratified-layers}
\begin{subtable}[H]{\linewidth}
\centering
\caption{ResNet34 on ImageNet}
\label{tab:resnet34-stratified}
\resizebox{\textwidth}{!}{%
\begin{tabular}{|l|c|c|c|c|c|c|c|}
\hline
\textbf{Layer Type} & \textbf{Accuracy}               & \textbf{Alpha ($\alpha$)}      & \textbf{\% in  Optimal $\boldsymbol{\alpha}$ Range}  & \textbf{$\boldsymbol{\rho}$} & \textbf{$\boldsymbol{R^2}$}  & \textbf{RMSE} & \textbf{K-$\boldsymbol{\tau}$} \\ \hline
Input Conv2d        & 0.395$\pm$ 0.029  & 14.381$\pm$ 1.817 & 0.000                       & 0.176           & 0.008          & 0.091         & 0.156             \\ \hline
Conv2d\_1           & 2.461$\pm$ 0.260   & 22.628$\pm$ 1.525 & 3.750                       & -0.178          & 0.033          & 3.232         & -0.117            \\ \hline
Conv2d\_2           & 36.35$\pm$ 1.610   & 23.882$\pm$ 1.637 & 3.750                       & -0.140           & 0.052          & 19.823        & -0.096            \\ \hline
Downsample          & 15.793$\pm$ 2.474 & 6.788$\pm$ 0.604  & 43.333                      & 0.634           & 0.335          & 11.052        & 0.456             \\ \hline
Output FC           & 0.098$\pm$ 0.009  & 11.829$\pm$ 2.095 & 0.000                       & 0.492           & 0.036          & 0.029         & 0.405             \\ \hline
\end{tabular}%
}
\end{subtable}

\begin{subtable}[H]{\linewidth}
\centering
\caption{ViT on ImageNet}
\label{tab:vit-stratified}
\resizebox{\textwidth}{!}{%
\begin{tabular}{|l|c|c|c|c|c|c|c|}
\hline
\textbf{Layer Type} & \textbf{Accuracy}               & \textbf{Alpha ($\alpha$)}      & \textbf{\% in  Optimal $\boldsymbol{\alpha}$ Range}  & \textbf{$\boldsymbol{\rho}$} & \textbf{$\boldsymbol{R^2}$}  & \textbf{RMSE} & \textbf{K-$\boldsymbol{\tau}$} \\ \hline
patch\_embed.proj   & 0.145$\pm$ 0.008  & 65.281$\pm$ 4.607 & 0.000                       & 0.467           & 0.241          & 0.021         & 0.422             \\ \hline
attn.qkv            & 77.808$\pm$ 0.071 & 17.978$\pm$ 1.218 & 1.667                       & 0.077           & 0.019          & 0.769         & 0.058             \\ \hline
attn.proj           & 77.469$\pm$ 0.223 & 13.345$\pm$ 0.725 & 6.667                       & -0.075          & 0.004          & 2.442         & -0.049            \\ \hline
mlp.fc1             & 52.805$\pm$ 2.806 & 16.678$\pm$ 0.911 & 0.000                       & 0.239           & 0.032          & 30.238        & 0.162             \\ \hline
mlp.fc2             & 71.847$\pm$ 1.902 & 20.088$\pm$ 1.157 & 1.667                       & -0.055          & 0.001          & 20.824        & -0.036            \\ \hline
head                & 0.128$\pm$ 0.013  & 14.194$\pm$ 3.329 & 0.000                       & -0.219          & 0.014          & 0.042         & -0.135            \\ \hline
\end{tabular}%
}
\end{subtable}

\begin{subtable}[H]{\linewidth}

\centering
\caption{GPT2 on WikiText103}
\label{tab:gpt2-stratified}
\resizebox{\textwidth}{!}{%
\begin{tabular}{|l|c|c|c|c|c|c|c|}
\hline
\textbf{Layer Type} & \textbf{Accuracy}               & \textbf{Alpha ($\alpha$)}      & \textbf{\% in  Optimal $\boldsymbol{\alpha}$ Range}  & \textbf{$\boldsymbol{\rho}$} & \textbf{$\boldsymbol{R^2}$}  & \textbf{RMSE} & \textbf{K-$\boldsymbol{\tau}$} \\ \hline
wpe            & 7.336$\pm$ 0.030   & 15.272$\pm$ 2.730  & 0.000                       & -0.127          & 0.148          & 0.087         & -0.067            \\ \hline
attn.c\_attn   & 36.472$\pm$ 0.355 & 16.644$\pm$ 0.926 & 0.000                       & -0.049          & 0.002          & 3.887         & -0.036            \\ \hline
attn.c\_proj   & 35.382$\pm$ 0.739 & 11.705$\pm$ 0.528 & 5.833                       & 0.077           & 0.003          & 8.080          & 0.058             \\ \hline
mlp.c\_fc      & 34.874$\pm$ 0.845 & 18.075$\pm$ 1.039 & 0.000                       & -0.120           & 0.003          & 9.237         & -0.079            \\ \hline
mlp.c\_proj    & 35.430$\pm$ 0.777  & 18.240$\pm$ 0.859  & 0.000                       & 0.146           & 0.000          & 8.512         & 0.098             \\ \hline
\end{tabular}%
}
\end{subtable}
\end{table}

\begin{table}[H]
\centering
\ContinuedFloat
\begin{subtable}[H]{\linewidth}
\centering
\caption{GPT2-Large on WikiText103}
\label{tab:gpt2-large-stratified}
\resizebox{\textwidth}{!}{%
\begin{tabular}{|l|c|c|c|c|c|c|c|}
\hline
\textbf{Layer Type} & \textbf{Accuracy}               & \textbf{Alpha ($\alpha$)}      & \textbf{\% in  Optimal $\boldsymbol{\alpha}$ Range}  & \textbf{$\boldsymbol{\rho}$} & \textbf{$\boldsymbol{R^2}$}  & \textbf{RMSE} & \textbf{K-$\boldsymbol{\tau}$} \\ \hline
wpe                 & 8.905$\pm$ 0.072  & 12.303$\pm$ 1.625 & 10.000                          & -0.818          & 0.734          & 0.117         & -0.644            \\ \hline
attn.c\_attn        & 43.581$\pm$ 0.012 & 22.934$\pm$ 0.838 & 0.000                           & -0.005          & 0.007          & 0.222         & -0.002            \\ \hline
attn.c\_proj        & 43.041$\pm$ 0.203 & 16.266$\pm$ 0.556 & 1.944                       & 0.002           & 0.000              & 3.856         & 0.003             \\ \hline
mlp.c\_fc           & 41.817$\pm$ 0.355 & 24.427$\pm$ 0.888 & 0.000                           & 0.059           & 0.005          & 6.724         & 0.038             \\ \hline
mlp.c\_proj         & 42.413$\pm$ 0.348 & 25.964$\pm$ 0.999 & 0.000                           & -0.058          & 0.000              & 6.599         & -0.039            \\ \hline
\end{tabular}%
}
\end{subtable}
\end{table}

\newpage
\section{Further Analysis on Data-Free Metrics}
\label{app:data_free_metrics}

In this section we extend our analysis to the following data-free metrics under the existing experimental setup. $W$ represents the weight matrix of the layer and $X$ is the empirical spectral density of the correlation matrix,\space $X = W^TW$, such that $p_{emp}(\lambda) \sim \lambda^{-\alpha}$, where $\lambda$ are the eigenvalues of $X$.

 \begin{itemize} 
    \itemsep0em
    \item[] \textbf{Alpha Weighted ($\hat{\alpha}$):} $\alpha \log(\lambda_{max})$, where $\lambda_{max}$ is the max eigenvalue from $X$~\citep{weightwatcher}.
    \item[] \textbf{Log Alpha Norm:}  $\log(||X||^{\alpha}_{\alpha})$, where $||X||^{\alpha}_{\alpha}$  = $\sum\limits_{i}^M\lambda_{i}^{\alpha}$, where $M$ is the rank of $W$~\citep{weightwatcher}.
    \item[] \textbf{MP Soft Rank:} is the ratio between the bulk edge of the $p_{emp}(\lambda)$,\space $\lambda^+$,  and the max eigenvalue,\space $\lambda_{max}$,   $\frac{\lambda^+}{\lambda_{max}}$~\citep{weightwatcher}.
    \item[] \textbf{Spectral Norm}: The max singular value of $W$ denoted as $||W||_{\infty}$.
    \item[] \textbf{Stable Rank:} The ratio of the squared Frobenius Norm and the squared Spectral Norm, denoted as $\frac{||W||^2_F}{||W||^2_2}$~\citep{rudelson2007sampling}.
    \item[] \textbf{Number of Power Law Spikes:} The number of eigenvalues that are outside the bulk region of the ESD, when fit to an Marchenko-Pastur distribution~\citep{weightwatcher}.
    \item[] \textbf{Entropy:}  $\frac{-1}{log(M)} \sum_i p_i \log p_i$, where $M$ is the rank of matrix $W$ and $p_i$ is $\frac{\sigma_i^2} {\sum_i (\sigma_i^2)}$ where $\sigma$ is the singular values of $W$~\citep{weightwatcher}.
    \item[] \textbf{Frobenius Norm}: The sum of the squared singular values of $W$ denoted as $||W||_{F}$.

\end{itemize}

Each metric can be found below with the appropriate subsection that corresponds to our analysis of these data-free metrics.

The correlation between the other data-free metrics formalised in Section~\ref{app:data_free_metrics} and the associated test accuracy when layers undergo re-initialisation (\textcolor{blue}{$\bullet$}) or re-randomisation (\textcolor{orange}{$\bullet$}) is shown in Figure~\ref{fig:genralised_fig}. For all metrics explored we observe approximately zero correlation between the metric values and the test accuracy, showing that other data-free metrics align with the findings presented for the alpha metric presented in the main body of the paper. In \S~\ref{app:ResNet34},\S~\ref{app:ViT} we show this holds for ImageNet models as well as for language models in \S~\ref{app:GPTsmall} and~\ref{app:GPTlarge}.

\subsection{Small Scale Analysis of Data-Free Metrics Under Re-Initialisation and Re-Randomisation}
\label{sec:gen_norms}
Norm-based metrics were originally shown to be too coarse a metric to measure the generalisability of the neural networks in~\cite{create-equal}. Figure~\ref{fig:genralised_fig}  strengthens these findings, highlighting that there is essentially no correlation between the a range of data-free metrics and the test accuracy of a model under re-randomisation and re-initialisation.

\begin{figure}[H]
    \centering
    \begin{subfigure}[htpb]{0.32\linewidth}
        \centering
        \includegraphics[width=\linewidth]{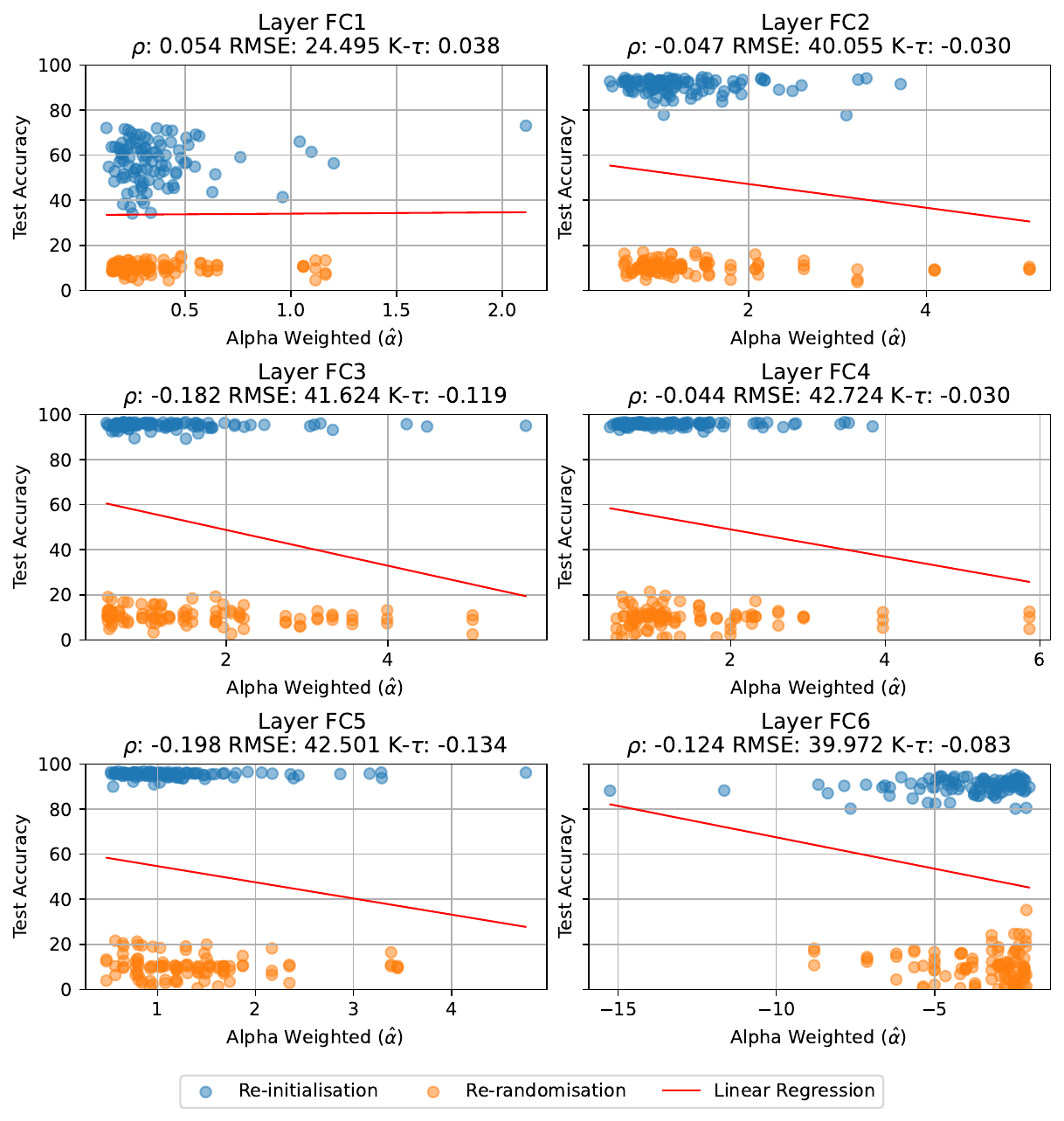}
        \caption{Alpha Weighted}
    \end{subfigure}
    \begin{subfigure}[htpb]{0.32\linewidth}
        \centering
        \includegraphics[width=\linewidth]{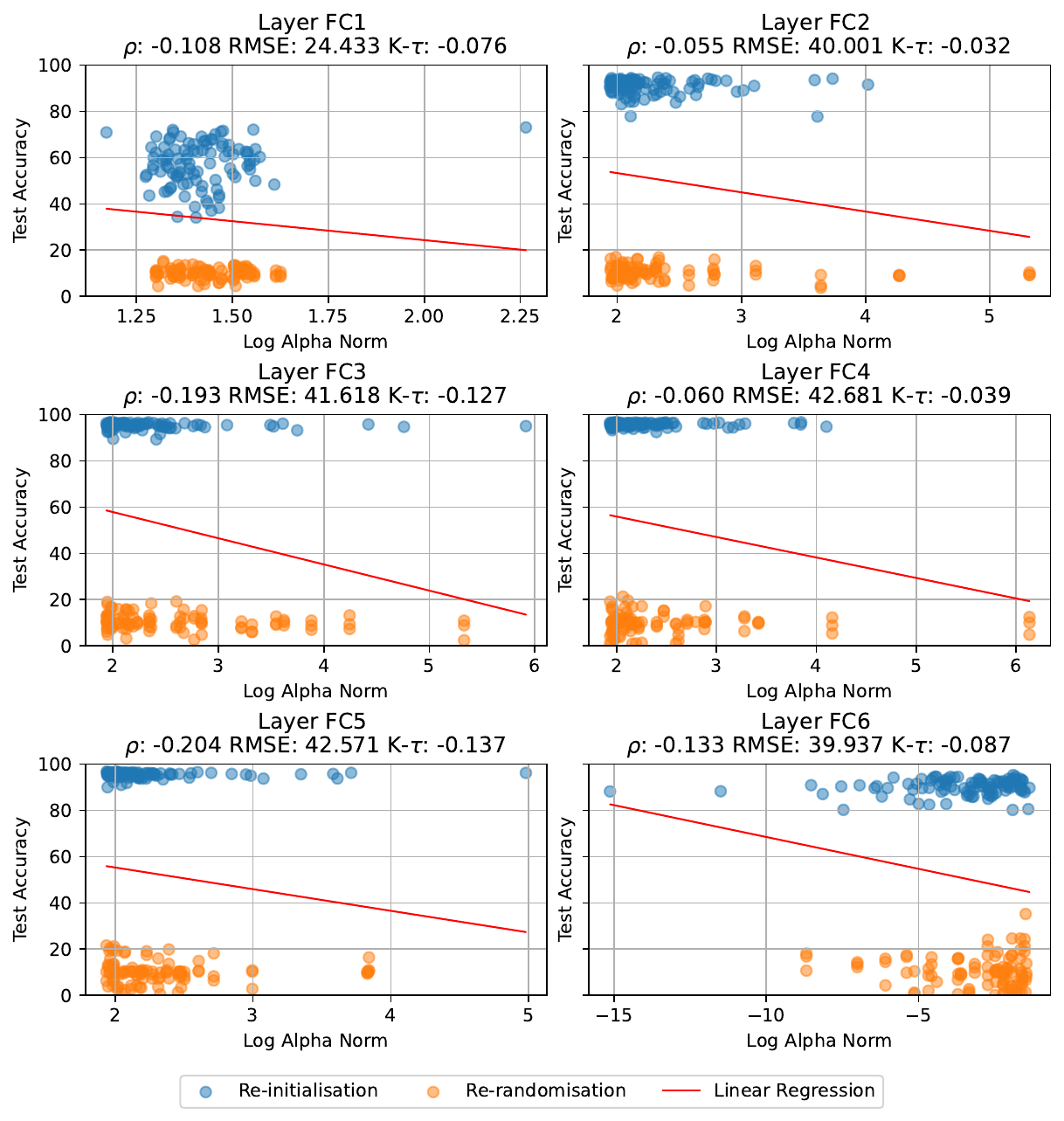}
        \caption{Log Alpha Norm}
    \end{subfigure}
    \begin{subfigure}[htpb]{0.32\linewidth}
        \centering
        \includegraphics[width=\linewidth]{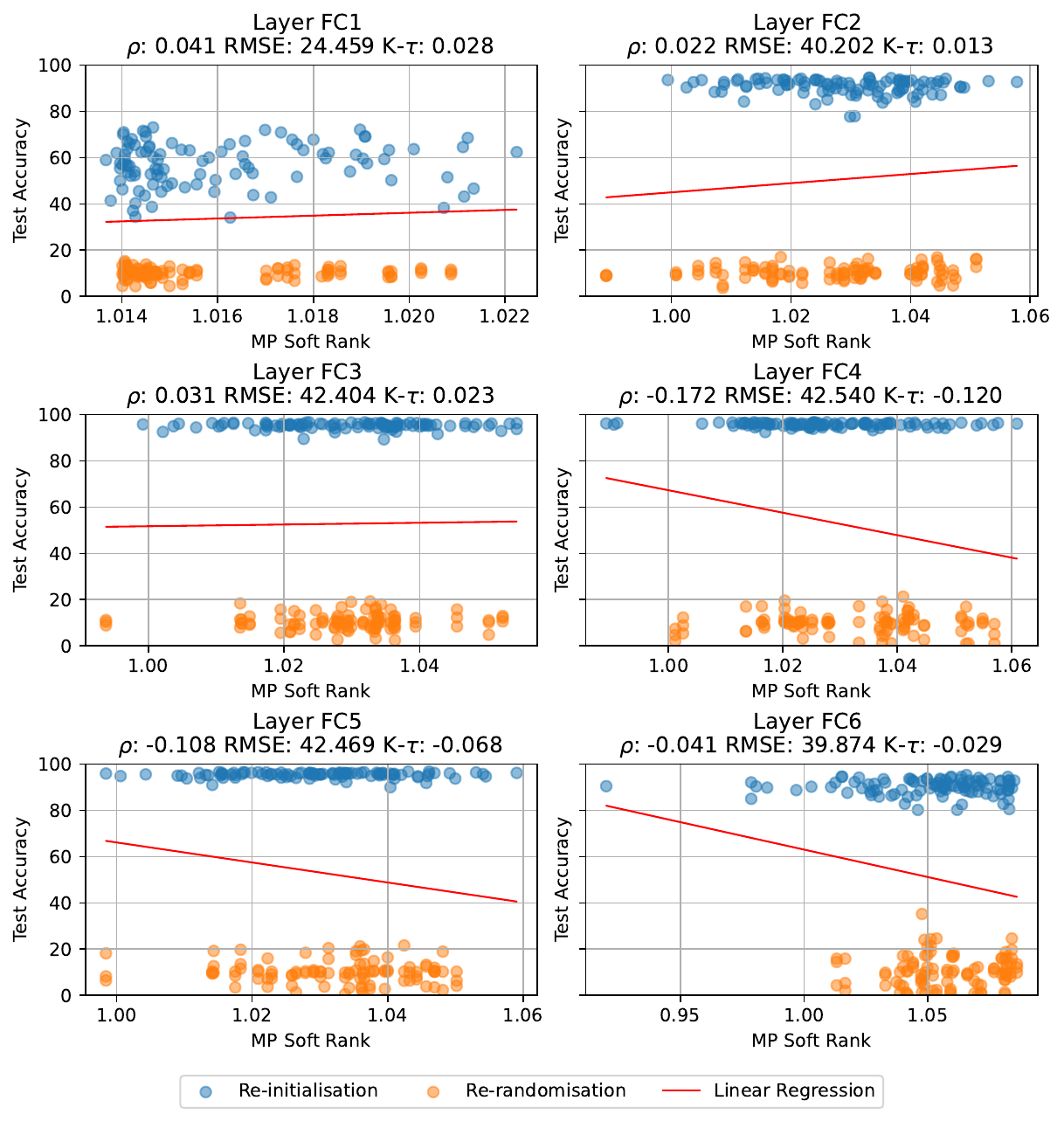}
        \caption{MP Soft Rank}
    \end{subfigure}

    \begin{subfigure}[htpb]{0.32\linewidth}
        \centering
        \includegraphics[width=\linewidth]{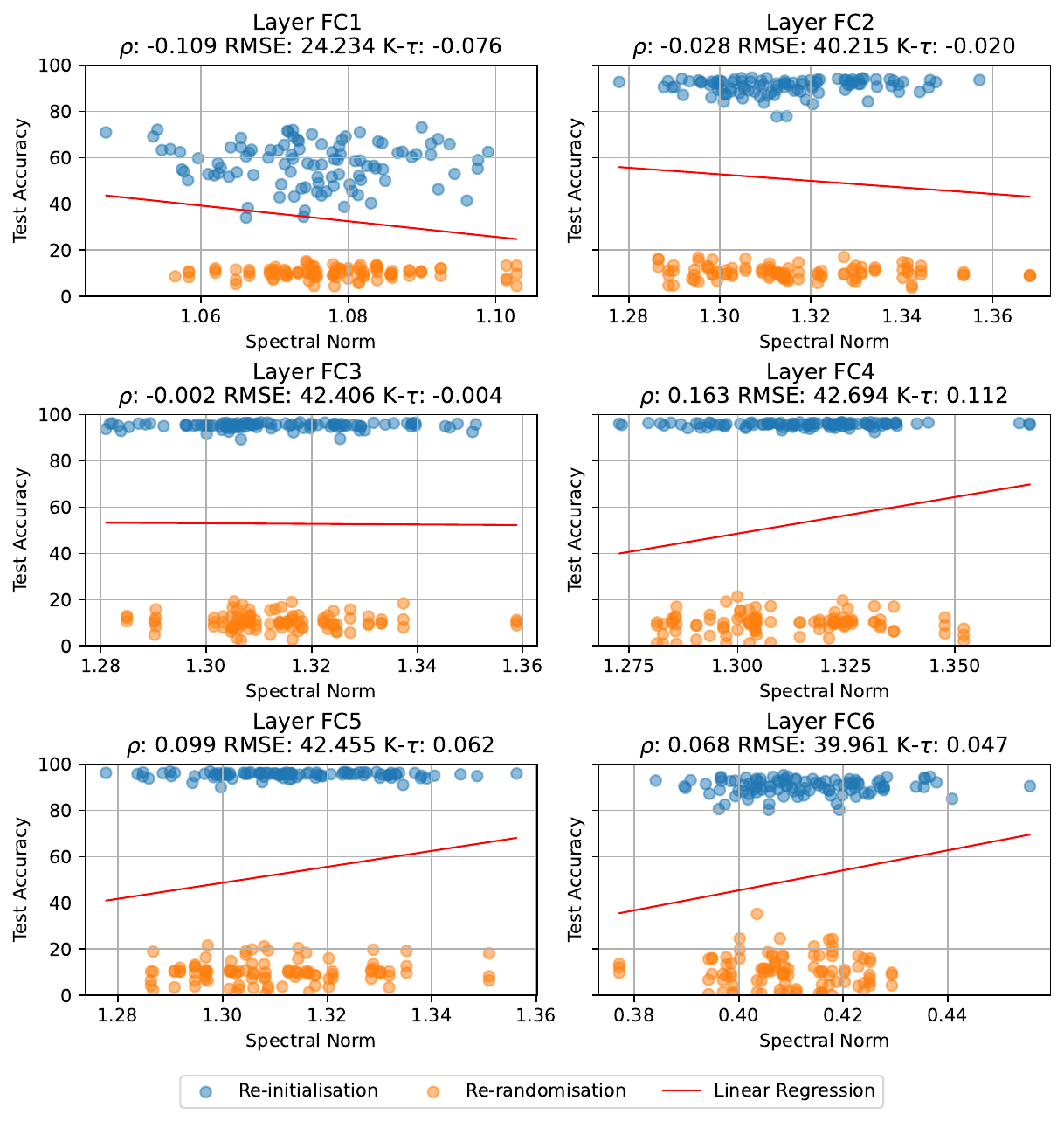}
        \caption{Spectral Norm}
    \end{subfigure}
    \begin{subfigure}[htpb]{0.32\linewidth}
        \centering
        \includegraphics[width=\linewidth]{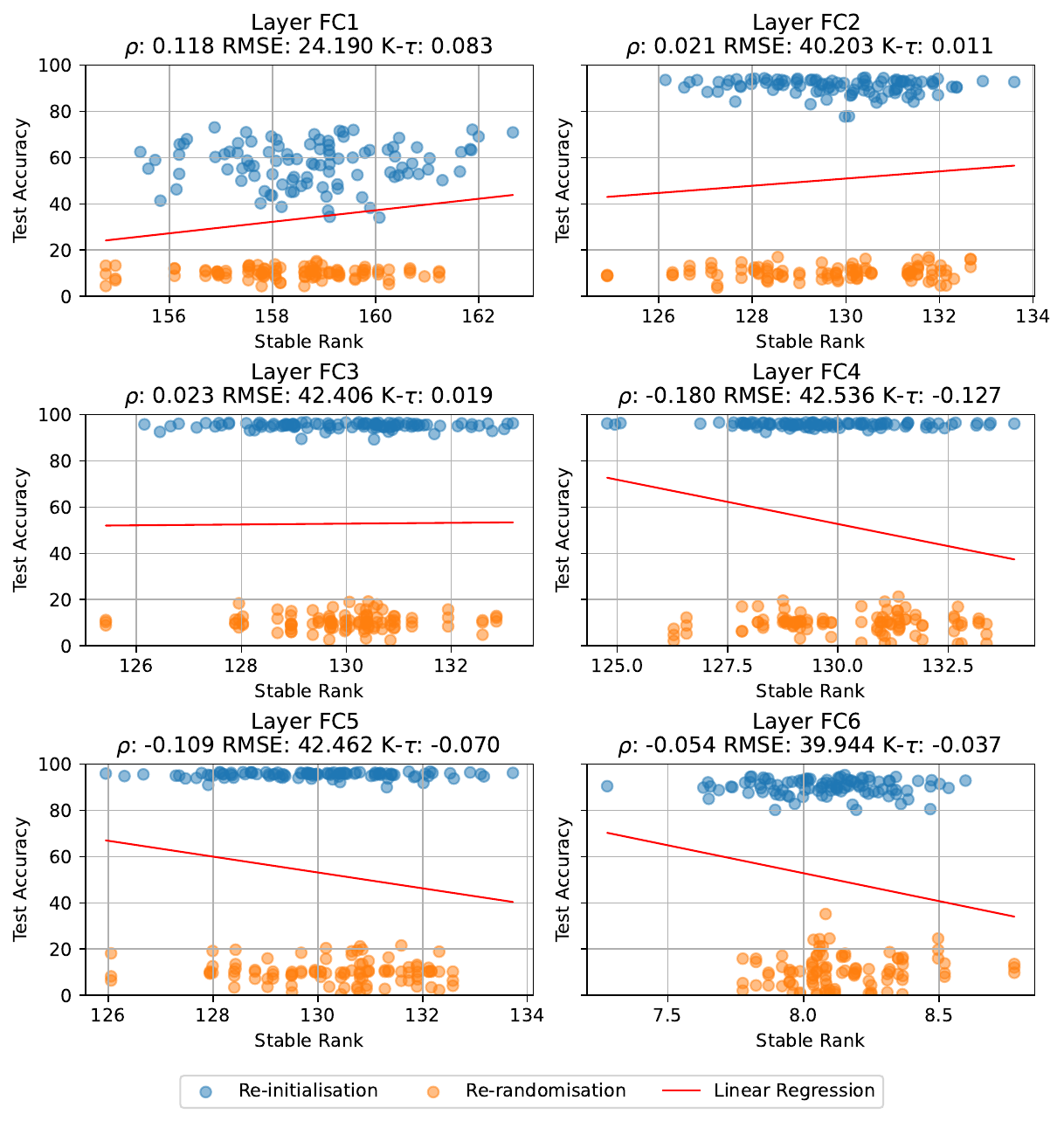}
        \caption{Stable Rank}
    \end{subfigure}
    \begin{subfigure}[htpb]{0.32\linewidth}
        \centering
        \includegraphics[width=\linewidth]{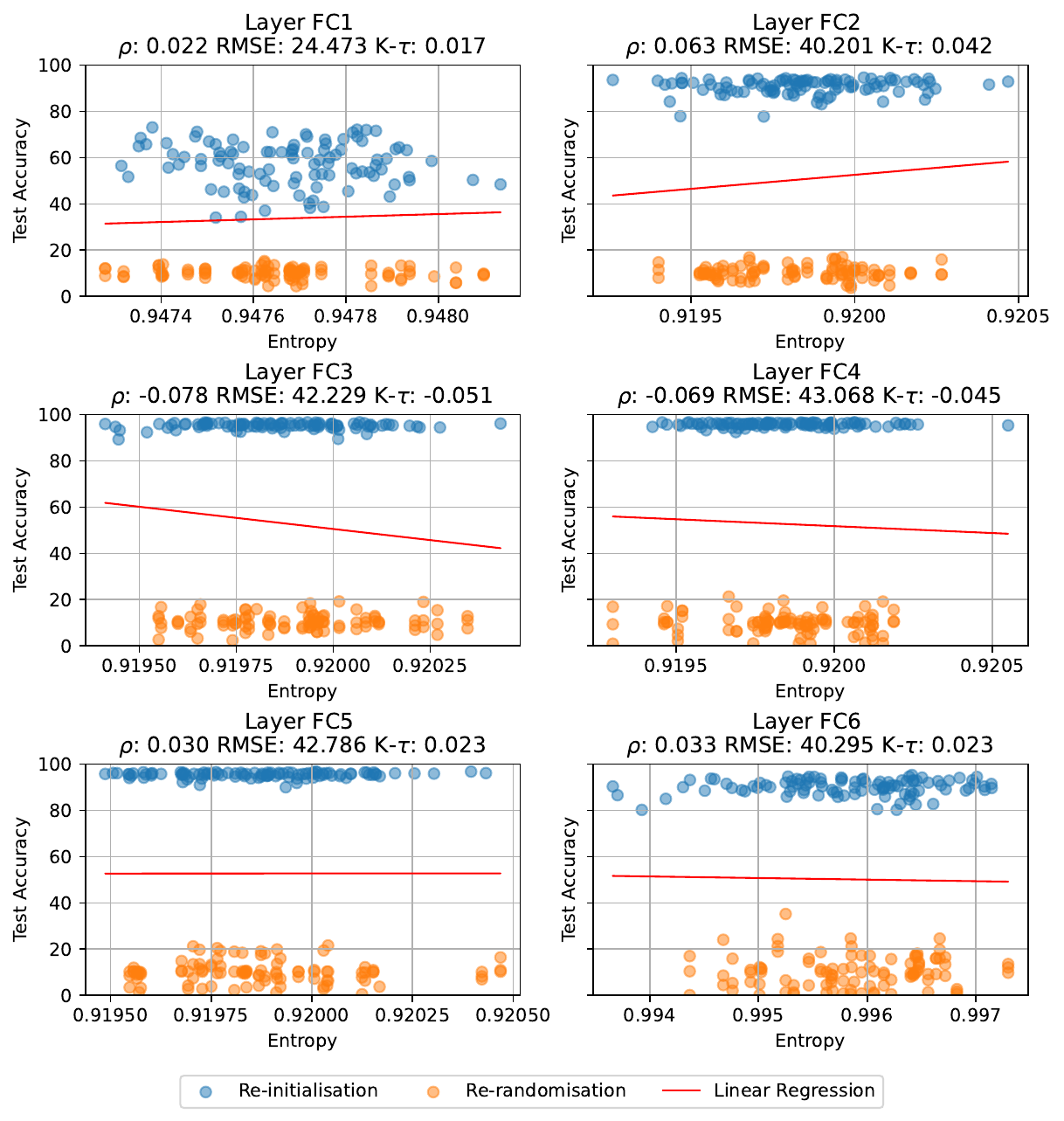}
        \caption{Entropy}
    \end{subfigure}
\end{figure}
\begin{figure}[H]
\ContinuedFloat
\centering
        \begin{subfigure}[htpb]{0.32\linewidth}
        \centering
        \includegraphics[width=\linewidth]{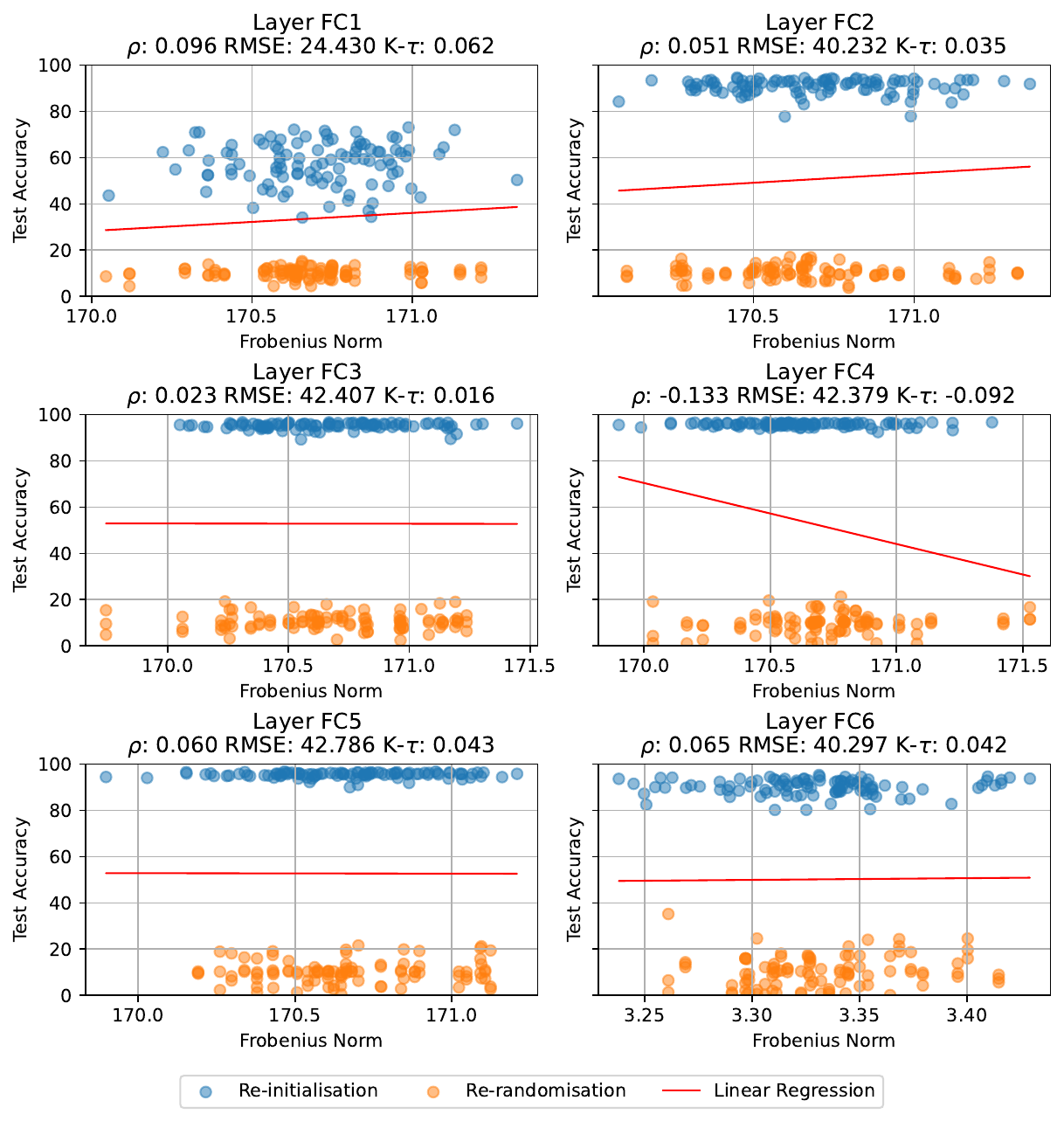}
        \caption{Frobenius Norm}
    \end{subfigure}
    \caption{Layer re-initialisation ({\color{blue}\textbf{$\bullet$}}) and re-randomisation ({\color{orange}\textbf{$\bullet$}}) test accuracy vs the respective metric. $\rho$ is the Spearman correlation coefficient,\space $RSME$ is the root mean square error of the linear regression (red line), and K-$\tau$ is the Kendall's tau measure, all with respect to the relationship between test accuracy and the metric.}
    \label{fig:genralised_fig}
\end{figure}

\newpage
\subsection{ResNet34 on ImageNet}
\label{app:ResNet34}
\begin{figure}[H]
    \centering
    \includegraphics[width=\linewidth]{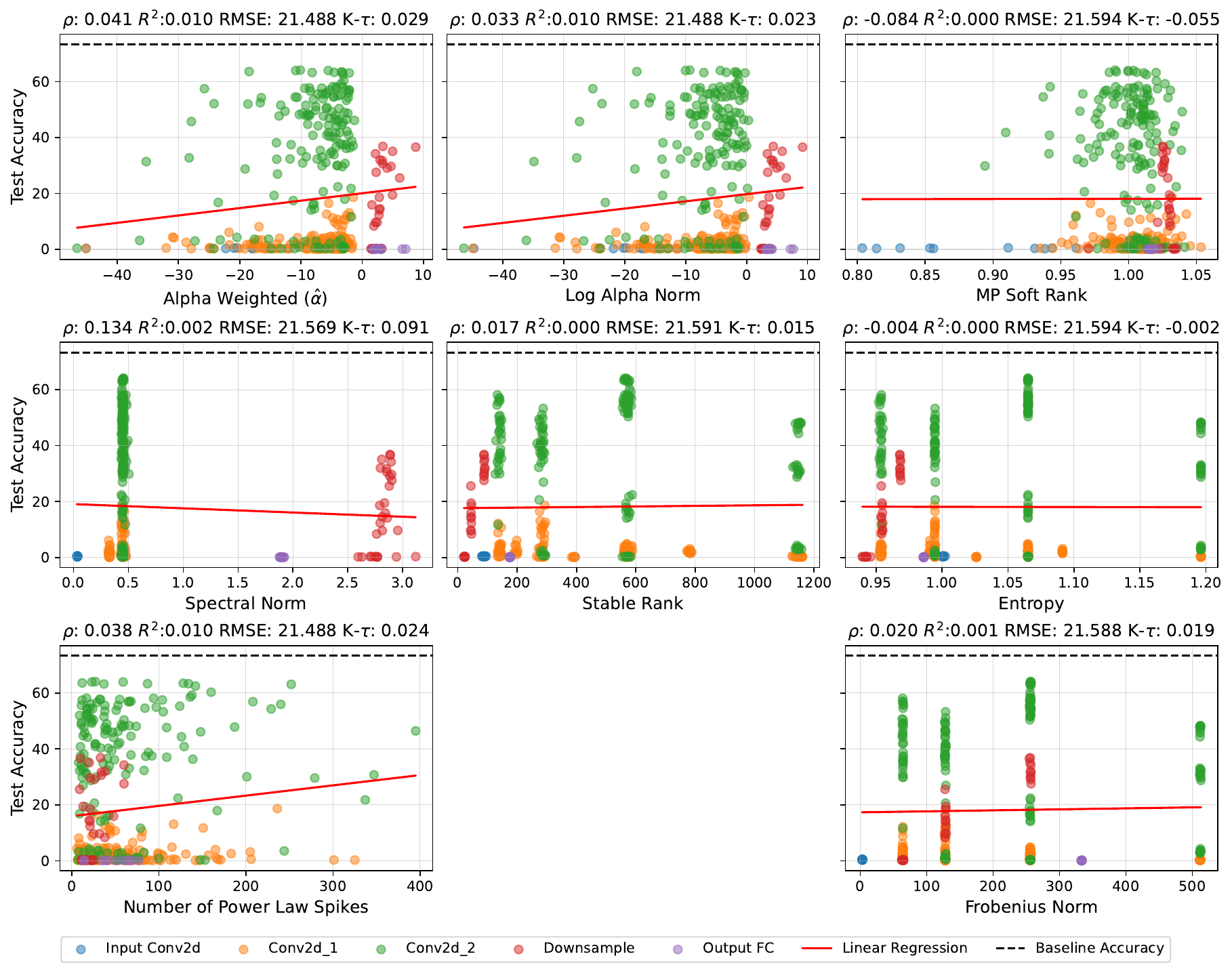}
    \caption{ResNet 34: Layer re-randomisation test accuracy vs the respective metric. $\rho$ is the Spearman correlation coefficient,\space $RSME$ is the root mean square error of the linear regression (red line), and K-$\tau$ is the Kendall's tau measure, all with respect to the relationship between test accuracy and the metric.}
    \label{fig:resnet34-metrics}
\end{figure}
In Figure~\ref{fig:resnet34-metrics}, when using a range of data-free metrics we observe the same trend as found with $\alpha$ in the main body of the paper. Overall for the ResNet43 on ImageNet data-free metrics fail to predict the performance difference that is derived from re-randomisation across layers. While some correlations are higher than circa 0, i.e. (Spectral Norm) it can be clearly observed that there is strong overlap between metric values for the layers that have good accuracy and poor accuracy through re-randomisation, showing that the artifact of slightly higher correlation shows no meaningful insight for these metrics.

\newpage
\subsection{ViT on ImageNet}
\label{app:ViT}
\begin{figure}[H]
    \centering
    \includegraphics[width=\linewidth]{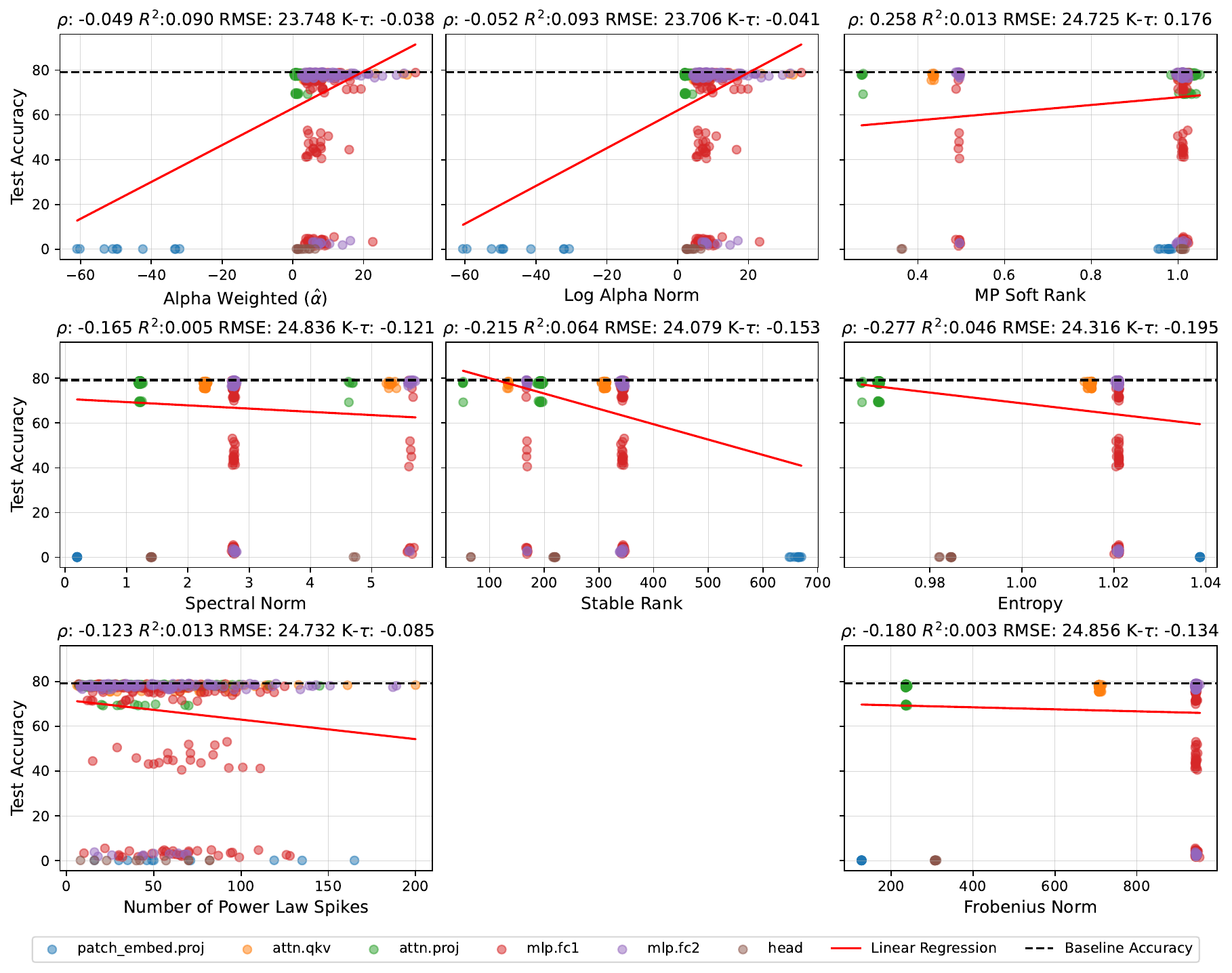}

    \caption{ViT: Layer re-randomisation test accuracy vs the respective metric. $\rho$ is the Spearman correlation coefficient,\space $RSME$ is the root mean square error of the linear regression (red line), and K-$\tau$ is the Kendall's tau measure, all with respect to the relationship between test accuracy and the metric.}
    \label{fig:vit-metrics}
\end{figure}
To confirm trends for the ImageNet dataset on the ViT Figure~\ref{fig:vit-metrics} shows the relationship of data-free metrics value distributions under re-randomisations. Once, again in this figure we find that for all data-free metrics that models can generalise and perform poorly while expressing the same metric values under re-randomisation. As a result the values have no meaningful prediction of performance. This trend is in line with the observations we found with $\alpha$ in the main body of the paper. Here, we do find weak correlations for MP soft rank, spectral norm, stable rank, entropy, number of power law spikes and Frobenius, however, there is significant overlap between the metric values and associated test accuracy (i.e. for strong accuracy and random accuracy). Furthermore for all metrics the $R^2$ value is close to zero indicating that the regression model has little to no predictive power. Overall, data-free metrics can be considered a weak predictor of model performance.
\newpage
\subsection{GPT2 on WikiText103}
\label{app:GPTsmall}

\begin{figure}[H]
    \centering
    \includegraphics[width=\linewidth]{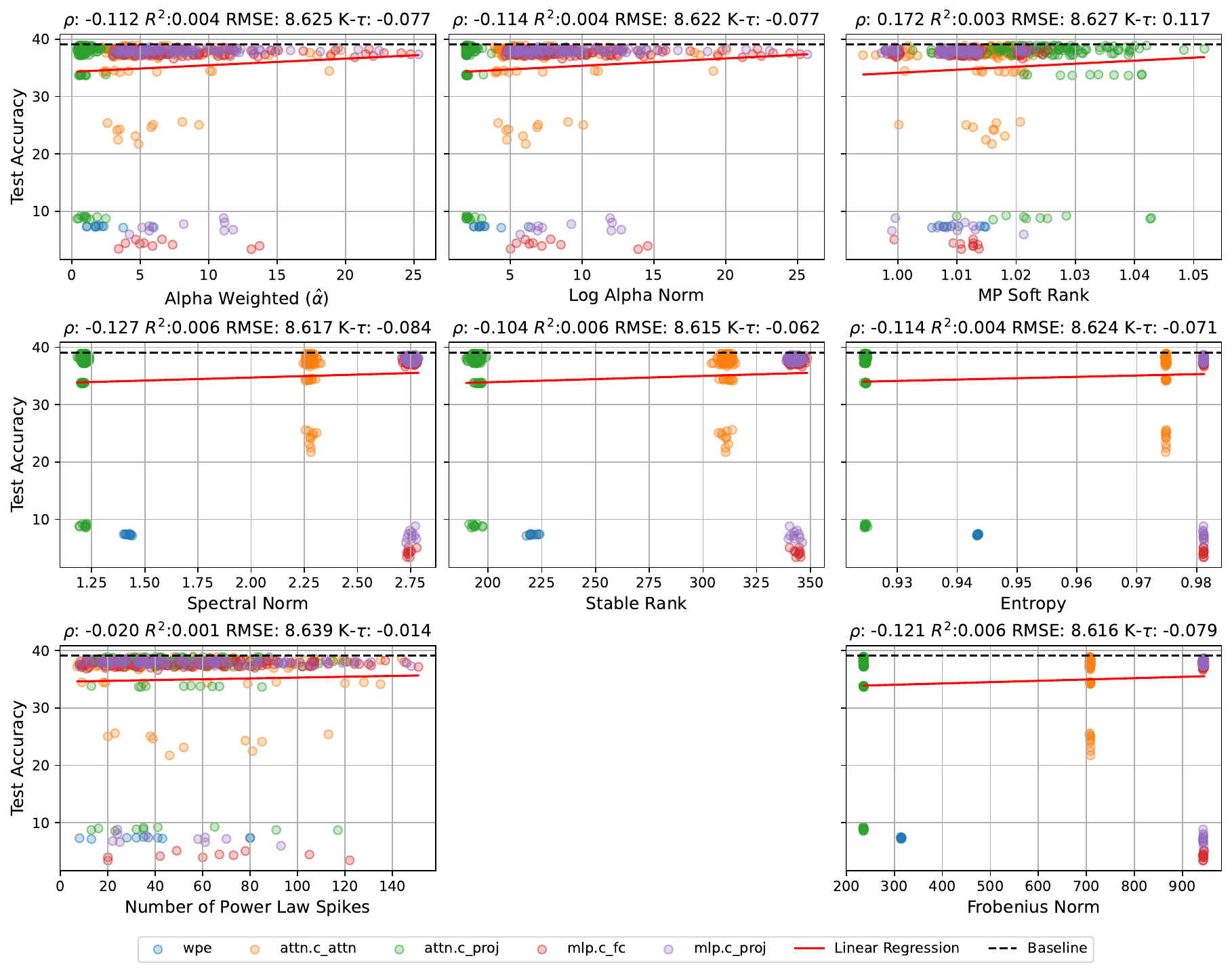}
    
    \caption{GPT2: Layer re-randomisation test accuracy vs the respective metric. $\rho$ is the Spearman correlation coefficient,\space $RSME$ is the root mean square error of the linear regression (red line), and K-$\tau$ is the Kendall's tau measure, all with respect to the relationship between test accuracy and the metric.}
    \label{fig:gpt2-metrics}
\end{figure}
To understand if there is generality in our findings of data-based metrics across modalities and parameters scales we observe the trends of performance and data-free metrics under re-randomisation for the GPT2 model. In Figure~\ref{fig:gpt2-metrics} we largely find a consistent trend that values for data-free metrics can simultaneously represent high and low performing models. This shows that analysing particular values for these data-free metrics provides little to no insight on performance. Here for all metrics there is a very weak correlation between metric values and performance. These trends align with those found with $\alpha$ in the main body of the paper. 
\newpage
\subsection{GPT2-Large on  WikiText103}
\label{app:GPTlarge}
\begin{figure}[H]
    \centering
    \includegraphics[width=\linewidth]{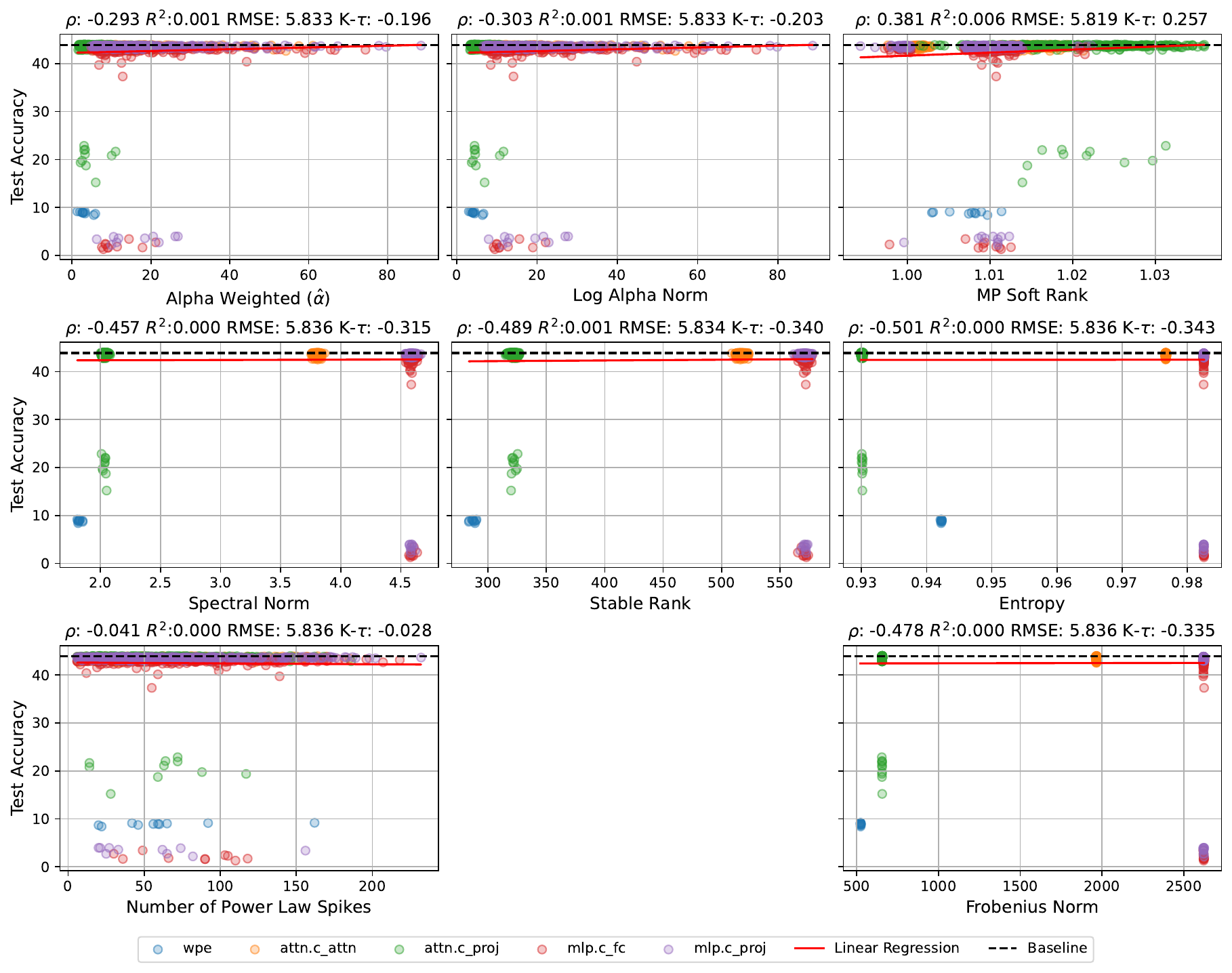}
    \caption{GPT2-Large: Layer re-randomisation test accuracy vs the respective metric. $\rho$ is the Spearman correlation coefficient,\space $RSME$ is the root mean square error of the linear regression (red line), and K-$\tau$ is the Kendall's tau measure, all with respect to the relationship between test accuracy and the metric.}
    \label{fig:gpt2-large-metrics}
\end{figure}
In Figure~\ref{fig:gpt2-large-metrics}, we explore the relationship of data-free metrics and accuracy under re-randomisation for the GPT-2 Large model. Here the trends mirror those for the GPT2 and the results with alpha presented in the main body. For the data-free metrics surveyed there is a moderate to weak correlation between the metric and the test accuracy of the model. However, in spite of this, there is no predictive capacity of the metrics when described with linear regression as shown with the $R^2$ close to zero. This can be attribute to overlapping values that correspond to high and low test accuracy. For the data-free metrics surveyed in this paper across parameter ranges in the GPT2 family we find no predictive capacity for the data-free metrics under re-randomisation. 
\newpage
\subsection{Pythia-1B on  WikiText103}
\label{app:Pythia-1B-all-metrics}
\begin{figure}[H]
    \centering
    \includegraphics[width=\linewidth]{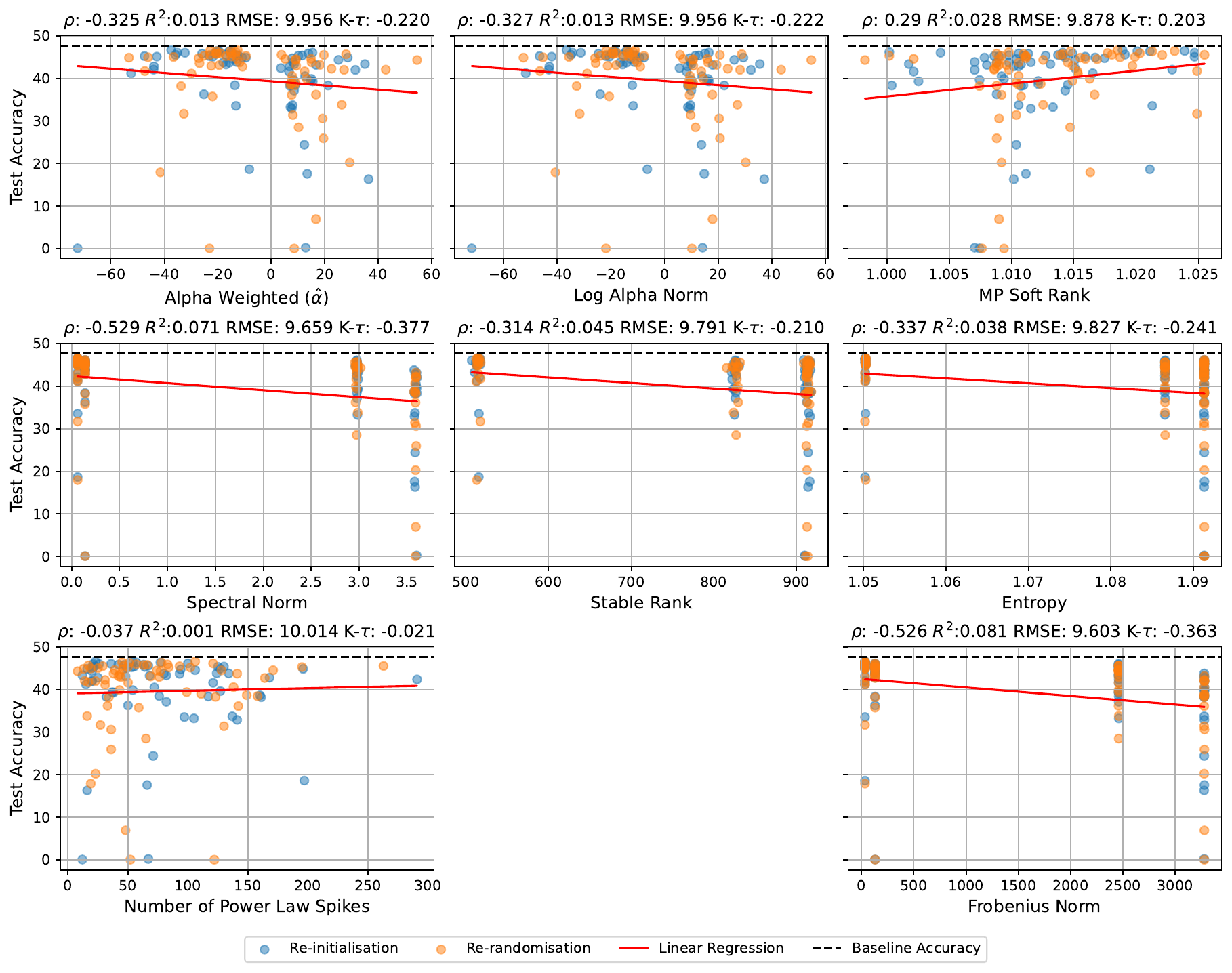}
    \caption{Pythia-1B: Layer re-initialisation (\textcolor{blue}{\textbf{$\bullet$}}) and re-randomisation (\textcolor{orange}{\textbf{$\bullet$}}) test accuracy vs the respective metric. $\rho$ is the Spearman correlation coefficient,\space $RSME$ is the root mean square error of the linear regression (red line), and K-$\tau$ is the Kendall's tau measure, all with respect to the relationship between test accuracy and the metric.}
    \label{fig:pythia-1b-metrics}
\end{figure}
In Figure~\ref{fig:pythia-1b-metrics}, we explore the relationship of data-free metrics and accuracy under re-initialisation (\textcolor{blue}{\textbf{$\bullet$}}) and re-randomisation (\textcolor{orange}{\textbf{$\bullet$}}) for the Pythia-1B model. Here the trends mirror those for the GPT2 models and the results with alpha presented in the main body. For the data-free metrics surveyed there is a moderate to weak correlation between the metric and the test accuracy of the model. However, in spite of this, there is no predictive capacity of the metrics when described with linear regression as shown with the $R^2$ close to zero. Which is due to overlapping values that correspond to high and low test accuracy. For the data-free metrics surveyed we find no predicative capacity for said metrics under re-initialisation and re-randomisation. 
\newpage
\section{Data-Based Metrics}
\label{sec:data-based-metrics}
\begin{figure}[H]
    \centering
    \begin{subfigure}[b]{0.49\linewidth}
        \includegraphics[width=\linewidth]{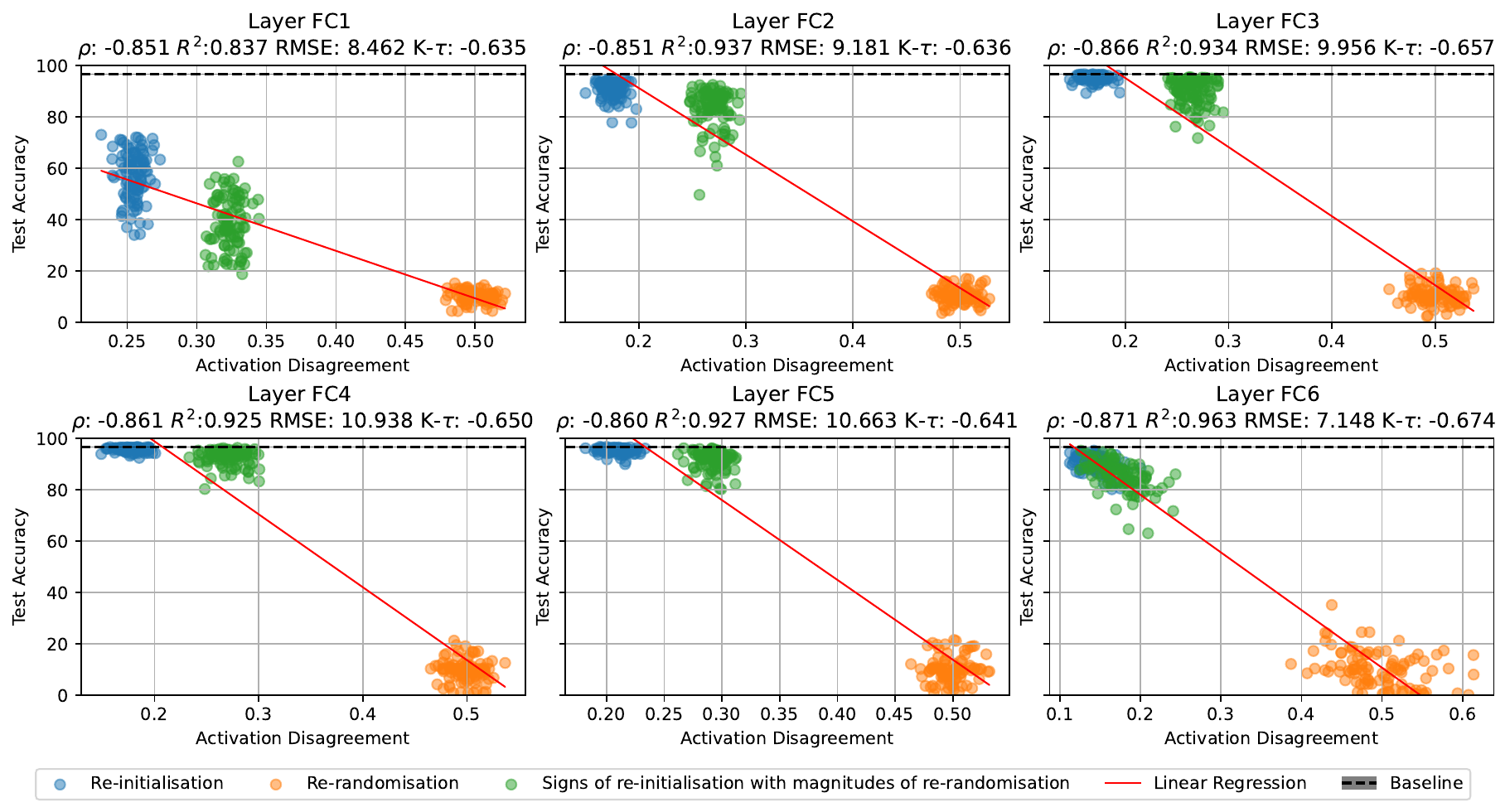}
        \caption{Activation Disagreement}
    \end{subfigure}
    \begin{subfigure}[b]{0.49\linewidth}
        \includegraphics[width=\linewidth]{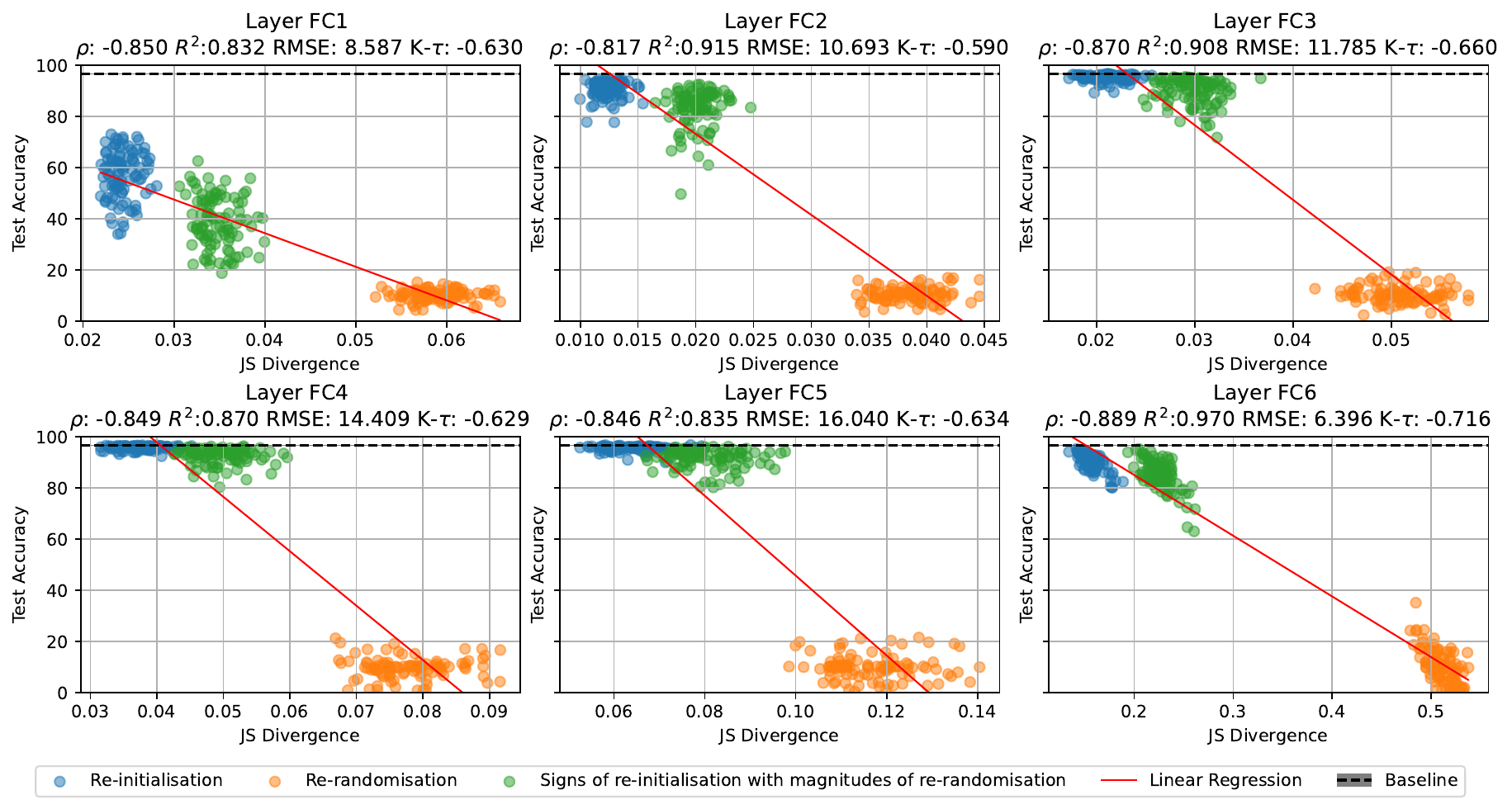}
        \caption{JS Divergence}
    \end{subfigure}

    \caption{Layer re-initialisation ({\color{blue}\textbf{$\bullet$}}), re-randomisation ({\color{orange}\textbf{$\bullet$}}) and signs of re-initialisation with magnitudes of re-randomisation (\textcolor{green}{\textbf{$\bullet$}}) test accuracy vs the respective metric.}
    \label{fig:activation-init-rand-signs}
\end{figure}
To compare data-free and data-based metrics we explore if two widely used data-based metrics can distinguish the output of a re-initialised layer from that of a re-randomised layer. The outputs of the original and modified layers are represented as $\mathcal{O}_l$ and $\mathcal{M}_l$, respectively. The model layer outputs are collected by passing through the test dataset,\space $\mathcal{D}_{test}$. The metrics explored are defined as follows, Linear CKA~\citep{cka} is explored in Appendix~\ref{sec:cka}):
\begin{itemize}
    \item \textbf{Activation Disagreement:} The mean percentage of times that the neurons from $\mathcal{O}_l$ and $\mathcal{M}_l$ fail to agree to activate on $\mathcal{D}_{test}$. 
    \item \textbf{Jensen–Shannon (JS) Divergence:} The mean weighted average of the Kullback–Leibler (KL) Divergence of the softmax( $\mathcal{O}_l$ on $\mathcal{D}_{test}$) compared to the softmax( $\mathcal{M}_l$ on $\mathcal{D}_{test}$)~\citep{js_divergence}.
\end{itemize}
When considering the results of the activation disagreement and JS divergence between the original model layer and each model with re-initialised and re-randomised layers in Figure~\ref{fig:activation-init-rand-signs}, it becomes clear that there is a stark difference in similarity between the re-initialised and re-randomised layers compared to the unmodified layer. The figure demonstrates that the difference between re-initialisation and re-randomisation can be explained by the amount the layer disagrees on activation compared to the original non-modified layer. It is evident that when there is less disagreement (re-initialisation), the model can maintain accuracy, while a lot of disagreement (re-randomisation) leads to a drop in accuracy. This observation is further supported by a mean Spearman correlation coefficient and Kendall's tau measure across layers of  -0.776 and -0.5375, respectively for Activation Disagreement, indicating a negative relationship between activation disagreement and accuracy. 
\paragraph{Connection to the Lottery Ticket Hypothesis:}
\label{sec:lottery-ticket-hypothesis}
For our MNIST experiments, both re-initialisation and re-randomisation are drawn from the Uniform distribution,\space $\mathcal{U}(-\sqrt{k}, \sqrt{k})$ where $k$ is $\frac{1}{\text{in\_features}}$. However, a layer can reduce to random accuracy under re-randomisation but not under re-initialisation, which suggests that an important factor for performance retention under re-initialisation is the signs of the weights. The signs of the weights, especially within networks that use the ReLU activation, essentially represent the pathway, or subnetworks within the model. The Lottery Ticket Hypothesis (LTH) suggests that at random initialisation, there exist subnetworks that, when trained, reaches the test accuracy of the original model that is trained with all the parameters and same compute budget~\citep{frankle2018lottery}. To explore the role of the signs, we take the re-randomised values for a layer and apply the signs of the re-initialised layer to the re-randomised weights. We explore this using the data-based metrics that can disambiguate between re-initialised and re-randomised. We find in Figure~\ref{fig:activation-init-rand-signs} that when applying the signs of the re-initialisation to the magnitudes of the re-randomisation layer (\textcolor{green}{\textbf{$\bullet$}}), this layer is closer to the original layer as observed with a reduction in difference, and that there is an increase in performance when compared to the re-randomised layer ({\color{orange}\textbf{$\bullet$}}). The results strongly suggest that for this network, the subnetworks defined at the start of training through initialisation do not dramatically change through training, providing credence to the LTH. Furthermore, this demonstrates the efficacy of data-based metrics to disambiguate between the subnetwork importance of layers in neural networks.

\section{Linear CKA (Data-Based Metric) Analysis }
\label{sec:cka}

\begin{figure}[H]
    \centering
    \includegraphics[width=0.8\linewidth]{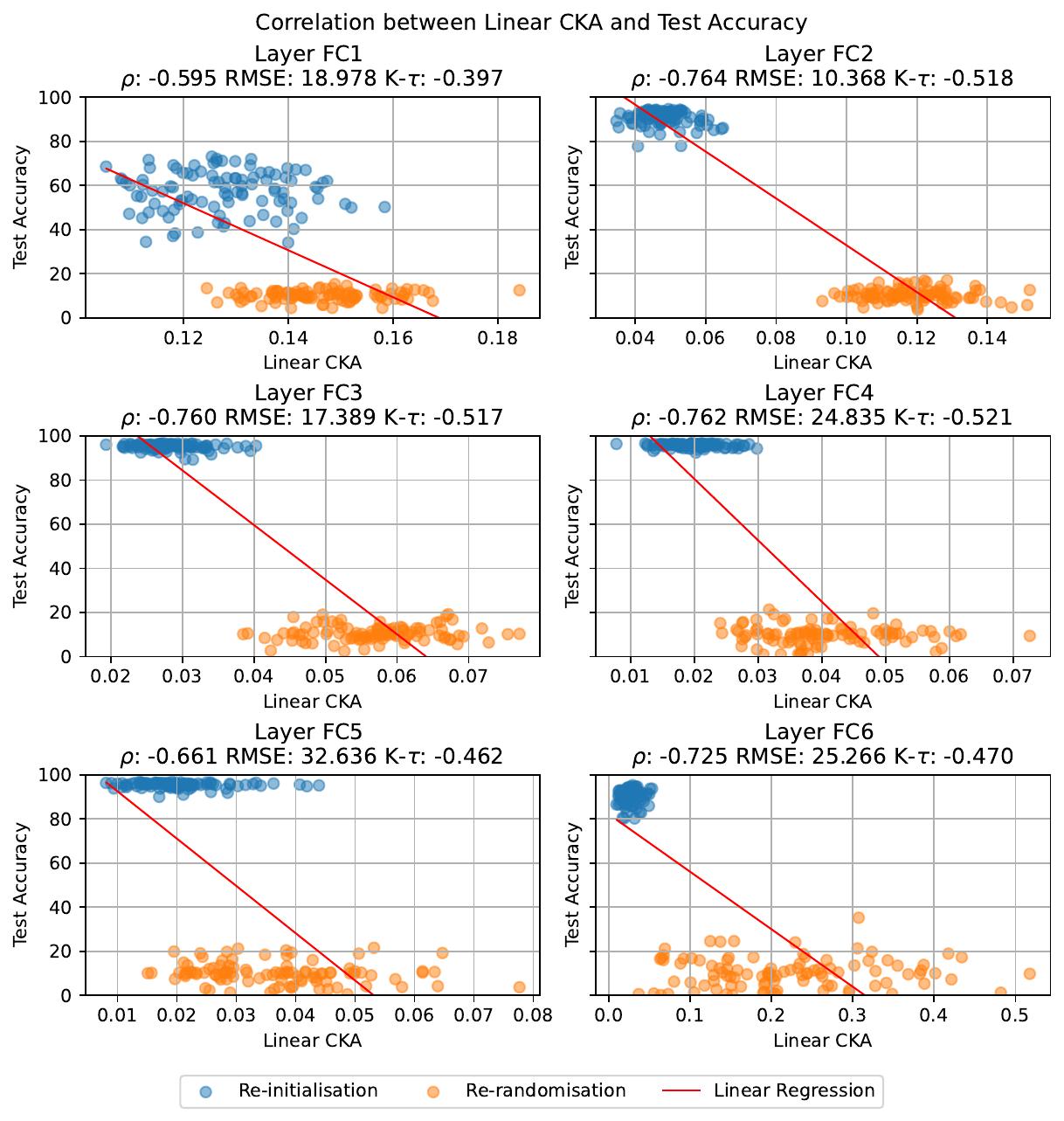}
    \caption{Layer re-initialisation ({\color{blue}\textbf{$\bullet$}}) and re-randomisation ({\color{orange}\textbf{$\bullet$}}) test accuracy vs the respective metric. $\rho$ is the Spearman correlation coefficient,\space $RSME$ is the root mean square error of the linear regression (red line), and K-$\tau$ is the Kendall's tau measure, all with respect to the relationship between test accuracy and the metric.}
    \label{fig:linear-cka}
\end{figure}

In this section, we explore an additional data based metric, Linear CKA~\citep{cka}. We asses the predictive power of Linear CKA to distinguish between the output of a re-initialised layer from that of a re-randomised layer. Overall we find that in line with other data-based metrics explored in the main paper that Linear CKA can disambiguate between a re-initialised and re-randomised layer. However, Linear CKA has less predictive power than the Activation Disagreement and JS-divergence metrics explored in the body of the paper, as demonstrated with lower correlation values and higher RMSE, see Figure~\ref{fig:linear-cka}. The results highlight an important caveat when considering data-based metrics, not all data-based metrics can be considered equal for predictive capacity. As a result, it remains an open challenge to find test beds for identifying and utilising the correct data-based metrics that effectively consider subnetwork relationships and disambiguate between robust and critical layers.

\section{SVD Ablation for Learning Dynamics}
\label{sec:exp-spectral-vs-training}

\begin{figure}[H]
    \centering
    \includegraphics[width=\linewidth]{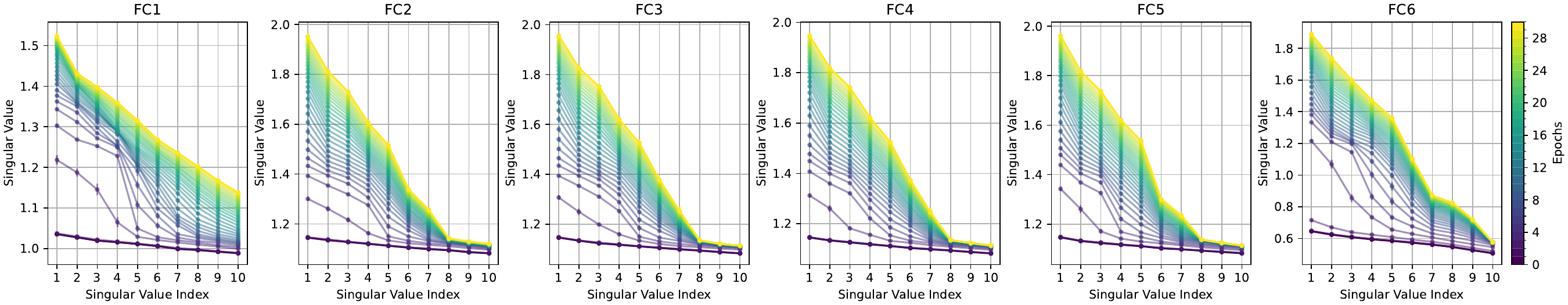}
    \caption{The first 10 singular values for each layer of a 5 hidden layer neural network with width 512 throughout training on MNIST. The viridis colormap is used to delineate epochs. \textcolor{dark-purple}{\textbf{Dark-purple}} represents initialisation and the last epoch, epoch 30, is in \textcolor{bright-yellow}{\textbf{bright-yellow}}. The mean (line) is reported with the standard error of the mean (error bars) from 10 runs.}
    \label{fig:singulars_through_training}
\end{figure}

The empirical experiments in this paper have  provided substantial evidence that data-free metrics may bear no relationship to the generalisation properties of neural networks. However, it can be argued that the spectral properties of weight matrices need to to change in order for a model to generalise, particularly since singular values do change substantially in standard training setups. In Figure~\ref{fig:singulars_through_training} we show how the singular values in an MNIST model change across epochs.

Given, that the singular values values do change substantially through training, a plausible interpretation of these metrics is that they record properties of the neural network that are important in the training process that results in a well generalising model. 

Motivated by this interpretation, we conduct experiments to establish the role of the spectral properties of the weight matrices in well generalising models. The following experiments explore how the factors in the SVD decomposition $W=U\Sigma V^T$ of weight matrices (i.e.\ the singular vectors and the singular values) affect training.
We explore each factor independently by keeping the others fixed during training, and only letting the corresponding factor update throughout training - to isolate a single factors effect. This enables a systematic intervention analysis to understand what factors primarily drive generalisation and which ones require interactions. The factors are also explored jointly to understand the role of the interaction of factors on generalisation. Finally, the factors are all trained together to create an effective baseline of complete interaction. 

If the model is unable to generalise without the singular values -- and thus spectral properties -- changing, this would support the interpretation that these metrics record property changes that are required for models to reach a minima that generalises well. Conversely, if a model can generalise without these properties changing, this would mean that these metrics are not casually relevant to finding a minima that generalises and thus has limited impact on generalisation in training.

\subsection{Experimental Setup}

To reduce confounding factors, we consider a linear layer where the weight matrices are parametrised via singular value decomposition,\space $W=U\Sigma V^T$, to from three layers where $U$ and $V^T$ are orthogonal, and $\Sigma$ is diagonal with non-negative entries. 

This parametrization allows for the orthogonal direction factors $U$\space and $V^T$\space and the scaling factor $\Sigma$\space of the weight matrix to be manipulated and changed independently. Thereby enabling a controlled analysis of their respective roles in the optimization process and generalisation.

For all experiments we use a 5 hidden layer MLP without biases with the ReLU activation and a hidden width of 512 on the MNIST dataset with cross entropy loss, SGD with a learning rate of 0.001 and momentum of 0.9, a batch size of 128 and trained for 30 epochs. To isolate the effect of parameter interventions, the training data order, random seeds and initialisation procedures (expect where explicitly varied) are held fixed across all runs. Model biases are not such that the factors are the only aspect impacting model performance.

To ensure that $U$\space and $V^T$\space maintained orthogonality throughout training we employed the standard PyTorch methods of matrix exponential for square matrices and the householder transformation for rectangular matrices\footnote{\url{https://docs.pytorch.org/docs/stable/generated/torch.nn.utils.parametrizations.orthogonal.html}}. This ensures that changes in the performance cannot be attributed to the loss of orthogonality. 

Using random $\Sigma$\space drawn from the Marchenko–Pastur distribution introduces implicit conditioning on the orthogonal directions of $U$\space and $V^T$\space. Therefore $\Sigma$\space is initialised as $\Sigma$\space$=\boldsymbol{I}$, where $\boldsymbol{I}$ is the identity matrix. This removes the scaling effects at initialisation and ensures that all singular values are treated equally by the optimiser. As a result, any observed differences in training dynamics can be attributed to the orthogonal directions of $U$\space and $V^T$\space or to the values of $\Sigma$ rather than the co-founding effects that could arise from the initial unequal spectral scaling. Furthermore initialising with $\Sigma$\space$=\boldsymbol{I}$ fixes the spectral properties across runs, facilitating a controlled comparison of the effects of the spectral properties between interventions. 

\subsubsection{Interventional Ablation Over Parametrisation Subsets}
\label{app:inverentional_abalalation}

To study the interactions of the SVD factors during training, we perform controlled interventions, in which gradient updates are enabled on specific subsets of parametrisations, while the remaining parametrisations are held fixed with the initial values - allowing for a controlled and isolated optimisation setup. 

For each experiment, a subset $\mathcal{T} \subseteq \{U, \Sigma, V^T \}$ is assigned as a trainable factor. All factors in the complement $\{$$U$ $\Sigma$ $V^T$ $\} \setminus \mathcal{T}$ are frozen and do not change throughout training. 

We consider the following intervention sets: 

\begin{itemize}
    \item Single-Factor Training $\mathcal{T} \in \{\{U\}, \{\Sigma\}, \{V^T\}\}$
    \item Two-Factor Training $\mathcal{T} \in \{\{U\Sigma\}, \{\Sigma V^T\}, \{UV^T\}\}$
    \item Full-Factor Training $\mathcal{T} \in \{\{U\Sigma V^T\}\}$
\end{itemize}

\paragraph{Single Factor Training:} The single factor training experiments ( $U$ $\Sigma$\space and $V^T$\space interventions) allow for understanding of the role of the of each component in the learning process. The performance in these conditions reveals whether training progress can be achieved through spectral scaling ($\Sigma$) or through rotations of the singular vectors ($U$\space and  $V^T$) alone. Failure of a factor to effectively generalise indicates that learning relies on the interaction between that factor and others. Whereas the success of a factor implies that the corresponding factor is a primary driver of the learning dynamics. 

\paragraph{$\boldsymbol{U}$\space Intervention:}

To isolate the effects of the $U$\space orthogonal directions during training,\space $\Sigma$\space is fixed as $\Sigma$\space$= \boldsymbol{I}$ and $V^T$\space is fixed as $V^T_0$, where $V^T_0$ is orthogonal matrix, throughout training. For each run $U$\space is initialised with a random orthogonal matrix and allowed to train. Keeping $V^T$\space fixed as $V^T_0$ throughout runs isolates the effects of variation in performance ensuring it cannot be attributed to changes in the initial conditions of $\Sigma$\space and $V^T$. 

\paragraph{$\boldsymbol{\Sigma}$\space Intervention:} To isolate the effects of the $\Sigma$\space during training,\space $\Sigma$\space is initialised as $\Sigma$\space$= \boldsymbol{I}$ the orthogonal matrices $U$\space and $V^T$\space are randomly initialised as orthogonal matrices and then held fixed throughout training. Between runs the initial values of $U$\space and $V^T$\space change to ensure that the results are not conditional on a single orthogonal direction.  

\paragraph{$\boldsymbol{V^T}$\space Intervention:} To isolate the effects of the $V^T$\space orthogonal directions during training,\space $\Sigma$\space is fixed as $\Sigma$\space$= \boldsymbol{I}$ and $U$\space is fixed as $U_0$, where $U_0$ is orthogonal matrix, throughout training. For each run $V^T$\space is initialised with a random orthogonal matrix and allowed to train. Keeping $U$\space fixed as $U_0$ throughout runs isolates  the effects of run variation in performance ensuring it cannot be attributed to changes in the initial conditions of $\Sigma$\space and $U$.

\paragraph{Two Factor Training:} The two factor training experiments (Joint $U\Sigma$,\space $\Sigma V^T$\space and $UV^T$\space interventions) allow for understanding the interactions between components and how this affects the training process and subsequent generalisation properties. Improvements in performance above the single factor cases, suggests that the interaction of the two factors is important to overall performance. Although if performance is comparable to full training this would suggests that the frozen factor does not significantly impede the training process.

\paragraph{Joint $\boldsymbol{U\Sigma}$\space Intervention:}
To isolate the extent to which performance improvements arise from the from the combined changes in $\Sigma$\space and $U$\space. The $U\Sigma$\space parametrisation jointly optimises $U$\space and $\Sigma$\space while keeping $V^T$\space fixed as $V^T_0$ throughout training. 

\paragraph{Joint $\boldsymbol{\Sigma V^T}$\space Intervention:}
To isolate the extent to which performance improvements arise from the from the combined changes in $\Sigma$\space and $V^T$\space. The $\Sigma V^T$\space parametrisation jointly optimises $\Sigma$\space and $V^T$\space while keeping $U$\space fixed as $U_0$ throughout training. 

\paragraph{Joint $\boldsymbol{UV^T}$\space Intervention:}
To isolate the extent to which performance improvements arise from the from the combined changes in $U$\space and $V^T$\space. The $UV^T$\space parametrisation jointly optimises $U$\space and $V^T$\space while keeping $\Sigma$\space is fixed throughout training. 

\paragraph{Full Factor Training $\boldsymbol{U\Sigma V^T}$:}
In the full factor training $U\Sigma V^T$ all components ($U$ $\Sigma$ and $V^T$) are jointly optimised. This corresponds to standard training on the weight matrix $W$, and thus serves as a reference for the impact of the  single and two-factor interventions on the performance. 

Table~\ref{tab:intervention-factors} summaries all the interventional conditions explored in this experiment. For this experiment instead of using 100 runs, 10 runs are used as the orthogonal constraint on the $U$\space and $V^T$\space components significantly increased training times. The test accuracy, loss and metrics are recorded at initialisation and throughout training at the end of every epoch.

\begin{table}[H]
\centering
\caption{Summary of factor interventional conditions.}
\label{tab:intervention-factors}
\begin{tabular}{|l|l|l|l|}
\hline
\rule{0pt}{2.5ex} 
\textbf{Trainable $\boldsymbol{U}$} & \textbf{Trainable $\boldsymbol{\Sigma}$} & \textbf{Trainable $\boldsymbol{V^T}$} & \textbf{Description}  \\ \hline \rule{0pt}{2.5ex} 
\cmark                    & \xmark                    & \xmark                    & Output Alignment Only \\ \hline \rule{0pt}{2.5ex} 
\xmark                    & \cmark                    & \xmark                    & Spectral Scaling Only \\ \hline \rule{0pt}{2.5ex} 
\xmark                    & \xmark                    & \cmark                    & Input Alignment Only  \\ \hline \rule{0pt}{2.5ex}
\cmark                    & \cmark                    & \xmark                    & Joint $U$\space and $\Sigma$\space ($U\Sigma$)         \\ \hline \rule{0pt}{2.5ex}
\xmark                    & \cmark                    & \cmark                    & Joint $\Sigma$\space and $V^T$\space ($\Sigma V^T$)         \\ \hline \rule{0pt}{2.5ex}
\cmark                    & \xmark                    & \cmark                    & Joint $U$\space and $V^T$\space ($UV^T$)         \\ \hline \rule{0pt}{2.5ex}
\cmark                    & \cmark                    & \cmark                    & Standard Training ($W=$\space$U\Sigma V^T$)   \\ \hline
\end{tabular}%
\end{table}

\subsection{Experimental Results}

The following experimental results cover the relationship between the factors and test performance (\S~\ref{sec:result-param-vs-test-acc}) and data-free metrics (\S~\ref{sec:result-metric-vs-test-acc}) throughout training.

\subsubsection{Factor Relationship with Model Performance}
\label{sec:result-param-vs-test-acc}

\begin{figure}[H]
    \centering
    \includegraphics[width=0.75\linewidth]{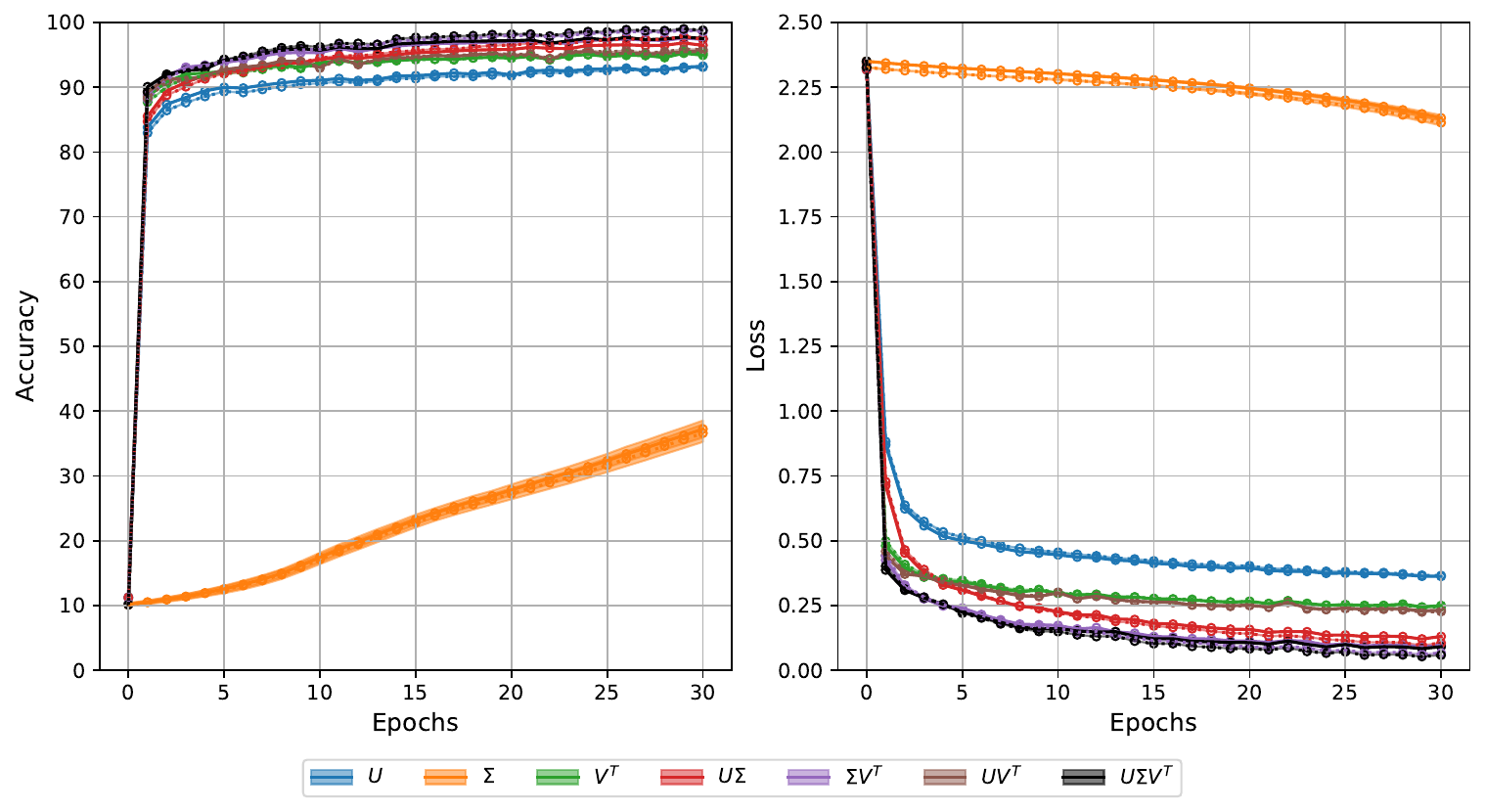}
    \caption{Model test performance throughout training on MNIST. (left) accuracy and (right) loss. The dotted line ($\dots$) is the train performance and the solid line (--) is the test performance, the $\circ$ represents a epoch. Each curve corresponds to training only one or more components of the parametrisation: $U$ $\Sigma$ $V^T$ $U\Sigma$,\space $\Sigma V^T$,\space $UV^T$\space and $U\Sigma V^T$. All other parameters are held fixed during training.  Performance is evaluated on the at the end of each epoch. The mean (line) and standard error from the mean (hue) are reported from 10 runs. }
    \label{fig:1_1_no_bias_test_perf}
\end{figure}

Figure~\ref{fig:1_1_no_bias_test_perf} compares the performance throughout training when different subsets of the models parameters are updated during training. Training all components jointly,\space $U\Sigma V^T$, as is done conventionally through optimisation directly on the weight matrix $W$, results in the model with the highest test accuracy at the end of training of $97.499\%$.

Training only a single component ($U$\space $\Sigma$\space or $V^T$) results in consistently lower performance than $U\Sigma V^T$. The $U$\space parametrisation reaches a high initial test accuracy after 1 epoch training ($83.811\%$) and then continues to improve in performance achieving a test accuracy of $93.236\%$ at the end of training. In contrast, updating only $\Sigma$\space results in a markedly slower initial improvement with slow but steady performance gains, yet remains significantly below the performance levels achieved by all other parametrisations throughout training resulting in a test accuracy of $37.246\%$. As found with only updating $U$\space updating only $V^T$\space  reaches a high initial test accuracy after 1 epoch training ($88.628\%$) a $4.817\%$ increase in test accuracy over $U$\space parametrisation. The $V^T$\space parametrisation then steadily increases in performance reaching a test accuracy of $94.924\%$ a $1.688\%$ improvement over the $U$\space parametrisation.

Jointly updating pairs of components ($U\Sigma$,\space $\Sigma V^T$\space and $UV^T$) improves the resulting test performance relative to the single-components ($U$\space $\Sigma$\space and $V^T$). Both $U\Sigma$\space and $\Sigma V^T$\space parametrisations follow the similar initial training trajectories as $U$\space and $V^T$\space respectively. After 1 epoch of training the $U\Sigma$\space parametrisation achieves a test accuracy of $85.506\%$ a $1.695\%$ improvement over the $U$\space parametrisation. The $U\Sigma$\space  parametrisation subsequently makes increased performance gains throughout training resulting in a test accuracy of $96.458\%$, only $1.041\%$ of training with all components jointly ($U\Sigma V^T$). Additionally after 1 epoch training the $\Sigma V^T$\space parametrisation achieves a test accuracy of $88.898\%$ a $0.270\%$ improvement over the $V^T$\space parametrisation and $1.168\%$ below $U\Sigma V^T$\space parametrisation. The $\Sigma V^T$\space  parametrisation subsequently makes increased performance gains throughout training and closely matches the training trajectory of $U\Sigma V^T$, resulting in a test accuracy of $97.373\%$, only $0.126\%$ of training with all components jointly ($U\Sigma V^T$). The $UV^T$\space parametrisation, closely matches the training trajectory of $V^T$\space with the initial epoch of training reaching a test performance of $89.244\%$ a $0.616\%$ improvement over the $V^T$\space parametrisation and a $5.433\%$ improvement over the $U$\space parametrisation. The $UV^T$\space parametrisation continues to closely match the $V^T$\space parametrisation with a resulting test performance of $95.524\%$ a $0.600\%$ improvement over the $V^T$\space parametrisation.

In summary, Figure~\ref{fig:1_1_no_bias_test_perf}, shows that the configurations exhibit a consistent ordering in test performance with $U\Sigma V^T$\space performing the best, followed by $\Sigma V^T$,\space $U\Sigma$,\space $UV^T$,\space $V^T$ $U$\space and $\Sigma$. With the difference between $U$\space and $\Sigma$\space of $55.990\%$ compared to a difference between $U\Sigma V^T$\space and $U$\space of $4.263\%$. Although $\Sigma$\space was unable to achieve comparable or close performance to $U\Sigma V^T$\space the inclusion of $\Sigma$\space to $U$\space and $V^T$\space parametrisations in $U\Sigma$\space and $\Sigma V^T$\space enabled improvements in test performance over $U$\space and $V^T$\space of $3.222\%$ and  $2.449\%$ respectively. The next section explores the relationship of data-free metrics throughout training to better understand how our intervention setup effects matrix (and data-free metric) properties during training and their relationship to accuracy. 

\subsubsection{Metric Relationship with Test Accuracy}
\label{sec:result-metric-vs-test-acc}
\begin{figure}[H]
    \centering

    \begin{subfigure}[htbp]{0.32\linewidth}
        \includegraphics[width=\linewidth]{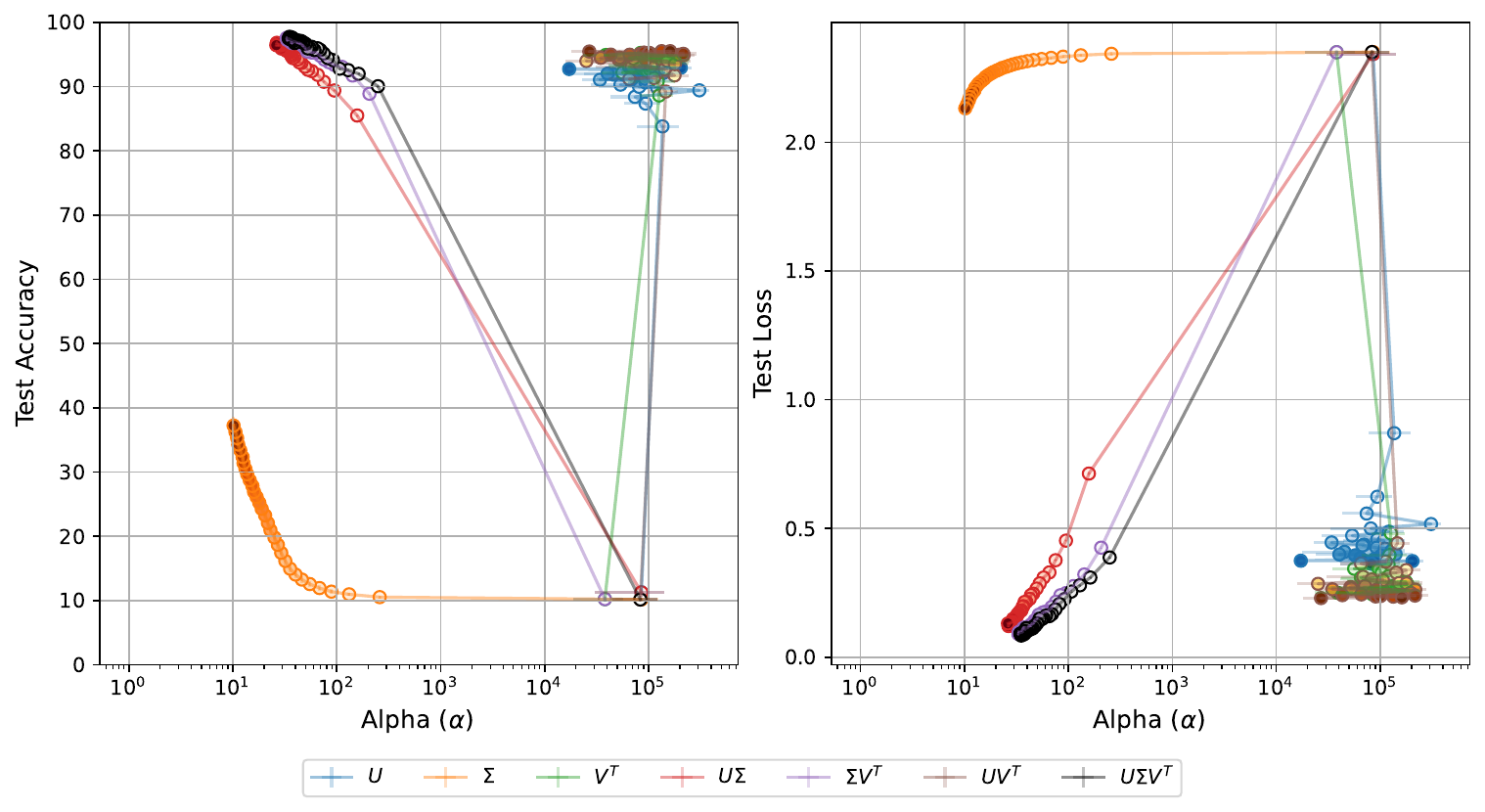}
        \caption{Alpha ($\alpha$) (log scaled)}
        \label{fig:alpha-trends_interventional_analysis}
    \end{subfigure}
    \begin{subfigure}[htbp]{0.32\linewidth}
        \includegraphics[width=\linewidth]{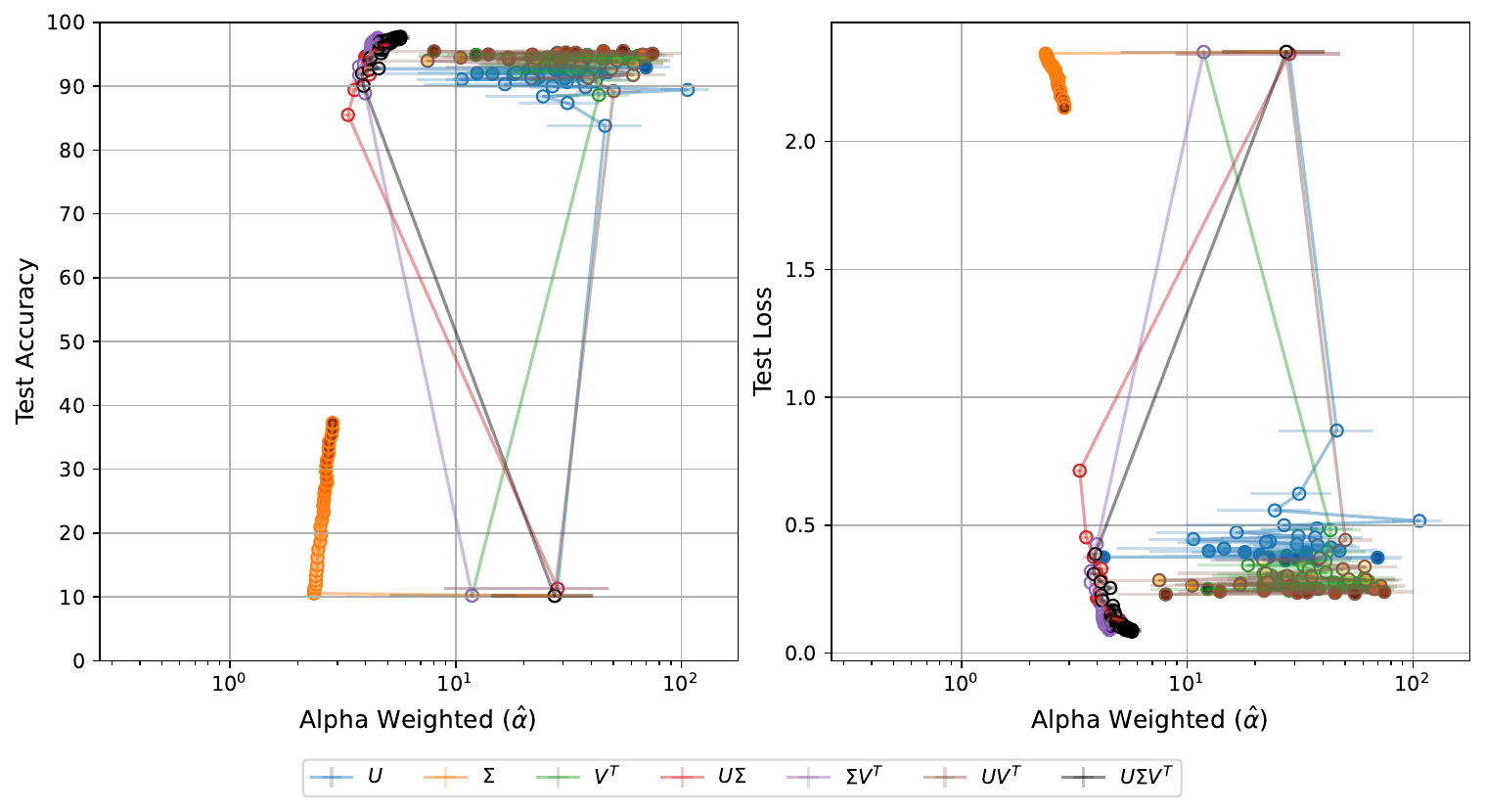}
        \caption{Alpha Weighted ($\hat{\alpha}$) (log scaled)}
        \label{fig:alpha-weighted-trends_interventional_analysis}
    \end{subfigure}
    \begin{subfigure}[htbp]{0.32\linewidth}
        \includegraphics[width=\linewidth]{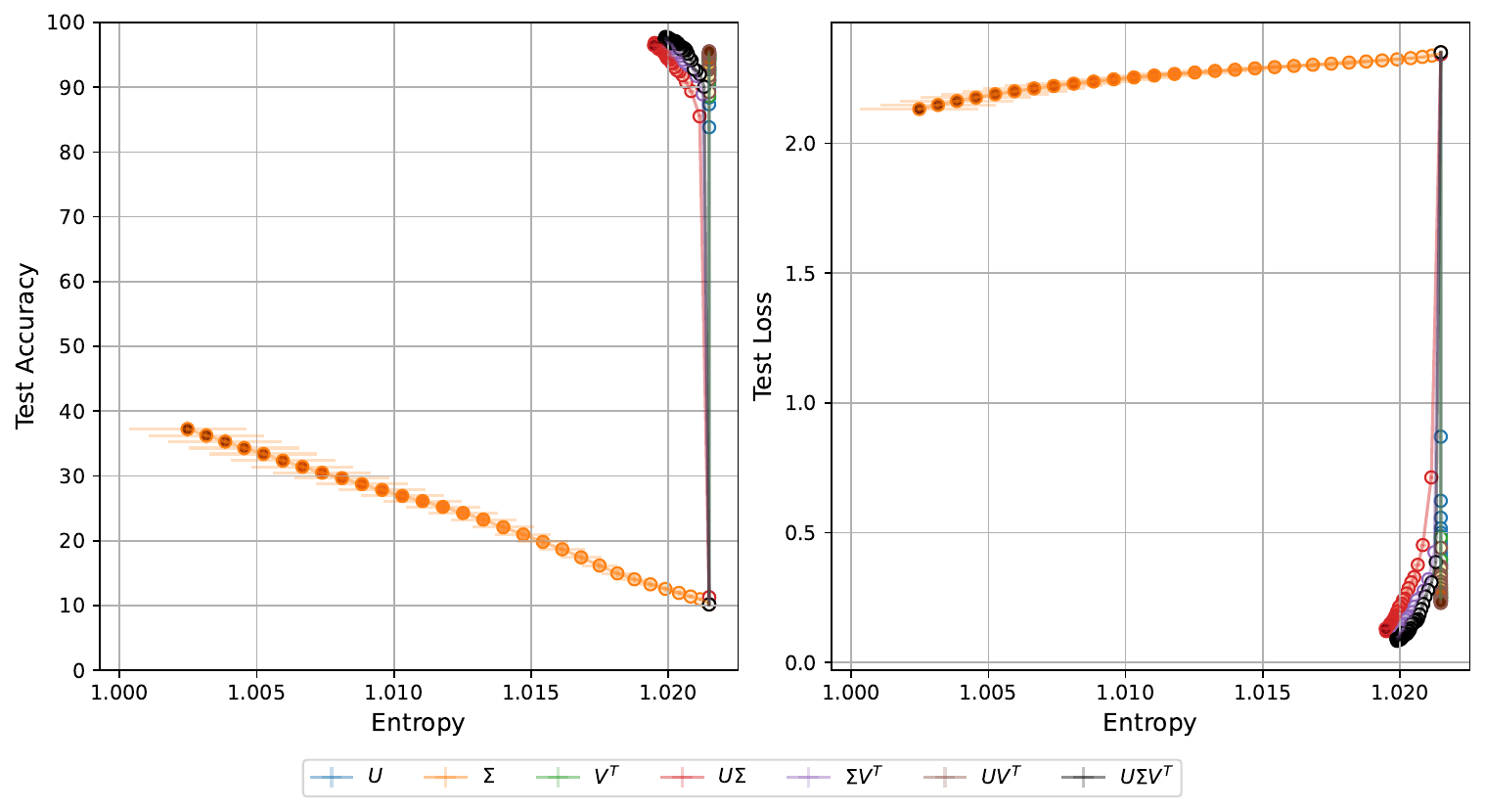}
        \caption{Entropy}
        \label{fig:entropy-trends_interventional_analysis}
    \end{subfigure}
    \hfill
    \begin{subfigure}[htbp]{0.32\linewidth}
        \includegraphics[width=\linewidth]{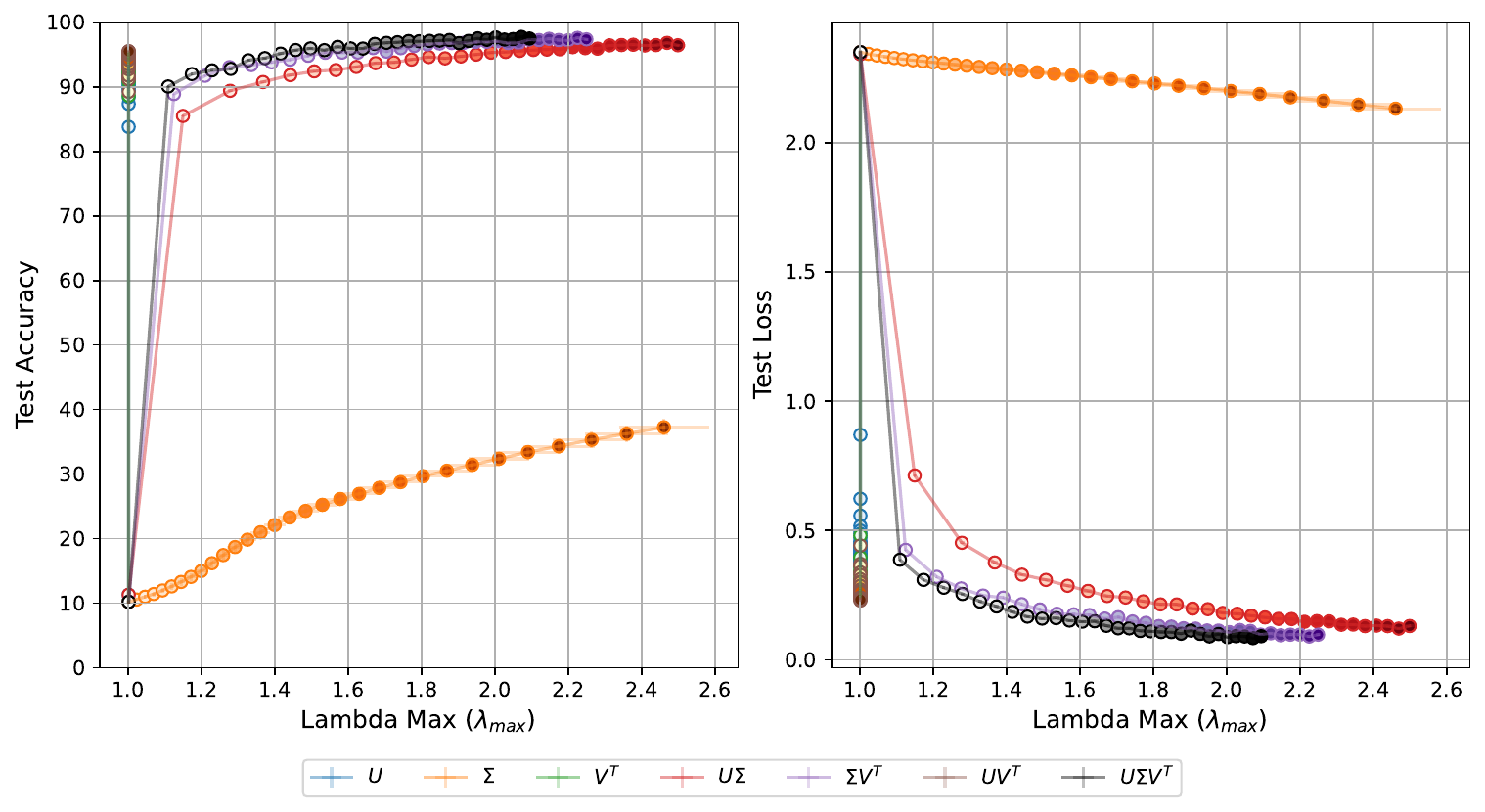}
        \caption{Lambda Max ($\lambda_{max}$)}
        \label{fig:lambda-max-trends_interventional_analysis}
    \end{subfigure}
    \begin{subfigure}[htbp]{0.32\linewidth}
        \includegraphics[width=\linewidth]{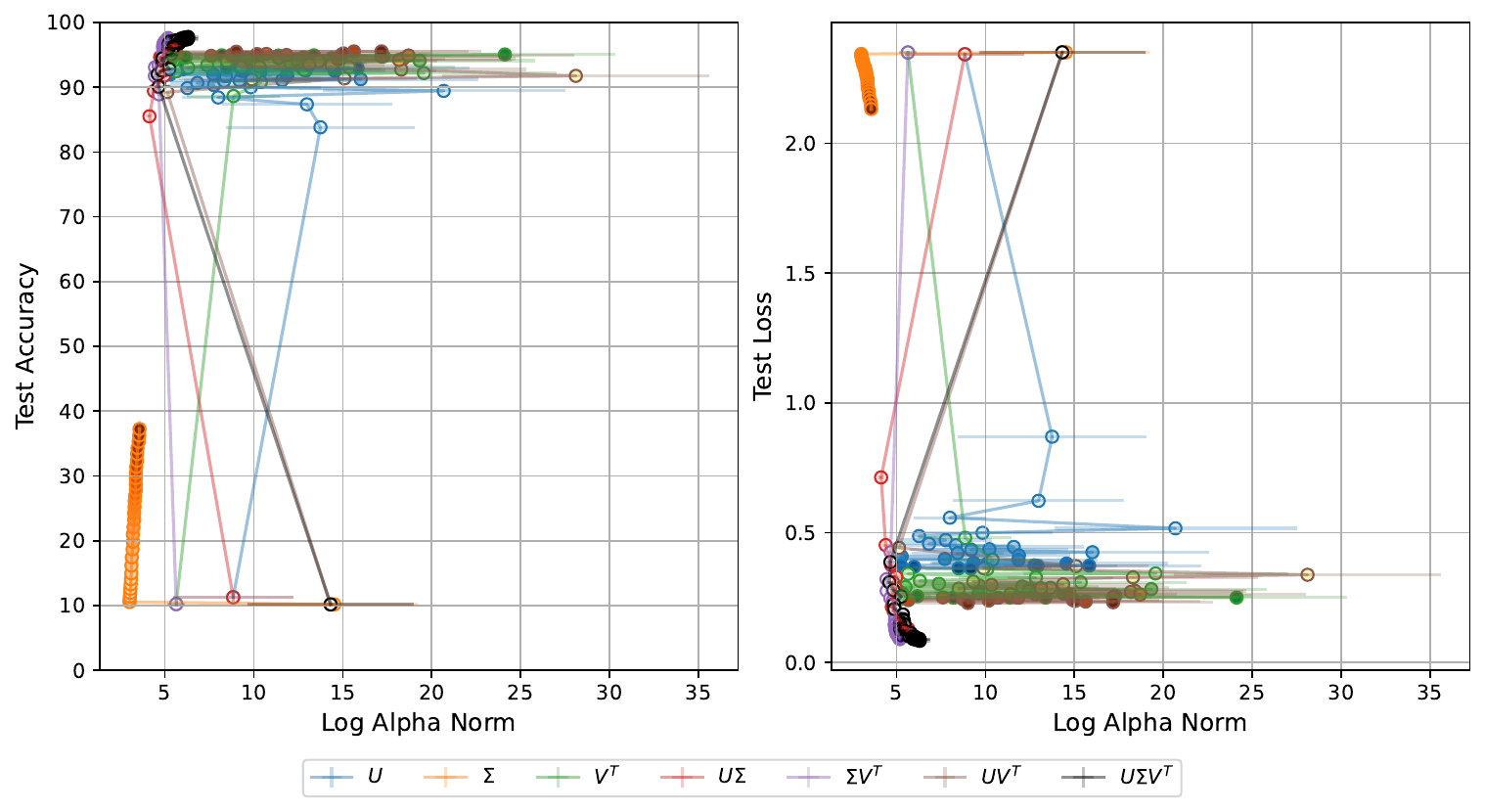}
        \caption{Log Alpha Norm ($\hat{\alpha}$}
        \label{fig:log-alpha-norm-trends_interventional_analysis}
    \end{subfigure}
    \begin{subfigure}[htbp]{0.32\linewidth}
        \includegraphics[width=\linewidth]{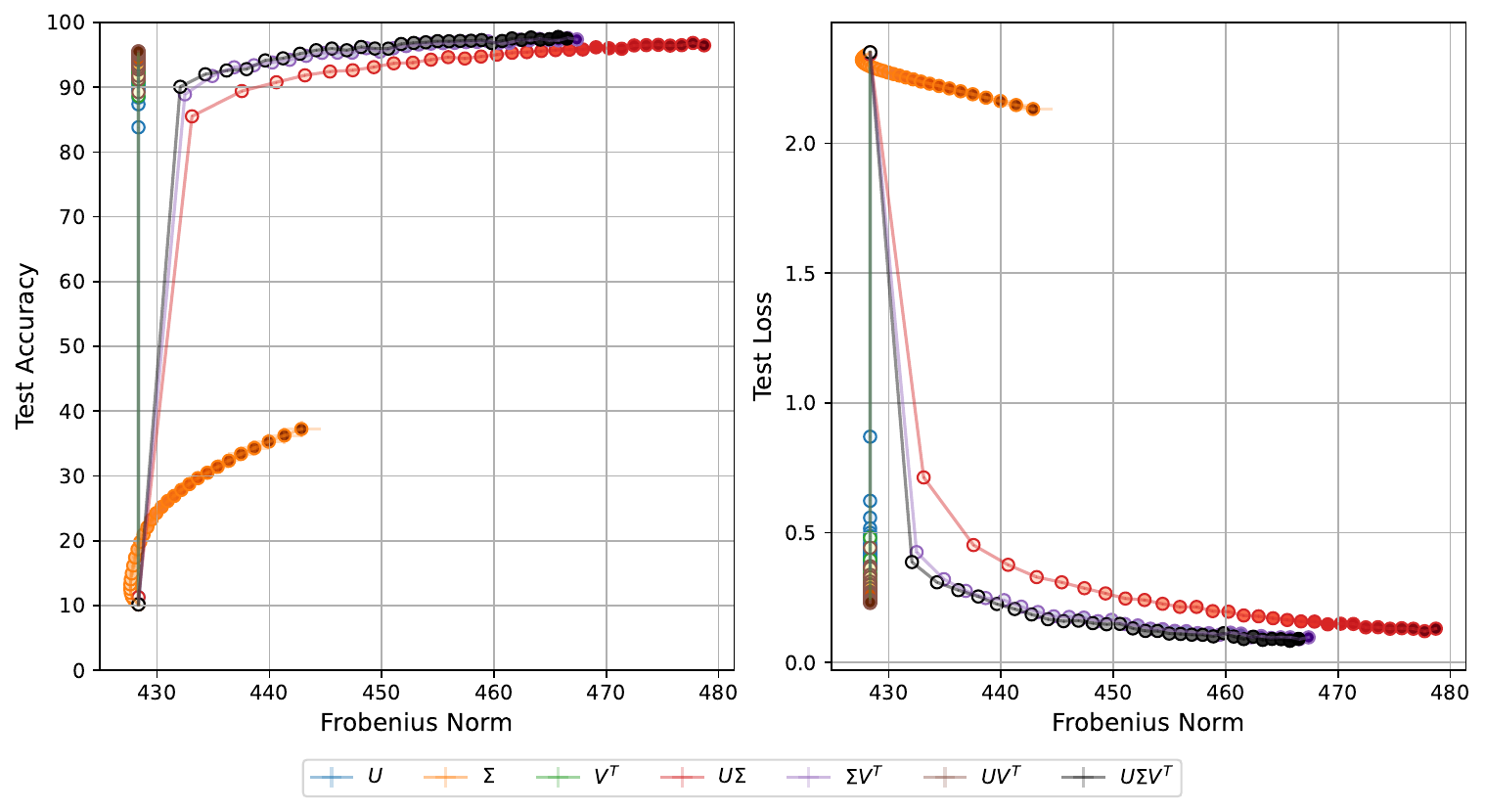}
        \caption{Frobenius Norm}
        \label{fig:fro-trends_interventional_analysis}
    \end{subfigure}
    \hfill
    \begin{subfigure}[htbp]{0.32\linewidth}
        \includegraphics[width=\linewidth]{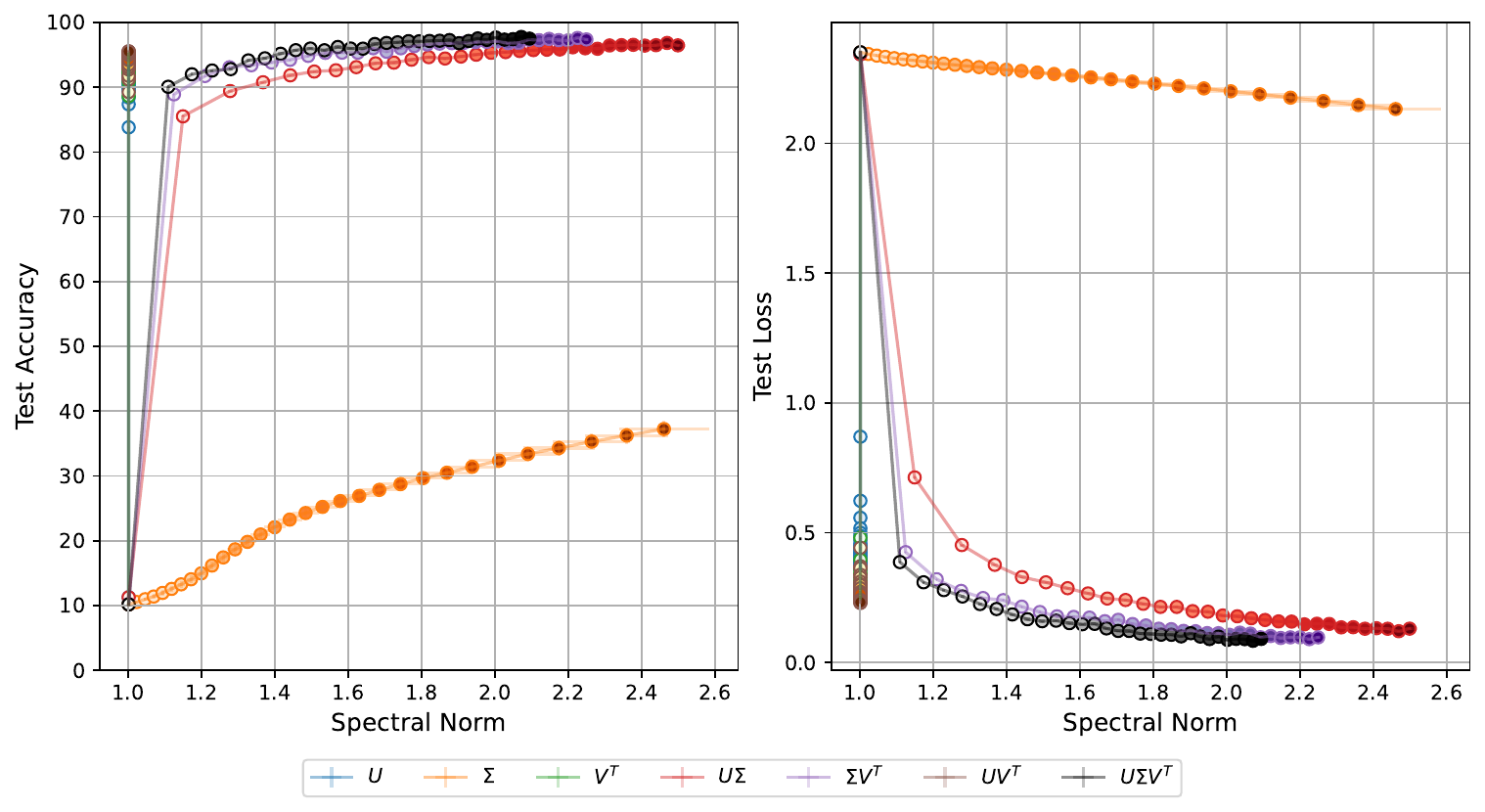}
        \caption{Spectral Norm}
        \label{fig:spec-trends_interventional_analysis}
    \end{subfigure}
    \begin{subfigure}[htbp]{0.32\linewidth}
        \includegraphics[width=\linewidth]{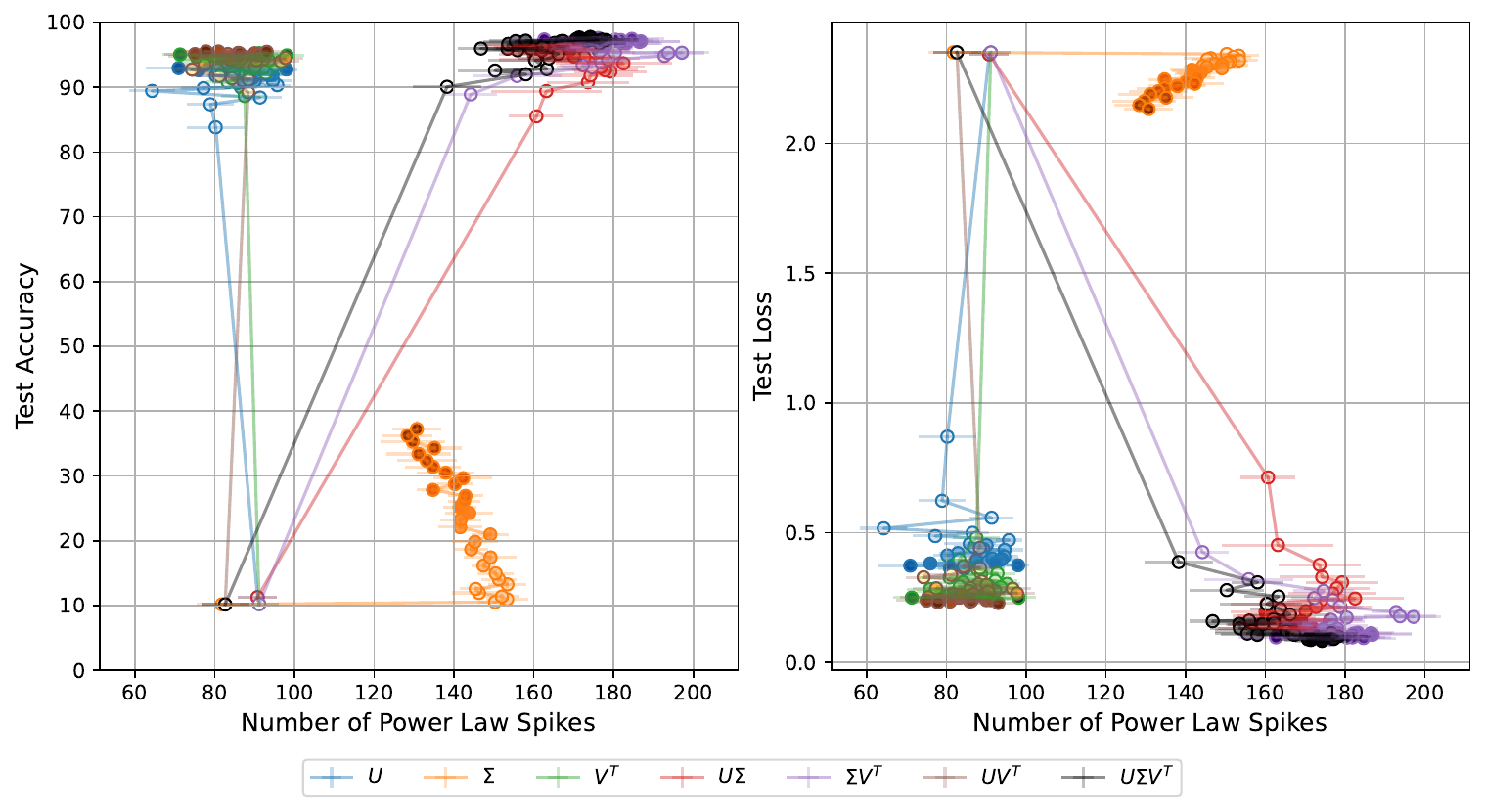}
        \caption{Number of Power Law Spikes}
        \label{fig:spikes-trends_interventional_analysis}
    \end{subfigure}
    \begin{subfigure}[htbp]{0.32\linewidth}
        \includegraphics[width=\linewidth]{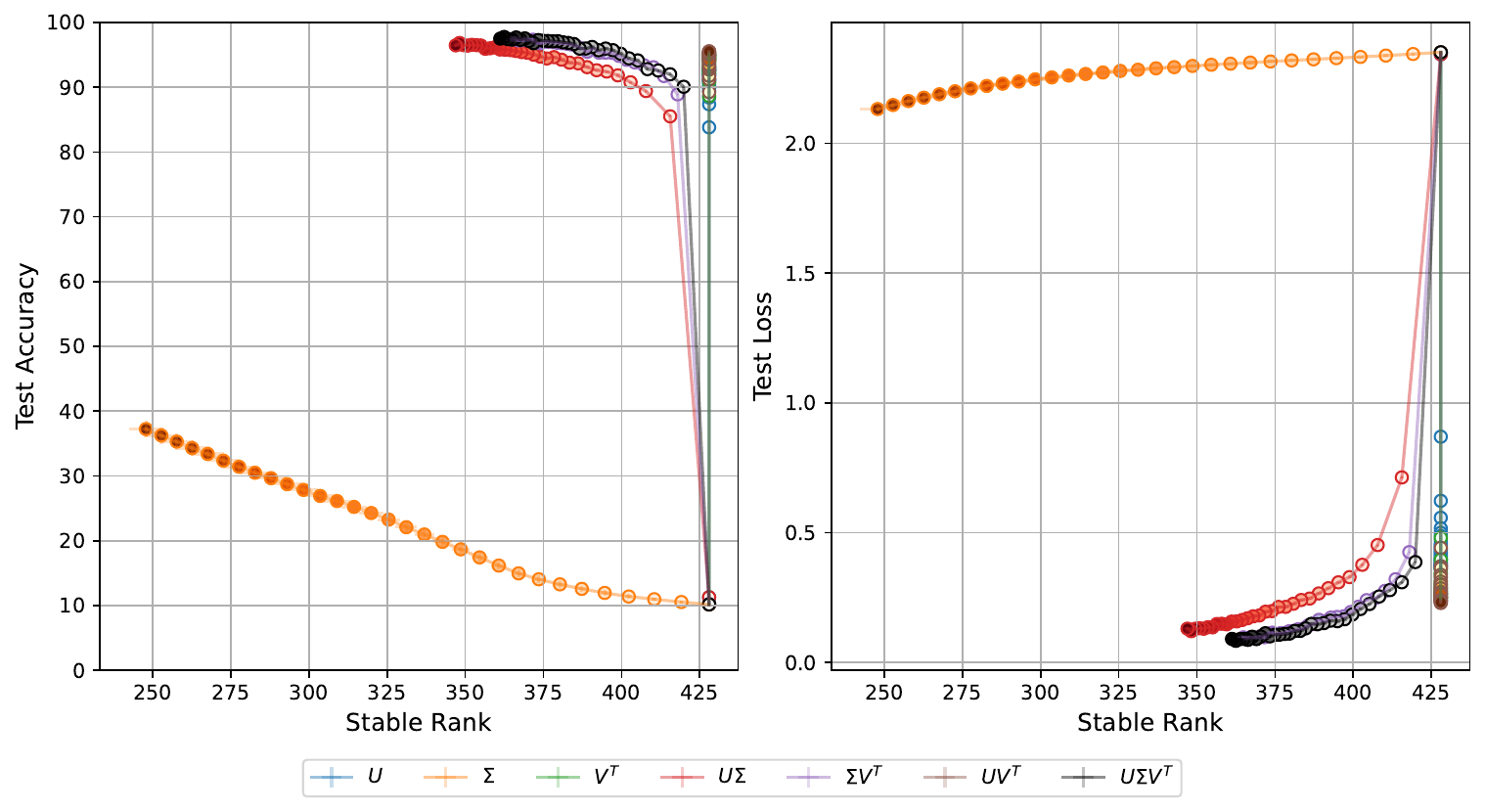}
        \caption{Stable Rank}
        \label{fig:stable-rank-trends_interventional_analysis}
    \end{subfigure}
    
    \caption{Model test performance throughout training on MNIST against data-free metrics. (left) test accuracy and (right) test loss. The $\circ$ represents a epoch. The colour intensity displays training progress, lighter points correspond to earlier epochs with darker points corresponding to later epochs. Each curve corresponds to training only one or more components of the parametrisation: $U$ $\Sigma$ $V^T$ $U\Sigma$,\space $\Sigma V^T$,\space $UV^T$\space and $U\Sigma V^T$. All other parameters are held fixed during training.  Performance is evaluated on the at the end of each epoch. The mean (line) and standard error from the mean (error bars) are reported from 10 runs. }
    \label{fig:trends_interventional_analysis}
\end{figure}
To understand the role of each respective metric in the training process we explore the each data-free metric's relationship with the test accuracy. Specifically focusing on how training with different model components effects the relationship each metric has with test accuracy throughout training.

\paragraph{Alpha ($\alpha$):} 
Figure~\ref{fig:alpha-trends_interventional_analysis} compares the performance throughout training with the global Alpha $\alpha$ when different subsets of the models parameters are updated during training. For parametrisations $U$, $V^T$\space and $UV^T$\space the $\alpha$ metric stays approximately around $10^5$. This behaviour is expected as the singular values are constrained to 1. The sight deviation from $10^5$ is introduction by the imperfect orthogonal constraints applied throughout training which results in slight negligible variations in $\Sigma$\space when the $U\Sigma V^T$\space is combined to calculate the metric. 

The parametrisations that utilise $\Sigma$\space exhibit different behaviour to $U$, $V^T$\space and $UV^T$. The $U\Sigma V^T$\space parametrisation achieves an  $\alpha$ value of 249.449 after the 1 epoch training, the $\alpha$ value continues to decrease throughout training while steadily improving in test performance,  reaching a global $\alpha$ value of 34.485 which a corresponding test accuracy of $97.499\%$. The $\Sigma$\space parametrisation reaches an $\alpha$ value of 259.177 within the first epoch of training correspond to a test accuracy of $10.542\%$. The  $\Sigma$\space parametrisation $\alpha$ value subsequently decreases, throughout training while improving in test performance reaching a value of 10.168 with a corresponding test accuracy of $37.246\%$. The introduction of $\Sigma$\space to $U$\space and $V^T$\space with $U\Sigma$\space and $\Sigma V^T$\space result in the global $\alpha$ values quickly deviating from the initial value after 1 epoch of training reaching an $\alpha$ value of 157.390 and 259.177 and test accuracies of $85.506\%$ and $88.898\%$ respectively. The $\alpha$ values continue to decrease throughout training reaching 26.220 and 33.338 while coinciding with a test accuracy of $96.458\%$ and $97.373\%$ for the  $U\Sigma$\space and $\Sigma V^T$\space  parametrisations respectively.

Overall, the Figure shows that optimal $\alpha$ metric values do not correlate with the resulting test performance of the model, as demonstrated with $U$, $V^T$\space and $UV^T$\space parametrisations achieving high accuracy while maintaining a large $\alpha$ value of $10^5$ and the $\Sigma$\space parametrisation achieving a low $\alpha$ metric of 10.168 with a test accuracy of $37.246\%$.

\paragraph{Alpha Weighted ($\hat{\alpha}$):}
Figure~\ref{fig:alpha-weighted-trends_interventional_analysis} compares the performance throughout training with the global Alpha Weighted $\hat{\alpha}$ when different subsets of the models parameters are updated during training. The Figure largely shows the same relationship between the parametrisations  $U$ $\Sigma$ $V^T$ $U\Sigma$,\space $\Sigma V^T$,\space $UV^T$\space and $U\Sigma V^T$ and the $\alpha$ metric observed in Figure~\ref{fig:alpha-trends_interventional_analysis}. With the $U$, $V^T$\space and $UV^T$\space parametrisations $\hat{\alpha}$ value  staying relatively fixed throughout training. 

The $\Sigma$\space parametrisation achieves an $\hat{\alpha}$ value of 2.359 after 1 epoch training, which then slowly increases as training progresses to 2.854 at the end of training. The $U\Sigma$\space and $\Sigma V^T$\space parametrisations follow a similar behaviour with a initial $\hat{\alpha}$ value of 3.338 and 3.975 after 1 epoch training respectively. With subsequent training the $\hat{\alpha}$ value tends towards slowly increasing resulting in an $\hat{\alpha}$ values of 4.851 and  4.602 for the $U\Sigma$\space and $\Sigma V^T$\space parametrisations respectively. In contrast the $U\Sigma V^T$\space parametrisation initially reaches an $\hat{\alpha}$ of 3.913 after 1 epoch of training and then subsequently fluctuates between increasing and decreasing values throughout training resulting in an $\hat{\alpha}$ of 5.683 at the end of training. 

The data in this section shows that there is not a strong correlation between the $\hat{\alpha}$ metric throughout training and the resulting test performance of the model, as shown with the large test performance difference witnessed regardless of the recorded $\hat{\alpha}$ value.

\paragraph{Entropy:} Figure~\ref{fig:entropy-trends_interventional_analysis} compares the performance throughout training with the global entropy when different subsets of the models parameters are updated during training. The $U$, $V^T$\space and $UV^T$\space parametrisations remain approximately constant throughout training while performance improves. The $\Sigma$\space parametrisation approximately linearly decreases in entropy throughout training, with an initial entropy of 1.021 and a end value of 1.002, marking a 0.019 decrease in entropy. The $U\Sigma$,\space $\Sigma V^T$\space and $U\Sigma V^T$ parametrisations also decrease in entropy throughout training, however the deviation from initialisation is a lot smaller, with a decrease in entropy of 0.002, 0.001 and 0.001 for each parametrisation respectively. These findings establish that there is a not a consistent correlation between model entropy and test performance that exists across parametrisations.

\paragraph{Lambda Max ($\lambda_{max}$):} Figure~\ref{fig:lambda-max-trends_interventional_analysis} compares the performance throughout training with the global $\lambda_{max}$ when different subsets of the models parameters are updated during training. The $U$, $V^T$\space and $UV^T$\space parametrisations remain approximately constant with a value of 1 throughout training while performance improves. The $\Sigma$\space parametrisation steadily increases the $\lambda_{max}$ value from 1.022 to 2.460 at the end of training. The initial increase in $\lambda_{max}$ is gradual as demonstrated with the change in $\lambda_{max}$ from epoch 1 to epoch 10 being 0.236, however later in training the increase in $\lambda_{max}$ is more substantial with the last 10 epochs resulting in a difference of 0.776. 

The inclusion of the $U$\space parametrisation to $\Sigma$\space via the  $U\Sigma$\space parametrisation, results in the $\lambda_{max}$ value exhibiting the opposite behaviour to the $\Sigma$\space parametrisation with a large initial increase in $\lambda_{max}$ at the start of training which then slows down as training progresses. Albeit the difference the $\lambda_{max}$ trajectory results in a similar $\lambda_{max}$ value of 2.498 at the end of training. The inclusion of of the $V^T$\space parametrisation to $\Sigma$\space via the  $\Sigma V^T$\space parametrisation, changes the training trajectory of $\lambda_{max}$. The training trajectory closely mimics the training trajectory of $U\Sigma$\space parametrisation however the change of $\lambda_{max}$ between epochs is consistently smaller, resulting in a $\lambda_{max}$ value of 2.248 at the end of training. The $U\Sigma V^T$\space parametrisation $\lambda_{max}$ training trajectory closely mimics the $\Sigma V^T$\space parametrisation. Starting with a small initial increase in $\lambda_{max}$ which then gradually increases through training resulting in a $\lambda_{max}$ value of 2.093 at the end of training. These results show that although $\lambda_{max}$ can correlate with performance increases, see $\Sigma$ $U\Sigma$,\space $\Sigma V^T$\space and $U\Sigma V^T$\space parametrisations, it is not required to change for a model to learn a generalisable solution, see $U$, $V^T$\space and $UV^T$\space parametrisations.

\paragraph{Log Alpha Norm:}
Figure~\ref{fig:log-alpha-norm-trends_interventional_analysis} compares the performance throughout training with the global Log Alpha Norm when different subsets of the models parameters are updated during training. The Figure largely shows the same relationship between the parametrisations $U$ $\Sigma$ $V^T$ $U\Sigma$,\space $\Sigma V^T$,\space $UV^T$\space and $U\Sigma V^T$ and the $\hat{\alpha}$ metric observed in  Figure~\ref{fig:alpha-weighted-trends_interventional_analysis}. The $U$, $V^T$\space and $UV^T$\space parametrisations Log Alpha Norm values vary significantly throughout training. This is expected, as Log Alpha Norm is defined as $\log(||X||^{\alpha}_{\alpha})$, where $||X||^{\alpha}_{\alpha}$  = $\sum\limits_{i}^M\lambda_{i}^{\alpha}$, (see Section~\ref{app:data_free_metrics}) and therefore inherits the variability from the calculation of the $\alpha$ metric. The $\Sigma$\space parametrisation Log Alpha Norm value quickly decreases after 1 epoch training going from a value of 14.527 to 3.012. The Log Alpha Norm value then slowly increases throughout training, with small changes between epochs, reaching a value of 3.573. The $U\Sigma$,\space $\Sigma V^T$, and $U\Sigma V^T$\space parametrisations exhibit similar behaviour to the $\Sigma$\space parametrisation however the initial decrease is lower going from 8.84 and 5.636 to 4.131 and 5.134 respectively. The parametrisations Log Alpha Norm then subsequently gradually increases throughout training as observed with the  $\Sigma$\space parametrisation. Figure~\ref{fig:log-alpha-norm-trends_interventional_analysis} shows that while the Log Alpha Norm can appear to correlate with improvements in test performance throughout training, this correlation is not consistently observed across all parametrisations namely,\space $U$, $V^T$\space and $UV^T$\space. 

\paragraph{Frobenius Norm:} Figure~\ref{fig:fro-trends_interventional_analysis} compares the performance throughout training with the global Frobenius Norm when different subsets of the models parameters are updated during training. 

The $U$, $V^T$\space and $UV^T$\space parametrisations improved in test performance throughout training while the global Frobenius Norm stayed approximately constant throughout training, reflecting that the singular values are held fixed at 1 throughout training. 

The $\Sigma$\space parametrisation global Frobenius Norm initially decreases early in training reaching its lowest value at epoch 5 of 427.618 a 0.726 decrease from the initial value before training while corresponding to an improvement in test accuracy of $2.395\%$. The global Frobenius Norm subsequently increases with every epoch passing the initial global Frobenius Norm at epoch 12, with a test accuracy of $19.828\%$ marking a $9.649\%$ improvement from initialisation. The resultant global Frobenius Norm at the end of training is 442.849 corresponding with a test accuracy of $37.246\%$.

The $U\Sigma$,\space $\Sigma V^T$\space and $U\Sigma V^T$\space parametrisations global Frobenius Norm follow a similar behaviour as observed with the Lambda Max $\lambda_{max}$ metric in Figure~\ref{fig:lambda-max-trends_interventional_analysis}. The $U\Sigma$\space has a large initial increase in the global Frobenius Norm value which then gradually increases throughout training to a value of 478.707 with a test accuracy of $96.458\%$. The $\Sigma V^T$\space parametrisation follows a similar training trajectory as the $U\Sigma V^T$\space parametrisation, however makes a larger initial increase during epoch 1 to a global Frobenius Norm of 432.480 in comparison $U\Sigma V^T$\space reaches a value of 432.068 which is 0.412 smaller. While the $U\Sigma V^T$\space parametrisation reaches a test accuracy of $90.066\%$ whereas the $\Sigma V^T$\space parametrisation reaches a test accuracy of $88.898\%$ marking a $1.168\%$ difference in test accuracy. Both the $\Sigma V^T$\space and $U\Sigma V^T$\space parametrisations steadily increase the global Frobenius Norm throughout training, with $\Sigma V^T$\space reaching a higher value of 467.382 at the end of training marking a 0.892 increase over the $U\Sigma V^T$\space parametrisation. While the $U\Sigma V^T$\space parametrisation reaches a test accuracy of $97.499\%$ marking a $0.126\%$ improvement over the $\Sigma V^T$\space parametrisation at the end of training.

In summary Figure~\ref{fig:fro-trends_interventional_analysis} shows that for a model to generalise and achieve good test accuracy the Frobenius norm of the layer weights is not required to change, see factors $U$\space $V^T$\space and $UV^T$. Furthermore enabling the Frobenius norm to change has limited impact on improving the overall performance of the model throughout training.

\paragraph{Spectral Norm:} Figure~\ref{fig:spec-trends_interventional_analysis} compares the performance throughout training with the global Spectral Norm when different subsets of the models parameters are updated during training. The training trajectories of the global Spectral Norm closely matches the behaviour observed with Lambda Max ($\lambda_{max}$) in Figure~\ref{fig:lambda-max-trends_interventional_analysis}. The $U$, $V^T$\space and $UV^T$\space parametrisations global Spectral Norm stays relatively fixed throughout training. The $\Sigma$\space parametrisation global Spectral Norm  gradually increases through training reaching the highest value while achieving the lowest test accuracy. The $U\Sigma$,\space $\Sigma V^T$\space and $U\Sigma V^T$\space parametrisations observe an initial large increase in the global Spectral Norm which then continues to gradually increase with $U\Sigma$\space reaching the highest value. The $\Sigma V^T$\space and $U\Sigma V^T$\space parametrisations reach approximately the same global Spectral Norm value.

\paragraph{Number of Power Law Spikes:} Figure~\ref{fig:spikes-trends_interventional_analysis} compares the performance throughout training with the global Number of Power Law Spikes when different subsets of the models parameters are updated during training.  As observed with the $\alpha$,\space $\hat{\alpha}$ and Log Alpha Norm metrics in Figures ~\ref{fig:alpha-trends_interventional_analysis}, ~\ref{fig:alpha-weighted-trends_interventional_analysis} and ~\ref{fig:log-alpha-norm-trends_interventional_analysis} respectively the $U$, $V^T$\space and $UV^T$\space parametrisations improved in test performance throughout training while the global Number of Power Law Spikes stayed relatively fixed throughout training, with the variation introduced by fitting a power law to singular values are all approximately 1 throughout training.

In contrast to the $\alpha$,\space $\hat{\alpha}$ and Log Alpha Norm metrics the  global Number of Power Law Spikes increases from the initial value for the $\Sigma$ $U\Sigma$,\space $\Sigma V^T$\space and $U\Sigma V^T$\space parametrisations. 

The $\Sigma$\space parametrisation global Number of Power Law Spikes subsequently fluctuates throughout training with alternating between increasing and decreasing within training epochs. Eventually resulting in a Number of Power Law Spikes of 130.717 with a test accuracy of $37.246\%$.

The $U\Sigma$\space parametrisation first epoch of training results in the largest  increase in the global Number of Power Law Spikes from 90.750 to 160.667, corresponding to a test accuracy of $85.506\%$. In the subsequent training epochs the global Number of Power Law Spikes oscillates with a variance of 51.772 between 155.133 and 182.483, resulting in a global Number of Power Law Spikes of 162.250 with a test accuracy of $96.458\%$ at the end of training. 

The $\Sigma V^T$\space parametrisation first epoch of training additional results in a large increase in the global Number of Power Law Spikes from 91.150 to 144.217, corresponding to a test accuracy of $88.898\%$, although a smaller increase than the $U\Sigma$\space parametrisation. As observed with the $U\Sigma$\space parametrisation in subsequent training epochs the global Number of Power Law Spikes oscillates between increasing and decreasing, resulting in a global Number of Power Law Spikes of 162.633 and a test accuracy of $97.373\%$. 

The $U\Sigma V^T$\space parametrisation follows a similar trend as the $U\Sigma$\space and $\Sigma V^T$\space parametrisations while improving in test accuracy, however with a smaller initial increase of the global Number of Power Law Spikes after 1 epoch training increasing from 82.633 to 138.267. The global Number of Power Law Spikes subsequently increases and decreases from epoch to epoch resulting in a global Number of Power Law Spikes of 171.750 and a test accuracy of $97.499\%$ at the end of training. 

These results indicate the the Number of Power Law Spikes is an inherently noisy metric with high variably from epoch to epoch, which can be attributed to the relationship it has with the fitting of $\alpha$, as the number of power law spikes directly corresponds to how many samples where used to estimate $\alpha$. Furthermore these results show that Number of Power Law Spikes does not need to change in order for the model to generalise.

\paragraph{Stable Rank:} Figure~\ref{fig:stable-rank-trends_interventional_analysis} compares the performance throughout training with the global Stable Rank when different subsets of the models parameters are updated during training. The training trajectories of the global Stable Rank closely matches the behaviour observed with Lambda Max ($\lambda_{max}$) and Spectral Norm in Figures~\ref{fig:lambda-max-trends_interventional_analysis} and \ref{fig:stable-rank-trends_interventional_analysis}. However instead of increasing the Stable rank decreases throughout training. The $U$,\space $V^T$\space and $UV^T$\space parametrisations global Stable Rank stays relatively fixed throughout training. The $\Sigma$\space parametrisation global Stable Rank  gradually decreases throughout training reaching the lowest value while achieving the lowest test accuracy. The $U\Sigma$,\space $\Sigma V^T$\space and $U\Sigma V^T$\space parametrisations observe an initial small decrease in the global Stable Rank which then continues to gradually decrease with $U\Sigma$\space reaching the lowest value. The $\Sigma V^T$\space and $U\Sigma V^T$\space parametrisations reach approximately the same global Stable Rank value.

\subsection{Discussion}
The results from this section show that a model can achieve good test accuracy and generalise on a given task with fixed unbiased singular values ($\Sigma = \mathbf{I}$) and therefore exhibit no change in the spectral properties of the layers of the model. Furthermore, when $\Sigma$ is trained on its own ($U$ and $V^T$ fixed) the model does not achieve comparable performance to any other factor, moreover the inclusion of $\Sigma$ with $U$ and $V^T$ only marginally improves performance over training with $U$ and $V^T$ independently with an increase of 3.222\% and 2.449\% respectively. These results suggest that the dominate factor in learning can be attributed to the left and right singular vectors. This means that over-interpreting the spectral properties of layers can lead to erroneous and misleading conclusions surrounding generalisation, given that an that they are not required to change for a model to effectively generalise.

\section{The Reliability of  the Power Law Fit (\texorpdfstring{$\alpha$}{α})}
\label{app:reliability_of_alpha}

One of the fundamental assumptions of HTSR theory is that the tail of the eigenvalues fit and can be accurately described by a power law distribution, hence the $\alpha$ metric~\cite{weightwatcher}. However, although the power law distribution can be fit to data, this does not mean it is the best or most apt distribution to describe the data~\cite{clauset2009power}. Therefore the following sections \S~\ref{app:fcn_gof}, \S~\ref{app:lsm_gof} explore the power law distribution assumption by performing a goodness of fit test comparing the power law distribution to the exponential, the log-normal, truncated power law and the stretched exponential distributions  with the power-law package~\cite{alstott2014powerlaw}. A p-value of 0.05 is used to disambiguate between statistically meaningful differences in the resulting test normalised likelihood ratios. A negative likelihood ratio indicates support for the alterative distribution. 

Fitting a power law requires finding the minimum value of the tail $x_{min}$, which is done utilising the Kolmogorov-Smirnov (KS) distance~\cite{clauset2009power}. When there exists one minimal across all possible $x_{min}$ values fitting the tail is trivial. However, in practise this is not the case, and furthermore  as this is fitted with a maximum likelihood estimation the selection of $x_{min}$ can be susceptible to noise.~\cite{clauset2009power} suggests as a rule of thumb that more than or equal to 50 samples can be used to obtain reliable estimates of distributional fit. Given the possibility of outliers in Figure~\ref{fig:model_init_vs_model_end} in \S~\ref{app:xmin_selection} we explore if the selection of $x_{min}$ can explain this occurrence and find that ensuring 50 points are used to estimate the tail that we find a stronger correlation between the initial $\alpha$ and the $\alpha$ at the end of training.

Overall this section shows that the eigenvalues of the ReLU FCN 5x512 models can be statistically significantly better described by the alternative distributions explored, namely the lognormal, truncated power law and the stretched exponential. For the large scale models there is inconclusive evidence to suggest that the power law distribution is the most appropriate distribution to describe the eigenvalues of these layers. Therefore in lieu of a clear statistical preference towards the power law distribution, as cautioned in~\cite{clauset2009power}, interpreting the data as fitting a power law may lead to erroneous and misleading conclusions about the nature of the data, in this case the spectral properties of the trained weight matrices.  Furthermore identifying the tail accurately has inherent challenges and selecting the $x_{min}$ with the smallest KS distance may result inaccurate findings. 

\subsection{Small Scale Models (ReLU FCN 5x512)}
\label{app:fcn_gof}
As all 100 ReLU FCN 5x512 models where trained on the same dataset with the same initialisation method and training strategy we ensemble the eigenvalues for each respective layer from the 100 models to provide a more representative sample of the eigenvalues of each layer. Concretely for FC1 to FC5 we have 51200 data points and FC6 we have 1000 data points. We compare the goodness of fit of the ensemble eigenvalues of each layer against the power law and alternative in Table~\ref{tab:gof-100-mnist}.  Table~\ref{tab:gof-100-mnist} shows that the statistically significant evidence that that eigenvalues of the layers can be better described as either a log-normal distribution, truncated power law or a stretched exponential (note negative normalised likelihood ratios and p-values < 0.5). 
This provides statistically significant evidence that the power-law distribution is not a suitable model for describing the eigenvalues of this model and demonstrates a clear preference to the alternatives. 

\begin{table}[H]
\centering
\caption{ReLU FCN 5x512: Goodness of fit test of the power law distribution against the alternatives with corresponding normalised likelihood ratio (LR) (a negative value indicates support for the alternative) and p-value. For reliable fitting a minimum of 100 samples were required to fit the tail.}
\label{tab:gof-100-mnist}
\resizebox{\textwidth}{!}{%
\begin{tabular}{|c|cc|cc|cc|cc|}
\hline
\multirow{2}{*}{\textbf{Layer}} & \multicolumn{2}{c|}{\textbf{Exponential}}                    & \multicolumn{2}{c|}{\textbf{Lognormal}}                      & \multicolumn{2}{c|}{\textbf{Truncated Power Law}}            & \multicolumn{2}{c|}{\textbf{Stretched Exponential}}          \\ \cline{2-9} 
                       & \multicolumn{1}{c|}{Normalised LR} & P-value        & \multicolumn{1}{c|}{Normalised LR} & P-value        & \multicolumn{1}{c|}{Normalised LR} & P-value        & \multicolumn{1}{c|}{Normalised LR} & P-value        \\ \hline
FC1                    & \multicolumn{1}{c|}{6.838}         & 0.000 & \multicolumn{1}{c|}{-3.829}        & 0.000 & \multicolumn{1}{c|}{-5.026}        & 0.000 & \multicolumn{1}{c|}{-3.151}        & 0.002 \\ \hline
FC2                    & \multicolumn{1}{c|}{7.522}         & 0.000 & \multicolumn{1}{c|}{-5.502}        & 0.000 & \multicolumn{1}{c|}{-5.303}        & 0.000 & \multicolumn{1}{c|}{-5.858}        & 0.000 \\ \hline
FC3                    & \multicolumn{1}{c|}{7.633}         & 0.000 & \multicolumn{1}{c|}{-5.764}        & 0.000 & \multicolumn{1}{c|}{-5.535}        & 0.000 & \multicolumn{1}{c|}{-6.152}        & 0.000 \\ \hline
FC4                    & \multicolumn{1}{c|}{7.135}         & 0.000 & \multicolumn{1}{c|}{-6.014}        & 0.000 & \multicolumn{1}{c|}{-5.767}        & 0.000 & \multicolumn{1}{c|}{-6.431}        & 0.000 \\ \hline
FC5                    & \multicolumn{1}{c|}{8.224}         & 0.000 & \multicolumn{1}{c|}{-5.121}        & 0.000 & \multicolumn{1}{c|}{-4.826}        & 0.000 & \multicolumn{1}{c|}{-5.450}        & 0.000 \\ \hline
FC6                    & \multicolumn{1}{c|}{-9.115}        & 0.000 & \multicolumn{1}{c|}{-4.732}        & 0.000 & \multicolumn{1}{c|}{-9.116}        & 0.000 & \multicolumn{1}{c|}{-4.864}        & 0.000 \\ \hline
\end{tabular}%
}

\end{table}
\subsection{Large Scale Models}
\label{app:lsm_gof}

In this section we explore the goodness of fit of the power law to eigenvalues of the pretrained models explored in \S~\ref{sec:large-scale} namely ResNet34, ViT, GPT2 GPT2-Large and Pythia-1B layers. We compare the goodness of fit for each layer and report how many layers either favoured each distribution compared or return an inconclusive results, where an inconclusive result is a result that reports a p-value greater than 0.05, The results for each respective model are shown in Tables~\ref{tab:alpha-reliability}. 

The Tables show that the power law distribution is minimally and rarely favoured across models and when it is favoured this  typically occurs in comparison to the exponential distribution. Furthermore the  majority of comparisons are inconclusive and thereby show no preference to a particular distribution. This means that there is not sufficient statistical evidence to support the use of the power law distribution to fit the tails of the eigenvalues as there exist multiple distributions that can provide comparably good fits to the data. 

\begin{table}[H]
\centering
    \caption{Comparison of distribution fits across layers using likelihood ratio tests. Counts show the number of layers favouring each distribution. Inconclusive indicates tests with a p-value greater than 0.05.}
    \label{tab:alpha-reliability}
    \begin{subtable}[b]{\textwidth}
        \centering
        \caption{ResNet34}
        \label{tab:resnet-alpha-reliability}
        \begin{tabular}{|c|c|c|c|}
            \hline
            \textbf{Comparison} & \textbf{Power Law Favoured} & \textbf{Alt Favoured} & \textbf{Inconclusive} \\ \hline
            Power Law vs Exponential & 4 (10.811\%) & 4 (10.811\%) & 29 (78.378\%) \\ \hline
            Power Law vs Log Normal & 0 (0.000\%) & 10 (27.027\%) & 27 (72.973\%) \\ \hline
            Power Law vs Truncated Power Law & 0 (0.000\%) & 17 (37.838\%) & 23 (62.162\%) \\ \hline
            Power Law vs Stretched Exponential & 1 (2.703\%) & 10 (27.027\%) & 26 (70.270\%) \\ \hline
        \end{tabular}
    
    \end{subtable}
    
    \begin{subtable}[b]{\textwidth}
        \centering
        \caption{ViT}
        \label{tab:vit-alpha-reliability}
        \begin{tabular}{|c|c|c|c|}
        \hline
        \textbf{Comparison} & \textbf{Power Law Favoured} & \textbf{Alt Favoured} & \textbf{Inconclusive} \\ \hline
        Power Law vs Exponential & 4 (8.000\%) & 2 (4.000\%) & 44 (88.000\%) \\ \hline
        Power Law vs Log Normal & 0 (0.000\%) & 7 (14.000\%) & 43 (86.000\%) \\ \hline
        Power Law vs Truncated Power Law & 0 (0.000\%) & 9 (18.000\%) & 41(82.000\%) \\ \hline
        Power Law vs Stretched Exponential & 0 (0.000\%) & 7 (14.000\%) & 43 (86.000\%) \\ \hline
        \end{tabular}        
    \end{subtable}

    \begin{subtable}[b]{\textwidth}
        \centering
        \caption{GPT2}
        \label{tab:gpt-alpha-reliability}
        \begin{tabular}{|c|c|c|c|}
            \hline
            \textbf{Comparison} & \textbf{Power Law Favoured} & \textbf{Alt Favoured} & \textbf{Inconclusive} \\ \hline
            Power Law vs Exponential & 7 (14.286\%) & 3 (6.122\%) & 39 (79.592\%) \\ \hline
            Power Law vs Log Normal & 0 (0.000\%) & 3 (6.122\%) & 46 (93.878\%) \\ \hline
            Power Law vs Truncated Power Law & 0 (0.000\%) & 7 (14.286\%) & 42 (85.714\%) \\ \hline
            Power Law vs Stretched Exponential & 1 (2.041\%) & 3 (6.122\%) & 45 (91.837\%) \\ \hline
        \end{tabular}
    \end{subtable}

    \begin{subtable}[b]{\textwidth}
        \centering
        \caption{GPT2-Large}
        \label{tab:gpt-large-alpha-reliability}
        \resizebox{\textwidth}{!}{%
        \begin{tabular}{|c|c|c|c|}
            \hline
            \textbf{Comparison}                & \textbf{Power Law Favoured} & \textbf{Alt Favoured} & \textbf{Inconclusive} \\ \hline
            Power Law vs Exponential           & 14 (9.655\%)                & 0 (0.000\%)               & 131 (90.345\%)        \\ \hline
            Power Law vs Log Normal            & 0 (0.000\%)                     & 0 (0.000\%)               & 145 (100.000\%)           \\ \hline
            Power Law vs Truncated Power Law   & 0 (0.000\%)                     & 18 (12.414\%)         & 127 (87.586\%)        \\ \hline
            Power Law vs Stretched Exponential & 0 (0.000\%)                     & 3 (2.069\%)           & 142 (97.931\%)        \\ \hline
        \end{tabular}%
        }
    \end{subtable}

    \begin{subtable}[b]{\textwidth}
        \centering
        \caption{Pythia-1B}
        \label{tab:pythia-1b-alpha-reliability}
        \resizebox{\textwidth}{!}{%
        \begin{tabular}{|c|c|c|c|}
            \hline
            \textbf{Comparison}                & \textbf{Power Law Favoured} & \textbf{Alt Favoured} & \textbf{Inconclusive} \\ \hline
            Power Law vs Exponential           & 5 (7.812\%)                 & 0 (0.000\%)               & 59 (92.188\%)         \\ \hline
            Power Law vs Log Normal            & 0 (0.000\%)                     & 0 (0.000\%)               & 64 (100.000\%)            \\ \hline
            Power Law vs Truncated Power Law   & 0 (0.000\%)                     & 2 (3.125\%)           & 62(96.875\%)          \\ \hline
            Power Law vs Stretched Exponential & 1 (1.562\%)                 & 0 (0.000\%)               & 63 (98.438\%)         \\ \hline
        \end{tabular}%
        }
    \end{subtable}

\end{table}

\subsection{Selection of \texorpdfstring{$\boldsymbol{x_{min}}$}{Xmin}}
\label{app:xmin_selection}
\begin{figure}[H]
    \centering
    \includegraphics[width=0.75\linewidth]{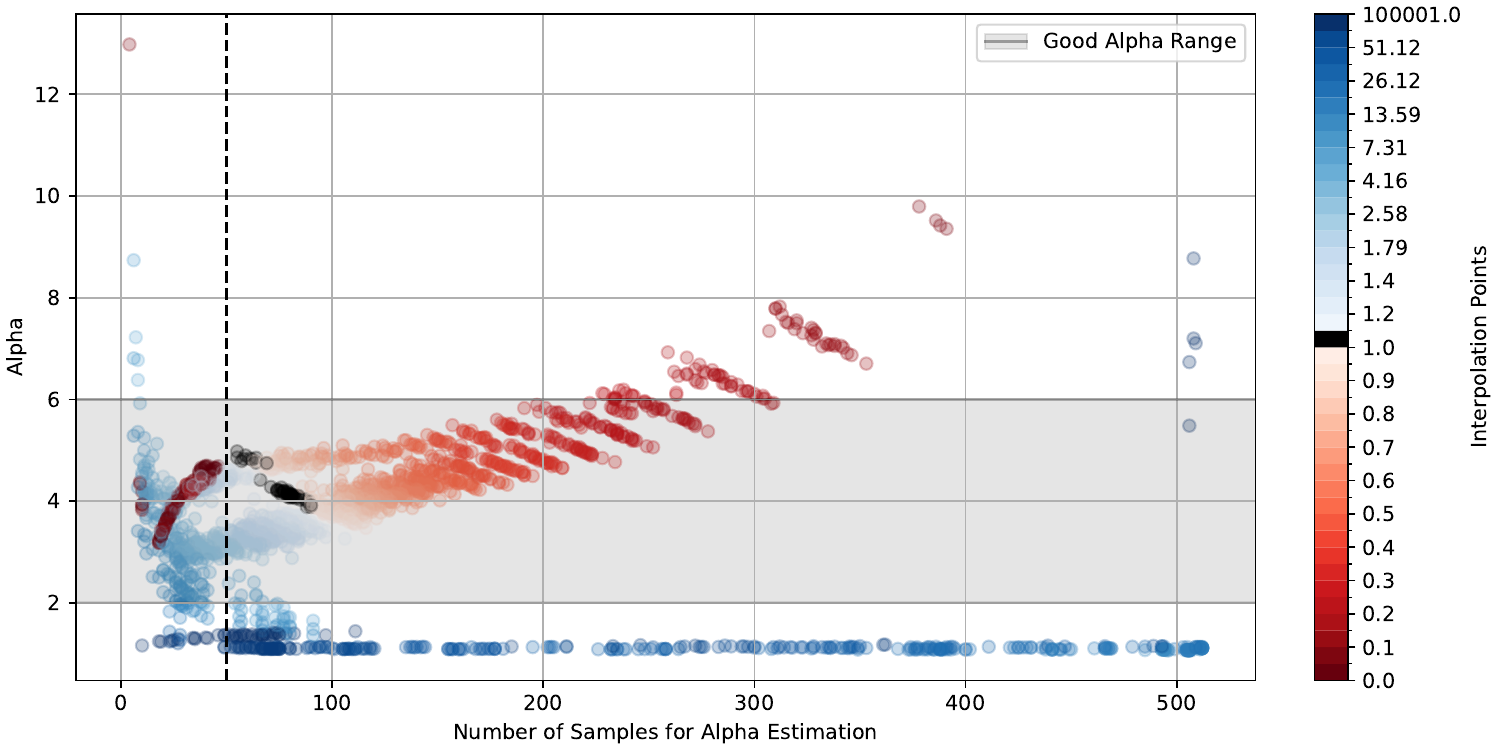}
    \caption{Number samples used for alpha estimation for each layer. Each dot represents a layer from a trained model }
    \label{fig:interpolation-samples-to-alpha}
\end{figure}

In \S~\ref{sec:alpha_init_training} Figure~\ref{fig:model_init_vs_model_end} we observe that there is a strong correlation with model $\alpha$ at initialisation and at the end of training, however for the interpolated points of 0.1-0 appeared to be outliers to this trend achieving $\alpha$ values lower than the trend would expect. As~\cite{clauset2009power} states that 50 samples can be used a rule of thumb for the minium required samples to obtain reliable estimates of distributional fit, we explore in Figure~\ref{fig:interpolation-samples-to-alpha} how many data points were used in fitting the power-law to the tail of the data (the black vertical line represents the 50 sample cut off). The Figure shows that many of the interpolation points $\alpha$ values are estimated with fewer than 50 points. 

\begin{figure}[H]
    \centering
    \includegraphics[width=0.8\linewidth]{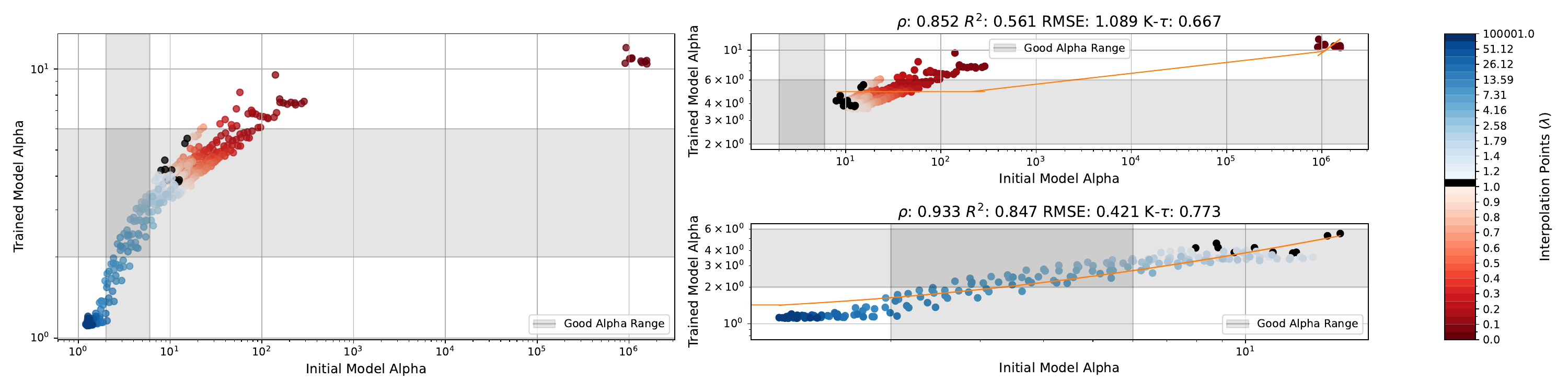}
    \caption{Trained model $\alpha$ vs initial model $\alpha$ (\textbf{Left}). Trained model $\alpha$ vs initial model $\alpha$ for interpolations $\lambda=0$ to $\lambda=1$ (\textbf{Top Right}) and $\lambda=1$ to $\lambda=100001$ (\textbf{Bottom Right})}
    \label{fig:model_init_vs_model_end_fixed}
\end{figure}

Given the results in Figure~\ref{fig:interpolation-samples-to-alpha} and guided by the role of thumb we re-estimate the $\alpha$ values using the power law package~\cite{alstott2014powerlaw} with a constraint of a minimum of 50 points used to estimate alpha for layers FC1-FC5, for layer FC6 $\alpha$ is still calculated from the 10 eigenvalues, to ensure the model average $\alpha$ is maintained and enable comparability with the original figure.  The results are shown in Figure~\ref{fig:model_init_vs_model_end_fixed}. It is important to note that $\lambda$ points 0 and 10001 are not plotted as suitable $x_{min}$ could not be found for the initialisation which expected given they are at the extremes of the interpolation. In this figure the trend that $\alpha$ at initialisation strongly correlates to the $\alpha$ at the end of training is substantially clearer, which provides further credence to the statement that the 'good' range [2,6] is a consequence of our standard initialisation regime and regularisation from SGD. This underscores that $\alpha$ does not necessarily emerge because this metric is indicative of generalisation.

\section{Large Scale Model Details}

The pretrained models were accessed from huggingface:
\begin{itemize}
    \item ResNet34: \url{https://huggingface.co/timm/resnet34.tv_in1k}
    \item ViT: \url{https://huggingface.co/timm/vit_base_patch16_224.augreg_in1k}
    \item GPT2: \url{https://huggingface.co/openai-community/gpt2}
    \item GPT2-Large: \url{https://huggingface.co/openai-community/gpt2-large}
    \item Pythia-1B: \url{https://huggingface.co/EleutherAI/pythia-1b}
\end{itemize} 

\section{Computational Resources}
\label{app:Resources}

Model training and test evaluation was conducted using A100 and L40S GPUs and the AMD EPYC 7742 CPU. 

\subsection{Small Scale Analysis of Alpha Under Re-Initialisation and Re-Randomisation}

To train each model took approximately 1 minute. Given 100 models were trained, training the models took approximately 100 GPU minutes. The evaluation time of the model was approximately 0.1 GPU minutes. Therefore total GPU usage was approximately 220 GPU minutes.

\subsection{Large Scale Analysis of Alpha Under Reparametrisation}
To evaluate the performance of the reparametrisation of each layer for each model took approximately:

\begin{itemize}
    \item ResNet34: 0.23 GPU minutes
    \item ViT: 1 GPU Minutes
    \item GPT2: 0.23 GPU minutes
    \item GPT2-Large: 1 GPU Minutes
    \item Pythia-1B: 1 GPU Minutes
\end{itemize}

The number of layers evaluated for each model was:

\begin{itemize}
    \item ResNet34: 37
    \item ViT: 43
    \item GPT2: 45
    \item GPT2-Large: 145
    \item Pythia-1B: 64
\end{itemize}

We evaluated the ResNet34, ViT, GPT2, GPT2-Large model with 10 random initialisation. For the Pythia-1B we use one initialisation and one random initialisation. This experiment took approximately: 2196.6 GPU minutes.

\subsection{Alpha At Initialisation \& Training With Any Alpha}

Training and evaluating each model took approximately 4.16 CPU minutes. We trained a total of 410 models (41 interpolation points and 10 random seed) totalling  approximately 1705.6 CPU minutes or 28.42 CPU hours.

\subsection{Generalisation With Fixed Alpha}
Training and evaluating and evaluating each model took approximately 90 GPU minutes. Given we explored 3 alpha conditions and averaged over 10 models the total training times was 2700 GPU minutes.

\subsection{SVD Ablation for Learning Dynamics}

Training with each factor took approximately:
\begin{itemize}
    \item $U$: 12 GPU minutes
    \item $\Sigma$: 12 GPU minutes
    \item $V$: 80 GPU minutes
    \item $U\Sigma$: 15 GPU minutes
    \item $\Sigma V:$ 85 GPU minutes
    \item $UV:$ 90 GPU minutes
    \item $U\Sigma V:$ 100 GPU minutes
\end{itemize}

Given we average over 10 models the total training time was approximately 3940 GPU minutes.

\subsection{Total Training and Evaluation time}

The total training and evaluation time was approximately: 8928.6 GPU minutes and 1705.6 CPU minutes, or more aptly: 148.81 GPU hours and 28.43 CPU hours. 

\section{Datasets and Licences}
\label{app:Licences}

\paragraph{Image Datasets}
\begin{itemize}
    \item MNIST~\cite{lecun2010mnist} has an CC licence.
    \item ImageNet~\cite{imagenet15russakovsky}  ImageNet states that: `The data is available for free to researchers for non-commercial use.' giving a CC BY-NC licence.
\end{itemize}

\paragraph{Language Datasets}
\begin{itemize}
    \item WikiText-103~\cite{merity2016pointer} has a CC BY-SA 4.0 licence. 
\end{itemize}

\newpage
\section*{NeurIPS Paper Checklist}

\begin{enumerate}

\item {\bf Claims}
    \item[] Question: Do the main claims made in the abstract and introduction accurately reflect the paper's contributions and scope?
    \item[] Answer: \answerYes{}
    \item[] Justification: In Section~\ref{sec:small-scale-analysis} and Section~\ref{sec:large-scale} we show from low complexity datasets (MNIST) and small scale models (ReLU FCN 5x512) settings to high complexity datasets (ImageNet and WikiText-103) on million (ResNet34 and ViT) and hundreds of million (GPT and GPT Large) parameter scales across modalities (Image and Language) that weight analysis provided by random matrix theory, primarily alpha, are not reparametrisation invariant under critical layer theory. Following this in Section~\ref{sec:fixed_alpha} we show that ideas around the criticality of optimal alpha values in models are misplaced; through simple modifications to optimisation we show that well generalising models can arise for an arbitrary alpha value. In addition, initialisation plays a strong role in the alpha found at the end of training. Our empirical evidence shows that optimal alpha values are not a prerequisite of generalisation and in Appendix Section~\ref{sec:exp-spectral-vs-training} we quantify the contribution by alpha through ablations studies using singular value decomposition. Finally, in Appendix Section~\ref{sec:data-based-metrics} we show in small scale settings that data-based metrics provide crucial insights between robust and critical layers that fundamentally relate to existing theories of the lottery ticket hypothesis, Section~\ref{sec:lottery-ticket-hypothesis}.  

\item {\bf Limitations}
    \item[] Question: Does the paper discuss the limitations of the work performed by the authors?
    \item[] Answer: \answerYes{}{}
    \item[] Justification: In Appendix Section~\ref{app:Limitations} we clearly articulate in which settings our findings hold and their extensions to metrics that are not explored in the paper. We also provide additional modalities where our findings would need to be confirmed.

\item {\bf Theory assumptions and proofs}
    \item[] Question: For each theoretical result, does the paper provide the full set of assumptions and a complete (and correct) proof?
    \item[] Answer: \answerYes{}{} 
    \item[] Justification: To analyse data-free metrics in Theorem~\ref{thm:DinhCopy} we show (through an extension of Dihn et al~\cite{dinh2017sharp}) that for a majority of metrics values can be arbitrarily modified while preserving the same functional representation of the model. To study alpha, our paper provides a set of extensive empirical evaluations to explore the relationship between data-free metrics and generalisation. We acknowledge that formal proofs would be useful for our study of alpha, however the results presented concretely show that optimal alpha values are not causal for generalisation.

    \item {\bf Experimental result reproducibility}
    \item[] Question: Does the paper fully disclose all the information needed to reproduce the main experimental results of the paper to the extent that it affects the main claims and/or conclusions of the paper (regardless of whether the code and data are provided or not)?
    \item[] Answer: \answerYes{}{} 
    \item[] Justification: We provided full explanations for the range of experiments conducted in this paper with metric formalisations, model architectural details, optimisation hyperparameters and dataset details.

\item {\bf Open access to data and code}
    \item[] Question: Does the paper provide open access to the data and code, with sufficient instructions to faithfully reproduce the main experimental results, as described in supplemental material?
    \item[] Answer: \answerYes{}
    \item[] Justification: Notebooks are provided at \url{https://github.com/GMW99/data-free-metrics-dont-work} which demonstrate the method and analysis for the MNIST and LLM cases.  

\item {\bf Experimental setting/details}
    \item[] Question: Does the paper specify all the training and test details (e.g., data splits, hyperparameters, how they were chosen, type of optimizer) necessary to understand the results?
    \item[] Answer: \answerYes{} 
    \item[] Justification: The experimental settings for the core of the paper are provided to a level of detail that enable reproduction of the results and allow for full understanding of their experimental context.

\item {\bf Experiment statistical significance}
    \item[] Question: Does the paper report error bars suitably and correctly defined or other appropriate information about the statistical significance of the experiments?
    \item[] Answer: \answerYes{}
    \item[] Justification: We provide large sample sizes for experiments in the paper, typically calculating means from 100 models or variations with a few instances where we calculate means with 10 models. In tables we report means and $\pm{}$ 1 Standard Error of The Mean to best represent uncertainty~\cite{belia2005researchers}, we also   analyse the correlation between data-free and data-based metrics and test accuracy using the Spearman correlation coefficient $\rho$, coefficient of determination $R^2$, the root mean square error (RMSE) of the linear regression and Kendall’s tau measure (K-$\tau$). When conducting likelihood tests we utilise a p-value of 0.05.

\item {\bf Experiments compute resources}
    \item[] Question: For each experiment, does the paper provide sufficient information on the computer resources (type of compute workers, memory, time of execution) needed to reproduce the experiments?
    \item[] Answer: \answerYes{} 
    \item[] Justification: In Appendix Section\ref{app:Resources} we provide details on the compute resources used to run the experiments in this paper, we also detail the estimated total compute required to re-run the experiments in the paper.

\item {\bf Code of ethics}
    \item[] Question: Does the research conducted in the paper conform, in every respect, with the NeurIPS Code of Ethics \url{https://neurips.cc/public/EthicsGuidelines}?
    \item[] Answer: \answerYes{}
    \item[] Justification: Every author has reviewed the code of ethics and can confirm that out paper confirms in every aspect with the NeurIPS Code of Ethics.

\item {\bf Broader impacts}
    \item[] Question: Does the paper discuss both potential positive societal impacts and negative societal impacts of the work performed?
    \item[] Answer: \answerNA{} 
    \item[] Justification: Our results can be considered under foundational research and provide no association with particular applications or deployments. However, we do acknowledge that our results show that data-free metrics may not be suitable for understanding generalisation which impact interpretability efforts leveraging data-free metrics for neural networks across data domains.
    
\item {\bf Safeguards}
    \item[] Question: Does the paper describe safeguards that have been put in place for responsible release of data or models that have a high risk for misuse (e.g., pre-trained language models, image generators, or scraped datasets)?
    \item[] Answer: \answerNA{}{} 
    \item[] Justification: In the paper we provide no new models or scrapped datasets that have risk of misuse.

\item {\bf Licenses for existing assets}
    \item[] Question: Are the creators or original owners of assets (e.g., code, data, models), used in the paper, properly credited and are the license and terms of use explicitly mentioned and properly respected?
    \item[] Answer: \answerYes{} 
    \item[] Justification: In the paper we have appropriately cited assets that have been used and provided url's to specific datasets, models and tools where required. In Appendix Section~\ref{app:Licences} we explicltiy state the licenses for each dataset used in the paper.

\item {\bf New assets}
    \item[] Question: Are new assets introduced in the paper well documented and is the documentation provided alongside the assets?
    \item[] Answer: \answerNA{} 
    \item[] Justification: Our paper does not release new assets. However, details of datasets and models used for the empirical analysis of the paper are provided in the paper.

\item {\bf Crowdsourcing and research with human subjects}
    \item[] Question: For crowdsourcing experiments and research with human subjects, does the paper include the full text of instructions given to participants and screenshots, if applicable, as well as details about compensation (if any)? 
    \item[] Answer: \answerNA{}
    \item[] Justification: Our paper does not involve crowdsourcing nor research with human subjects. 

\item {\bf Institutional review board (IRB) approvals or equivalent for research with human subjects}
    \item[] Question: Does the paper describe potential risks incurred by study participants, whether such risks were disclosed to the subjects, and whether Institutional Review Board (IRB) approvals (or an equivalent approval/review based on the requirements of your country or institution) were obtained?
    \item[] Answer: \answerNA 
    \item[] Justification: Our paper does not involve crowdsourcing nor research with human subjects.

\item {\bf Declaration of LLM usage}
    \item[] Question: Does the paper describe the usage of LLMs if it is an important, original, or non-standard component of the core methods in this research? Note that if the LLM is used only for writing, editing, or formatting purposes and does \emph{not} impact the core methodology, scientific rigor, or originality of the research, declaration is not required.
    \item[] Answer: \answerNA{} 
    \item[] Justification: At no point in core method development of the paper were LLM's used.  

\end{enumerate}

\end{document}